\documentclass{article}

\usepackage[margin=1in]{geometry}
\usepackage{times}
\usepackage{microtype}
\usepackage{amsmath, amssymb, amsthm, mathtools}
\usepackage{bm}
\usepackage{graphicx}
\usepackage{subcaption}
\usepackage{booktabs}
\usepackage{multirow}
\usepackage{bbm}
\usepackage{algorithm}
\usepackage[noend]{algpseudocode}
\usepackage{natbib}
\usepackage{hyperref}
\usepackage[capitalize,noabbrev]{cleveref}
\usepackage{xcolor}
\usepackage{enumitem}
\usepackage{titling}
\usepackage{placeins}
\usepackage{tikz}

\usetikzlibrary{arrows.meta, shapes.geometric}

\newcommand{\maybeincludegraphics}[2][]{\includegraphics[#1]{#2}}

\newcommand{\E}{\mathbb{E}}

\newcommand{\A}{\mathcal{A}}
\newcommand{\B}{\mathcal{B}}
\newcommand{\Fwd}{F}

\newcommand{\design}{a}
\newcommand{\policy}{\pi_\theta}
\newcommand{\reward}{r}
\newcommand{\dist}{d}
\newcommand{\Dtrain}{\mathcal{D}}
\newcommand{\KL}{\mathrm{KL}}

\newcommand{\method}{\textsc{RLSF}}
\newcommand{\methodfull}{Reinforcement Learning from Simulator Feedback}
\newcommand{\bench}{\textsc{SciDesignBench}}

\title{\bench{}: Benchmarking and Improving Language Models for Scientific Inverse Design}

\author{
David van Dijk \quad Ivan Vrkic \\[4pt]
CellType Inc. \\[2pt]
\texttt{\{david,ivan\}@celltype.com}
}
\date{}

\begin{document}
\maketitle

\begin{abstract}
Many of the most important problems in science and engineering are inverse problems: given a desired outcome, find a design that achieves it. Evaluating whether a candidate meets the spec is often routine; a binding energy can be computed, a reactor yield simulated, a pharmacokinetic profile predicted. But searching a combinatorial design space for inputs that satisfy those targets is fundamentally harder. We introduce \bench{}, a benchmark of 520 simulator-grounded tasks across 14 scientific domains and five settings spanning single-shot design, short-horizon feedback, long-horizon refinement, and seed-design optimization. On the 10-domain shared-core subset, the best zero-shot model reaches only 29.0\% success despite substantially higher parse rates. Simulator feedback helps, but the leaderboard changes with horizon: Sonnet~4.5 is strongest in one-turn \emph{de novo} design, whereas Opus~4.6 is strongest after 20 turns of simulator-grounded refinement. Providing a starting seed design reshuffles the leaderboard again, demonstrating that constrained modification requires a fundamentally different capability from unconstrained de novo generation. We then introduce \method{}, a simulator-feedback training recipe. An \method{}-tuned 8B model raises single-turn success rates by 8--17 percentage points across three domains. Together, these results position simulator-grounded inverse design as both a benchmark for scientific reasoning and a practical substrate for amortizing expensive test-time compute into model weights.

\end{abstract}

\section{Introduction}
\label{sec:intro}

Across science and engineering, many of the hardest problems are \emph{inverse problems}: infer or construct an input that produces a desired output under a known forward process. Inverse problems are difficult because the inverse map is often ill-posed, many-to-one, and highly sensitive to constraints or noise~\citep{stuart2010inverse, arridge2019solving}. \emph{Inverse design} is the constructive version of this challenge. Designing a molecule with target ADMET properties, a dosing policy with target pharmacokinetics, a reactor with target conversion, or a quantum circuit with target state fidelity all fit the same template: evaluate a candidate with a simulator or predictor, then search a large design space for one that satisfies several quantitative targets simultaneously~\citep{gomez2018automatic, sanchez2020inverse, lee2023materials}.

This framing exposes a question that current scientific LLM evaluation leaves unanswered: \emph{can frontier language models solve scientific inverse problems, rather than merely explain them?} Existing benchmarks test scientific knowledge retrieval~\citep{rein2024gpqa, hendrycks2021mmlu}, research reasoning~\citep{laurent2024labbench, shao2024bixbench}, and mathematical problem-solving~\citep{arora2025scibench}, but they rarely require the model to emit a structured \emph{design} whose correctness is decided only after executing a scientific oracle. They also do not measure \emph{long-horizon refinement}: whether a model can use repeated simulator feedback over many steps to climb toward a solution.

We answer this question with \bench{}, a benchmark comprising 520 tasks (260 \emph{de novo}, 260 optimization) across 14 scientific domains, each equipped with a frozen forward oracle. Tasks are written as natural-language goals; the model must produce a structured JSON design, which is then validated, executed, and scored by the oracle. We evaluate seven frontier language models---GPT-5.2, Claude Opus~4.6, Claude Sonnet~4.5 and~4.6, Gemini~3.1~Pro, Gemini~2.0~Flash, and GPT-4o---across five evaluation modes that separate single-shot design from long-horizon refinement: 1-turn \emph{de novo}, 5-turn feedback, 20-turn long-horizon feedback, 1-turn optimization, and 20-turn long-horizon optimization.

Three benchmark findings stand out. First, one-turn \emph{de novo} design remains hard: even the strongest shared-core model reaches only 29.0\% success, far below its parse rate. Second, \emph{long-horizon} capability is distinct from one-turn competence. Sonnet~4.5 leads one-turn \emph{de novo} design, but Opus~4.6 becomes the strongest model after 20 turns of simulator-guided refinement. Models with similar zero-shot scores diverge sharply once feedback enters the loop. Third, seed-design optimization is a different problem again: Opus~4.6 leads 1-turn optimization even though Sonnet~4.5 leads 1-turn \emph{de novo} design, confirming that modifying an existing design is not the same capability as designing from scratch.

These results paint a clear picture: frontier models encode substantial scientific knowledge, but simulator-grounded inverse design remains unsolved. Simulator feedback helps, yet it is expensive at inference and still plateaus well below expert level on many domains. This motivates a second question: if an oracle can \emph{score} a design, can the same oracle be used to train a stronger goal-conditioned model? Our training contribution is a simulator-feedback recipe rather than a new optimizer. We treat the benchmark's forward oracles as RL environments, initializing a model with SFT and applying standard GRPO using simulator feedback as the reward signal:

\begin{enumerate}[nosep,leftmargin=*]
  \item \textbf{SFT}: Fine-tune Qwen3-8B with QLoRA on simulator-generated (goal, design) pairs to establish format compliance and basic domain priors.
  \item \textbf{GRPO}: Apply Group Relative Policy Optimization~\citep{shao2024deepseekmath} using the forward simulator as the reward signal, enabling the model to discover designs that actually achieve target outcomes.
\end{enumerate}

\paragraph{Contributions.}
\begin{enumerate}[nosep,leftmargin=*]
  \item \textbf{\bench{}}: An open benchmark of 520 simulator-grounded inverse-design tasks across 14 scientific domains, each with a reference forward oracle, difficulty levels L1--L4, and five evaluation settings that separate one-turn design from long-horizon refinement.
  \item \textbf{Frontier-model evidence}: A systematic evaluation of seven frontier language models showing that parse rates substantially overstate real scientific competence, that long-horizon feedback utilization is a separate capability, and that optimization reorders the leaderboard relative to \emph{de novo} design.
  \item \textbf{\method{} training recipe}: A benchmark-aligned simulator-feedback recipe (SFT $\to$ GRPO) that turns the same oracle interface used for evaluation into an RL training environment.
  \item \textbf{Proof-of-concept case studies}: A Qwen3-8B trained with \method{} improves from 30\%$\to$41\% on ADMET optimization, 24\%$\to$36\% on PK/PD dosing design, and 42\%$\to$59\% on molecular docking, showing that simulator-backed training can materially improve a smaller model on selected scientific design domains.
\end{enumerate}

\begin{figure*}[htbp]
\centering
% === Panel (a): Concept diagram (PaperBanana-generated) ===
\begin{subfigure}[t]{\textwidth}
\centering
\maybeincludegraphics[width=0.92\linewidth]{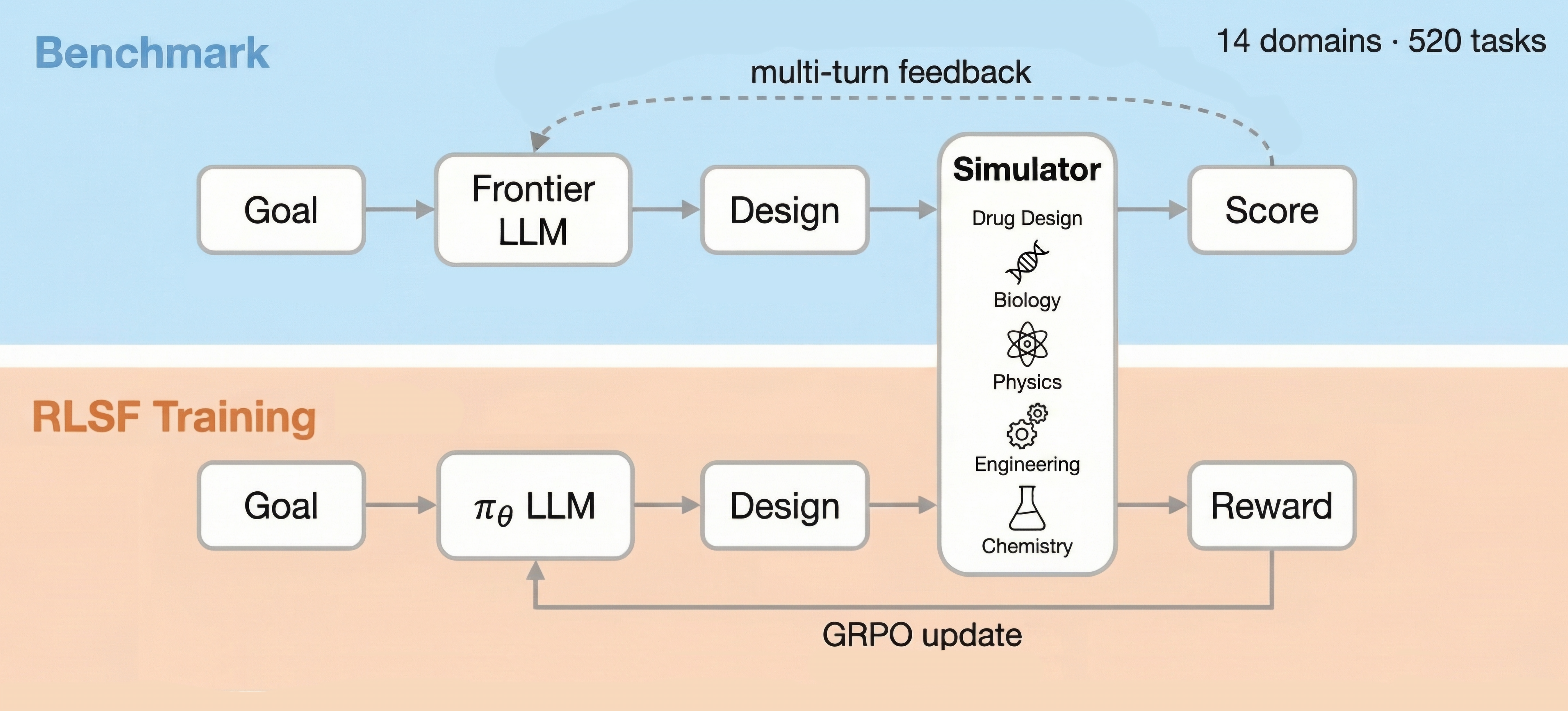}
\vspace{-4pt}
\caption{Framework: each task pairs a natural-language goal with a forward simulator across 14 scientific domains. At inference, the LLM produces a structured design that the simulator scores. Multi-turn evaluation feeds scores back as text (dashed); \method{} uses scores as RL rewards to train $\pi_\theta$ via GRPO (solid).}
\label{fig:overview_concept}
\end{subfigure}

\vspace{6pt}

% === Bottom row: panels (b) and (c) ===
\begin{subfigure}[t]{0.52\textwidth}
\centering
\maybeincludegraphics[width=\linewidth]{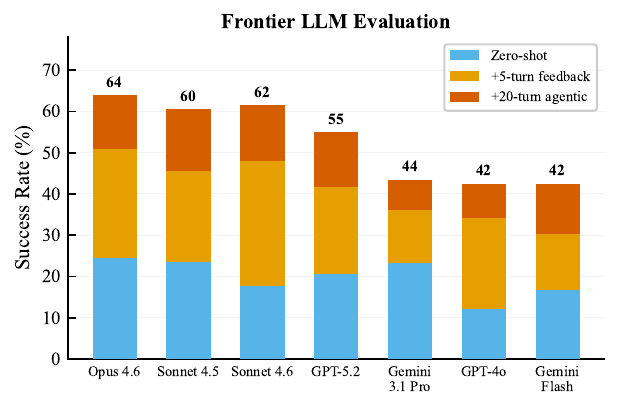}
\vspace{-12pt}
\caption{Frontier LLM evaluation: simulator feedback roughly doubles zero-shot performance, but models plateau below 68\%.}
\label{fig:overview_bar}
\end{subfigure}
\hfill
\begin{subfigure}[t]{0.44\textwidth}
\centering
\maybeincludegraphics[width=\linewidth]{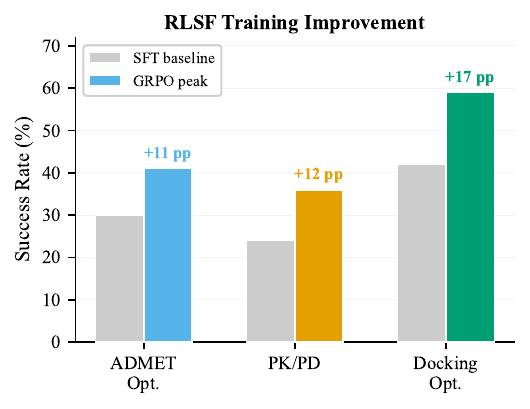}
\vspace{-12pt}
\caption{\method{} training: the same oracle interface can be reused as an RL environment for selected case-study domains.}
\label{fig:overview_training}
\end{subfigure}

\caption{\textbf{\bench{} overview.} (a)~Each benchmark task pairs a natural-language design goal with a ground-truth forward simulator. We evaluate frontier LLMs across five settings and train small models via \method{}. (b)~Frontier LLMs achieve only 12--26\% zero-shot; simulator feedback helps but saturates below 68\%. (c)~\method{} training improves an 8B model on three case studies: ADMET optimization (30\%$\to$41\%), PK/PD dosing (24\%$\to$36\%), and docking (42\%$\to$59\% with real AutoDock Vina).}
\label{fig:overview}
\end{figure*}

\section{Related Work}
\label{sec:related}

\paragraph{Inverse problems and inverse design.}
Inverse problems are a core theme in applied mathematics and computational science because the inverse map is often ill-posed, unstable, and many-to-one~\citep{stuart2010inverse, arridge2019solving}. Inverse design specializes this template to the constructive setting: rather than recovering a hidden cause, we seek a valid artifact or intervention that achieves a target behavior. This framing appears across molecular design, materials discovery, control, and engineering optimization~\citep{gomez2018automatic, sanchez2020inverse, lee2023materials}. \bench{} operationalizes this literature as a benchmark question for general-purpose language models: given only a target specification and a forward oracle, can a model produce a design that actually works?

\paragraph{Science benchmarks.}
A growing family of benchmarks evaluates LLMs on scientific tasks. \emph{Knowledge benchmarks} like GPQA~\citep{rein2024gpqa}, MMLU~\citep{hendrycks2021mmlu}, SciBench~\citep{arora2025scibench}, and SciEval~\citep{sun2024scieval} test factual recall and forward reasoning, but not design. \emph{Research-workflow benchmarks} like LAB-Bench~\citep{laurent2024labbench}, BixBench~\citep{shao2024bixbench}, and SciAssess~\citep{cai2024sciassess} probe literature use, experimental planning, or analysis, but they do not require generating artifacts scored by scientific simulators. \emph{Domain-specific benchmarks} like MatBench~\citep{dunn2020matbench} and Llamole~\citep{liu2024llamole} are closest in spirit, but they are narrower in domain scope and evaluation horizon. \bench{} differs along three axes simultaneously: breadth (14 domains), executable oracle-backed scoring, and multiple interaction horizons including long-horizon refinement.

\paragraph{General LLM benchmarks.}
Beyond science, LLM benchmarks span code generation (HumanEval~\citep{chen2021codex}, SWE-bench~\citep{jimenez2024swebench}) and mathematical reasoning (GSM8K~\citep{cobbe2021gsm8k}). These benchmarks validate execution or exact answers but do not test whether a model can solve a scientific inverse problem under domain-specific constraints.

\paragraph{Scientific LLMs.}
Domain-specific models like Galactica~\citep{taylor2022galactica}, SciGLM~\citep{zhang2024sciglm}, and ChemLLM~\citep{zhang2024chemllm} are trained on scientific corpora. While they improve question answering and literature synthesis, they are not trained with simulator feedback and have not been evaluated on inverse design tasks.

\paragraph{RL with execution or simulator feedback.}
RLHF~\citep{ouyang2022rlhf} and DPO~\citep{rafailov2024dpo} align LLMs with human preferences. GRPO~\citep{shao2024deepseekmath}, adopted by DeepSeek-R1~\citep{guo2025deepseekr1}, introduced group-relative advantages for efficient policy optimization. CodeRL~\citep{le2022coderl} applies execution feedback to code generation. OPRO~\citep{yang2024opro} uses LLMs as optimizers by iteratively prompting with past solutions, FunSearch~\citep{romera2024funsearch} evolves LLM-generated programs with execution feedback, and AlphaEvolve~\citep{novikov2025alphaevolve} scales population-based evolutionary search with LLM-guided mutation. EvoPrompting~\citep{chen2023evoprompting} combines LLMs with evolutionary search for neural architecture design. These are inference-time optimization approaches requiring costly per-goal search. \method{} instead trains the model weights with simulator feedback so that novel goals can be solved in a single forward pass.

\paragraph{Classical and learned inverse design.}
Many \bench{} domains have established solvers: OptKnock~\citep{burgard2003optknock} for metabolic engineering, GRAPE~\citep{khaneja2005grape} for quantum control, Ziegler--Nichols~\citep{ziegler1942optimum} for PID tuning, RNAinverse~\citep{hofacker1994rnafold} for RNA folding, Parks--McClellan~\citep{parks1972chebyshev} for filter design, and REINVENT~\citep{olivecrona2017reinvent} for molecular generation. Modern ML approaches use variational autoencoders~\citep{gomez2018automatic}, diffusion models~\citep{gruver2024finetuned}, GFlowNets~\citep{bengio2021gflownet}, and Bayesian optimization~\citep{stanton2022accelerating}. These methods are powerful but typically remain per-domain and often per-goal. \method{} targets a different operating point: an amortized, goal-conditioned language model that interacts with many simulators through the same natural-language plus JSON interface.

\FloatBarrier
\section{\bench{}: The Benchmark}
\label{sec:benchmark}

\subsection{Task Formulation}

Each \bench{} task consists of:
\begin{enumerate}[nosep,leftmargin=*]
  \item A \textbf{goal specification} $g = (\text{description}, \text{targets}, \text{constraints}, \text{difficulty})$ describing the desired outcome in natural language with quantitative targets.
  \item A \textbf{forward oracle} $\Fwd: \A \to \B$ that maps a design $\design \in \A$ to an outcome $b \in \B$.
  \item A \textbf{reward function} $\reward(b, g)$ measuring how well the outcome achieves the goal.
  \item A \textbf{success predicate} implemented per domain, which checks whether the required targets and constraints are satisfied.
\end{enumerate}

The LLM receives the goal as a natural-language prompt and must produce a JSON design. The pipeline is: \textbf{Goal} $\to$ \textbf{LLM} $\to$ \textbf{JSON Design} $\to$ \textbf{Parse} $\to$ \textbf{Validate} $\to$ \textbf{Execute} $\to$ \textbf{Score}. Success is therefore determined only after executing the forward oracle, which makes \bench{} a benchmark of simulator-grounded inverse problems rather than scientific recall.

\subsection{Domains}

\bench{} spans 14 scientific domains, each with a distinct forward oracle (\Cref{tab:domains}). Detailed descriptions of the inverse design tasks, design spaces, and oracles for all 14 domains are provided in \Cref{app:benchmark_details}.

\begin{table*}[htbp]
\centering
\caption{Overview of \bench{} domains grouped by scientific area. Each domain has 20 tasks (5 per difficulty level L1--L4) in both \emph{de novo} (design from scratch) and \emph{optimization} (modify existing design) settings. \textbf{Fidelity}: \emph{Exact} = mathematically precise; \emph{Model} = simplified scientific model solved exactly; \emph{Empirical} = data-derived correlations. Perturbation is de novo only; ADMET is optimization only.}
\label{tab:domains}
\scriptsize
\setlength{\tabcolsep}{2.5pt}
\begin{tabular}{@{}l l l l p{4.0cm} p{4.0cm}@{}}
\toprule
\textbf{Domain} & \textbf{Oracle (Time)} & \textbf{Design Space} & \textbf{Fidelity} & \textbf{De Novo Task} & \textbf{Optimization Task} \\
\midrule
\multicolumn{6}{@{}l}{\textit{Drug Design}} \\[2pt]
ADMET      & RDKit ($<$1\,ms)       & SMILES molecule      & Empirical & ---     & Modify a drug candidate to improve ADMET profile \\
PK/PD      & SciPy ODE (${\sim}$25\,ms) & Dose (mg), freq.\ (hr) & Model & Design a dosing regimen for target plasma concentrations & Adjust dosing when PK metrics are suboptimal \\
Docking    & AutoDock Vina (${\sim}$1.5\,s) & SMILES molecule & Model & Design a ligand for a protein binding pocket & Modify a lead compound to improve binding fit \\
\midrule
\multicolumn{6}{@{}l}{\textit{Biology}} \\[2pt]
FBA        & COBRApy LP (${\sim}$70\,ms) & Gene knockouts       & Model & Engineer \emph{E.~coli} for target metabolite production & Modify a strain to redirect metabolic flux \\
SSA        & Gillespie (${\sim}$130\,ms)  & Network + rates      & Model & Design a gene circuit with target stochastic behavior & Tune circuit parameters for new targets \\
RNA Design & ViennaRNA ($<$1\,ms)         & ACGU sequence        & Model & Design a sequence folding into a target structure & Modify a sequence for a new folding target \\
Perturbation & Linear GRN ($<$1\,ms)      & Gene name            & Empirical & Identify a knockout for desired expression changes & --- \\
\midrule
\multicolumn{6}{@{}l}{\textit{Physics}} \\[2pt]
Quantum    & NumPy ($<$1\,ms)         & Gate sequence        & Exact & Synthesize a circuit preparing a target quantum state & Improve an existing circuit's fidelity \\
Thin Film  & Transfer matrix ($<$1\,ms) & Layer stack (nm)   & Exact & Design an optical coating for target reflectance & Modify a coating for new spectral requirements \\
\midrule
\multicolumn{6}{@{}l}{\textit{Engineering}} \\[2pt]
Controls   & SciPy (${\sim}$25\,ms)   & $K_p, K_i, K_d$     & Exact & Tune a PID controller for a new plant & Retune a controller for tighter specifications \\
Sig.\ Proc. & SciPy ($<$1\,ms)        & Filter params        & Exact & Design a digital filter for target frequency response & Adjust filter cutoff or response characteristics \\
Alloy      & CALPHAD ($<$1\,ms)       & Composition, temp.   & Empirical & Design an alloy for target mechanical properties & Modify composition for new property targets \\
\midrule
\multicolumn{6}{@{}l}{\textit{Chemical Engineering}} \\[2pt]
Reactor    & SciPy CSTR ($<$1\,ms)    & $V$, $T$, $F$, $c_0$ & Model & Design CSTR conditions for target conversion & Adjust operating conditions for new targets \\
Heat Exch. & LMTD ($<$1\,ms)          & Tube geometry        & Model & Design a heat exchanger for target performance & Modify geometry for changed process conditions \\
\bottomrule
\end{tabular}
\end{table*}

\subsection{Oracle Fidelity and Execution Cost}

As shown in \Cref{tab:domains}, our oracles span a fidelity spectrum from \emph{exact} (mathematically precise: Quantum, Controls, Signal Processing, Thin Film) through \emph{model-exact} (simplified scientific model solved exactly: PK/PD, Docking, FBA, SSA, Reactor, Heat Exchanger, RNA Design) to \emph{empirical} (data-derived correlations: ADMET, Alloy, Perturbation). Exact oracles test pure mathematical inversion, model-exact oracles test design under validated but simplified models, and empirical oracles test optimization of data-derived predictors where reward hacking is a known risk~\citep{stanton2022accelerating}. Oracle execution times range from sub-millisecond to $\sim$5--10\,s (Docking); most are fast enough for online RL training.

\subsection{Difficulty Levels}

Each domain defines four difficulty levels:
\begin{itemize}[nosep,leftmargin=*]
  \item \textbf{L1}: Single target metric, simple system configuration, 30\% relative tolerance.
  \item \textbf{L2}: Two target metrics, moderate complexity, 25\% relative tolerance.
  \item \textbf{L3}: Three--four targets, complex systems, 20\% relative tolerance.
  \item \textbf{L4}: All metrics, most complex configurations, 15\% relative tolerance.
\end{itemize}

\subsection{Evaluation Metrics and Protocol}

For each task, we run $K=3$ attempts.\footnote{While $K=3$ is relatively low, evaluating 7 models $\times$ 5 settings $\times$ 520 tasks $\times$ up to 20 turns makes larger $K$ computationally prohibitive.} We report parse rate, validity rate, success rate (all required targets satisfied), and best reward. Domain scores are averaged across tasks; the overall \bench{} score is the mean success rate (\%) across domains. We report both \emph{full-benchmark} aggregates over the official manifest and \emph{shared-core} aggregates over the subset of domains supported by all compared models. We evaluate five settings---1-turn \emph{de novo}, 5-turn feedback, 20-turn long-horizon feedback, 1-turn optimization, and 20-turn long-horizon optimization---that probe distinct capabilities; the full protocol is detailed in \Cref{app:benchmark_details}.

In addition to model-facing evaluation, each official task is accompanied by a deterministic calibration artifact generated from the same manifest-driven harness: a simple baseline hit rate, a random-search reward distribution under a fixed oracle budget, a binary solvability label, and a coarse difficulty bucket. These artifacts are used to check that the public task set is neither trivially easy nor silently impossible under the released oracles.

\subsection{De Novo vs.\ Optimization}
\label{sec:denovo_vs_opt}

Each domain supports two complementary task types. In \emph{de novo} design, the model must construct a valid design from scratch given only target specifications---mirroring greenfield scientific design. In \emph{design optimization}, the model receives an existing design with measured properties and must modify it to achieve new targets---mirroring the common real-world setting of iterating on existing designs. In the optimization settings, success requires both meeting the new targets and changing the starting design. Thirteen of 14 domains include de novo tasks; thirteen include optimization tasks (all except Perturbation; ADMET appears only in optimization form, as generating molecules purely for ADMET properties without a structural starting point is not a realistic workflow).

\FloatBarrier
\section{\method{}: \methodfull}
\label{sec:method}

\subsection{From Forward Oracle to RL Environment}

We call our training recipe \method{}, short for \methodfull. The contribution is a training interface, not a new RL objective: initialize a goal-conditioned policy with supervised fine-tuning, then apply standard GRPO using the benchmark oracle as reward. The key insight is that any forward oracle $\Fwd : \A \to \B$ can be wrapped as a reinforcement learning environment for training an LLM policy. This mirrors the execution-feedback paradigm that has proven successful in code generation~\citep{chen2021codex}, where unit tests serve as reward signals---except here scientific simulators and predictors provide rich, continuous reward signals grounded in domain structure.

\paragraph{Amortized inverse design.} Classical approaches to inverse design (genetic algorithms, Bayesian optimization, bilevel programming) solve each target independently, requiring $N$ expensive optimization runs for $N$ goals. \method{} instead trains a \emph{goal-conditioned policy} that amortizes across a goal distribution: after an offline training phase, many novel goals can be attempted in a single forward pass rather than with fresh per-goal optimization. Rather than paying the expensive oracle cost at inference for every new target, \method{} spends the simulator budget offline, distilling search dynamics into the model's weights for fast, zero-shot deployment.

\paragraph{Reward formulation.} Given a task goal specification $g$ (\Cref{sec:benchmark}) encoding desired targets and constraints, we define a reward function:
\begin{equation}
\reward(\design \mid g) = -\dist\big(\Fwd(\design), g\big) + \lambda_{\text{feas}} \cdot \text{feasibility}(\design) + \lambda_{\text{pars}} \cdot \text{parsimony}(\design),
\label{eq:reward}
\end{equation}
where $\dist$ is a domain-specific distance between the simulated outcome and the targets encoded in $g$ (e.g., relative error for scalar targets, state fidelity for quantum), and the optional bonus terms encourage valid designs (e.g., schema-valid JSON) and parsimonious or minimally changed solutions (e.g., fewer gene knockouts in FBA, edits that remain close to the provided starting design in optimization tasks). In practice, $\lambda_{\text{pars}}$ is set to zero for most domains; the extra terms are included only when they survive task-specific ablations.

The LLM policy $\policy(\design \mid g)$ receives the goal as a natural-language prompt and generates a JSON design. The environment executes $\Fwd(\design)$, computes $\reward$, and returns the scalar reward. This enables standard RL algorithms to train the policy using simulator feedback with no human preference labels and no hand-written expert rewards beyond the task-specific oracle.

\subsection{Training Pipeline}

We employ a two-stage recipe: supervised fine-tuning (SFT) for format compliance and basic patterns, followed by standard Group Relative Policy Optimization (GRPO) for reward maximization.

\paragraph{Stage 1: Supervised Fine-Tuning (SFT).}
We generate training data by sampling goals from each domain's goal distribution, executing candidate designs, and retaining successful or high-reward (goal, design) pairs. We keep these training goals and seeds disjoint from the frozen benchmark tasks. We fine-tune a base LLM using QLoRA~\citep{dettmers2023qlora} on instruction-formatted examples where the model receives a goal description and must produce a JSON design. SFT teaches the model to (1) produce syntactically valid JSON in the expected schema, (2) output values in reasonable ranges, and (3) associate domain terminology with design parameters.

\paragraph{Stage 2: Group Relative Policy Optimization (GRPO).}
GRPO~\citep{shao2024deepseekmath} improves upon PPO~\citep{schulman2017ppo} by eliminating the need for a separate value network. For each goal $g_i$ in a batch, we sample a group of $K$ designs $\{\design_{i,1}, \ldots, \design_{i,K}\}$ from the current policy and compute their rewards $\{r_{i,1}, \ldots, r_{i,K}\}$ using the forward oracle. The group-relative advantage for design $k$ is:
\begin{equation}
\hat{A}_{i,k} = \frac{r_{i,k} - \mu_i}{\sigma_i + \epsilon}, \quad \mu_i = \frac{1}{K}\sum_{k=1}^K r_{i,k}, \quad \sigma_i = \sqrt{\frac{1}{K}\sum_{k=1}^K (r_{i,k} - \mu_i)^2}.
\label{eq:grpo_advantage}
\end{equation}
The policy is updated to maximize:
\begin{equation}
\mathcal{L}_{\text{GRPO}} = \E_{g \sim p(g)} \left[ \frac{1}{K} \sum_{k=1}^K \min\!\left(\rho_k \hat{A}_k,\; \text{clip}(\rho_k, 1\!-\!\epsilon, 1\!+\!\epsilon)\hat{A}_k\right) \right] - \beta \, \KL(\policy \| \pi_{\text{ref}}),
\label{eq:grpo_loss}
\end{equation}
where $\rho_k = \policy(\design_k \mid g) / \pi_{\text{old}}(\design_k \mid g)$ is the importance ratio and $\beta$ controls KL divergence from the SFT checkpoint $\pi_{\text{ref}}$.

\begin{algorithm}[t]
\caption{\method{}: training language models with simulator feedback}
\label{alg:rlsf}
\begin{algorithmic}[1]
\State \textbf{Input:} forward oracle $\Fwd$, goal distribution $p(g)$, base model $\pi_{\text{base}}$
\State \textbf{Stage 1: SFT}
\State Generate $\Dtrain_{\text{SFT}}$ by sampling goals, executing candidate designs, and keeping pairs with reward $> \tau_{\text{SFT}}$
\State $\pi_{\text{ref}} \leftarrow \text{QLoRA-SFT}(\pi_{\text{base}}, \Dtrain_{\text{SFT}})$
\State \textbf{Stage 2: GRPO}
\State Initialize $\policy \leftarrow \pi_{\text{ref}}$
\For{iterations $t = 1, \ldots, T$}
  \State Sample batch $\{g_i\}_{i=1}^M \sim p(g)$
  \For{each $g_i$}
    \State Generate group: $\{\design_{i,k}\}_{k=1}^K \sim \policy(\cdot \mid g_i)$
    \State Execute: $r_{i,k} = \reward(\design_{i,k} \mid g_i)$ using $\Fwd$
    \State Compute group advantages $\hat{A}_{i,k}$ via \Cref{eq:grpo_advantage}
  \EndFor
  \State Update $\policy$ via clipped objective (\Cref{eq:grpo_loss}) with KL penalty
\EndFor
\State \textbf{Output:} Trained policy $\policy$
\end{algorithmic}
\end{algorithm}

\subsection{Practical Considerations}

\paragraph{Efficiency.} QLoRA reduces memory requirements significantly, enabling training on a single H100 GPU. Each GRPO iteration requires $M \times K$ oracle calls; with fast simulators, this takes seconds per batch.

\paragraph{Chain-of-thought reasoning.} During GRPO, we allow the model to generate free-form reasoning before its JSON design. The reward is computed only on the parsed design, but the RL gradient reinforces reasoning chains that lead to high-reward outputs. This mirrors test-time compute scaling~\citep{openai2024o1} but applied to scientific domains: the model learns to reason about parameter relationships (``R$_0 = \beta/\gamma$, so for R$_0 = 2.5$ with $\gamma=0.1$, $\beta \approx 0.25$'') before committing to a design.

\paragraph{Constrained generation.} We use schema validation to filter invalid outputs before oracle execution. Designs that fail JSON parsing or constraint validation receive a fixed penalty reward, providing gradient signal to avoid malformed outputs.

\paragraph{Domain generality.} The pipeline applies to any domain in \bench{} without framework changes. The only domain-specific components are the forward oracle $\Fwd$ and reward function $\reward$, both encapsulated in a standardized environment abstraction. In this paper, however, we make training claims only on selected case-study domains rather than across the entire benchmark. We find that \method{} is most effective when three conditions hold: (1) the executor is \emph{fast and reliable} (seconds per call, deterministic or low-variance), (2) the design space is \emph{structured and symbolic} (JSON parameters, molecular strings, gate sequences) such that LLMs can manipulate them with constraints, and (3) \emph{diverse goals} are naturally available, making amortization across the goal distribution valuable. Under these conditions, \method{} can trade expensive long-horizon test-time search for cheaper one-pass inference at deployment.

\paragraph{Release.} We provide the \bench{} benchmark suite (manifest, tasks, scoring code, and reference oracle implementations) as well as the training environment wrappers and configs used for \method{}, enabling reproducible evaluation and simulator-grounded RL training.

\FloatBarrier
\section{Experiments}
\label{sec:experiments}

\subsection{Frontier LLM Evaluation}

\paragraph{Setup.} We evaluate seven frontier LLMs on the frozen \bench{}~v1 release: GPT-5.2 and GPT-4o~\citep{openai2024gpt4o}, Claude Opus~4.6, Claude Sonnet~4.6, and Claude Sonnet~4.5~\citep{anthropic2025claude}, and Gemini~3.1~Pro and Gemini~2.0~Flash~\citep{google2025gemini}. The benchmark contains 14 unique scientific domains overall, with 13 official \emph{de novo} domains and 13 official optimization domains (ADMET appears only in optimization; Perturbation only in \emph{de novo}), for 520 tasks total. Each task is attempted $K=3$ times with temperature 0.7. Models receive the goal as a natural-language prompt specifying target values, constraints, and the expected JSON schema.

\paragraph{Results.} \Cref{tab:frontier_results} reports per-domain success rates on the 13 official \emph{de novo} domains. The score column uses the 10-domain shared-core subset so that all seven models are compared on the same domain set.

\begin{table}[htbp]
\centering
\caption{Frontier LLM evaluation on the 13 official de novo SciDesignBench~v1 domains. Success rate (\%) $\pm$ standard error per domain (20 tasks each, $K{=}3$ attempts). The score column is the manifest-defined shared-core score over 10 domains, enabling fair comparison across all seven models. \textbf{Bold}: best. \underline{Underline}: second best.}
\label{tab:frontier_results}
\scriptsize
\setlength{\tabcolsep}{2.4pt}
\resizebox{\linewidth}{!}{%
\begin{tabular}{@{}l cc cccc cc ccc cc c@{}}
\toprule
& \multicolumn{2}{c}{\textit{Drug Design}}  & \multicolumn{4}{c}{\textit{Biology}}  & \multicolumn{2}{c}{\textit{Physics}}  & \multicolumn{3}{c}{\textit{Engineering}}  & \multicolumn{2}{c}{\textit{Chemical Engineering}} & \\
\cmidrule(lr){2-3} \cmidrule(lr){4-7} \cmidrule(lr){8-9} \cmidrule(lr){10-12} \cmidrule(lr){13-14}
\textbf{Model} & \textbf{PK/PD} & \textbf{Docking} & \textbf{FBA} & \textbf{SSA} & \textbf{RNA Design} & \textbf{Perturbation} & \textbf{Quantum} & \textbf{Thin Film} & \textbf{Controls} & \textbf{Signal Processing} & \textbf{Alloy} & \textbf{Reactor} & \textbf{Heat Exchanger} & \textbf{Score} \\
\midrule
Sonnet 4.5 & \underline{38.0$_{\pm 8}$} & \underline{53.3$_{\pm 9}$} & 0.0$_{\pm 0}$ & \textbf{20.0$_{\pm 9}$} & 3.3$_{\pm 2}$ & 50.0$_{\pm 10}$ & \textbf{48.3$_{\pm 11}$} & \textbf{33.3$_{\pm 10}$} & 1.7$_{\pm 2}$ & 30.0$_{\pm 11}$ & \textbf{30.0$_{\pm 11}$} & \textbf{45.0$_{\pm 8}$} & 26.7$_{\pm 7}$ & \textbf{29.0} \\
Opus 4.6 & 23.3$_{\pm 9}$ & \textbf{56.7$_{\pm 9}$} & 3.3$_{\pm 2}$ & 13.3$_{\pm 7}$ & 5.0$_{\pm 4}$ & 53.3$_{\pm 11}$ & \underline{33.3$_{\pm 10}$} & 28.3$_{\pm 10}$ & \underline{3.3$_{\pm 3}$} & 33.3$_{\pm 11}$ & \underline{25.0$_{\pm 10}$} & 30.0$_{\pm 10}$ & 20.0$_{\pm 5}$ & 23.8 \\
Sonnet 4.6 & 29.0$_{\pm 9}$ & 46.7$_{\pm 8}$ & 1.7$_{\pm 2}$ & 16.7$_{\pm 8}$ & \underline{20.0$_{\pm 7}$} & \underline{63.3$_{\pm 9}$} & 26.7$_{\pm 10}$ & 28.3$_{\pm 10}$ & 0.0$_{\pm 0}$ & 30.0$_{\pm 11}$ & 0.0$_{\pm 0}$ & 28.3$_{\pm 9}$ & \textbf{33.3$_{\pm 9}$} & 22.9 \\
Gemini 3.1 Pro & \textbf{48.0$_{\pm 9}$} & 31.7$_{\pm 10}$ & 1.7$_{\pm 2}$ & 13.3$_{\pm 7}$ & \textbf{23.3$_{\pm 7}$} & \textbf{70.0$_{\pm 10}$} & 15.0$_{\pm 7}$ & 18.3$_{\pm 8}$ & \textbf{8.3$_{\pm 6}$} & 31.7$_{\pm 10}$ & 23.3$_{\pm 9}$ & 0.0$_{\pm 0}$ & \underline{28.3$_{\pm 8}$} & 24.0 \\
GPT-5.2 & \underline{38.0$_{\pm 10}$} & --- & \underline{18.3$_{\pm 5}$} & \underline{18.3$_{\pm 7}$} & --- & 55.0$_{\pm 10}$ & \underline{33.3$_{\pm 10}$} & --- & 0.0$_{\pm 0}$ & 30.0$_{\pm 11}$ & 13.3$_{\pm 6}$ & \underline{31.7$_{\pm 10}$} & 18.3$_{\pm 8}$ & \underline{25.6} \\
Gemini 2.0 Flash & 15.0$_{\pm 7}$ & 10.0$_{\pm 5}$ & 0.0$_{\pm 0}$ & \textbf{20.0$_{\pm 9}$} & 0.0$_{\pm 0}$ & 33.3$_{\pm 10}$ & 23.3$_{\pm 9}$ & \underline{31.7$_{\pm 10}$} & 0.0$_{\pm 0}$ & \textbf{40.0$_{\pm 11}$} & 20.0$_{\pm 7}$ & 20.0$_{\pm 9}$ & 23.3$_{\pm 8}$ & 19.5 \\
GPT-4o & 0.0$_{\pm 0}$ & --- & \textbf{20.0$_{\pm 6}$} & \underline{18.3$_{\pm 9}$} & --- & 16.7$_{\pm 6}$ & 0.0$_{\pm 0}$ & --- & 0.0$_{\pm 0}$ & \underline{35.0$_{\pm 10}$} & 11.7$_{\pm 6}$ & 18.3$_{\pm 7}$ & 8.3$_{\pm 3}$ & 12.8 \\
\bottomrule
\end{tabular}
}
\end{table}

Frontier models remain far from saturating the benchmark. The strongest shared-core one-turn model, Sonnet~4.5, reaches only 29.0\% success. The benchmark is genuinely heterogeneous: no single model family dominates. Gemini~3.1~Pro is strongest on PK/PD and RNA Design, Opus~4.6 leads Docking, GPT-4o leads FBA, and Sonnet~4.5 leads the shared-core aggregate. Perturbation, Reactor, Heat Exchanger, Docking, and PK/PD are among the easiest \emph{de novo} domains, while FBA remains challenging and strongly model-dependent. Zero-shot SSA scores remain low and highly model-dependent. Across most domains, parse rates are substantially higher than success rates, so failures are predominantly \emph{scientific} rather than syntactic (\Cref{fig:parse_vs_success_app}). Per-domain analysis is in \Cref{app:benchmark_details}.

\subsection{Long-Horizon Evaluation with Simulator Feedback}

We evaluate two feedback settings: \textbf{5-turn feedback} (3 independent attempts, each with up to 5 rounds of simulate-revise) and \textbf{20-turn long-horizon feedback} (a single attempt with up to 20 rounds). After each turn, models receive structured feedback showing which targets were met, which were missed, and the relative error for each metric.

\paragraph{Results.} \Cref{tab:multiturn_results} summarizes the three \emph{de novo} settings on the 10 shared-core domains.

\begin{table}[htbp]
\centering
\caption{Shared-core comparison across the three de novo benchmark settings. All scores are computed over the 10 manifest-defined shared-core de novo domains. \textbf{Bold}: best per column. \underline{Underline}: second best.}
\label{tab:multiturn_results}
\scriptsize
\setlength{\tabcolsep}{4pt}
\begin{tabular}{@{}l ccc cc@{}}
\toprule
\textbf{Model} & \textbf{1-turn} & \textbf{5-turn} & \textbf{20-turn} & \textbf{$\Delta$ (1$\to$5)} & \textbf{$\Delta$ (5$\to$20)} \\
\midrule
Sonnet 4.5 & \textbf{29.0} & \textbf{66.5} & 69.5 & 37.5 & 3.0 \\
Opus 4.6 & 23.8 & \underline{65.5} & \textbf{76.0} & \underline{41.7} & \underline{10.5} \\
Sonnet 4.6 & 22.9 & \textbf{66.5} & \underline{71.5} & \textbf{43.6} & 5.0 \\
Gemini 3.1 Pro & 24.0 & 57.3 & 60.5 & 33.4 & 3.2 \\
GPT-5.2 & \underline{25.6} & 62.5 & 67.0 & 36.9 & 4.5 \\
Gemini 2.0 Flash & 19.5 & 43.2 & 60.5 & 23.7 & \textbf{17.3} \\
GPT-4o & 12.8 & 53.0 & 55.0 & 40.2 & 2.0 \\
\bottomrule
\end{tabular}
\end{table}

\begin{figure*}[htbp]
  \centering
  \maybeincludegraphics[width=\linewidth]{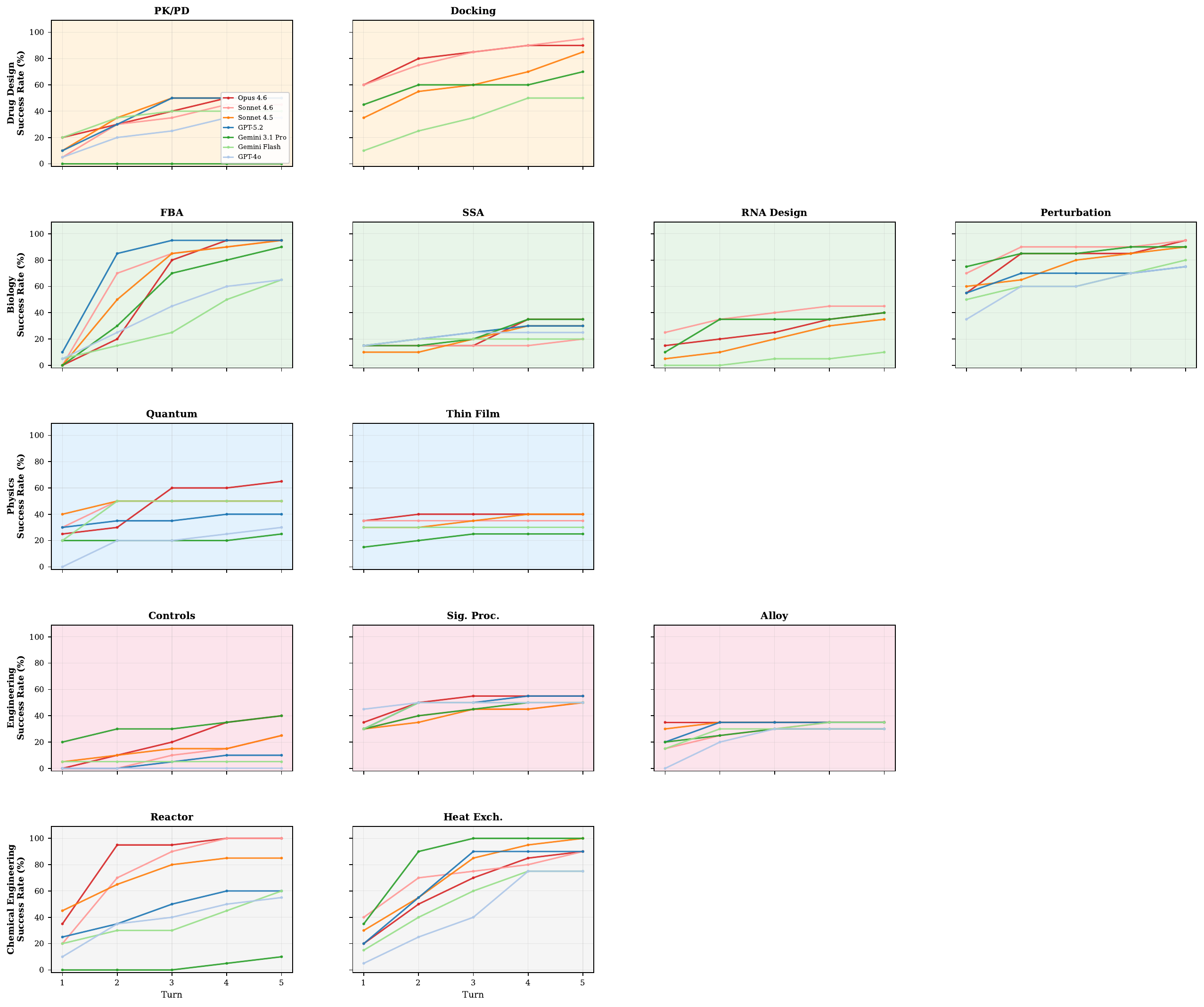}
  \caption{Success rate vs.\ turn number in the 20-turn long-horizon setting, per domain. Each line represents one model. Perturbation converges quickly; Controls and SSA show sustained gains through 20 turns; domains such as FBA remain highly model-dependent even with longer trajectories.}
  \label{fig:turn_curves_domain}
\end{figure*}

\begin{figure}[htbp]
  \centering
  \maybeincludegraphics[width=0.65\linewidth]{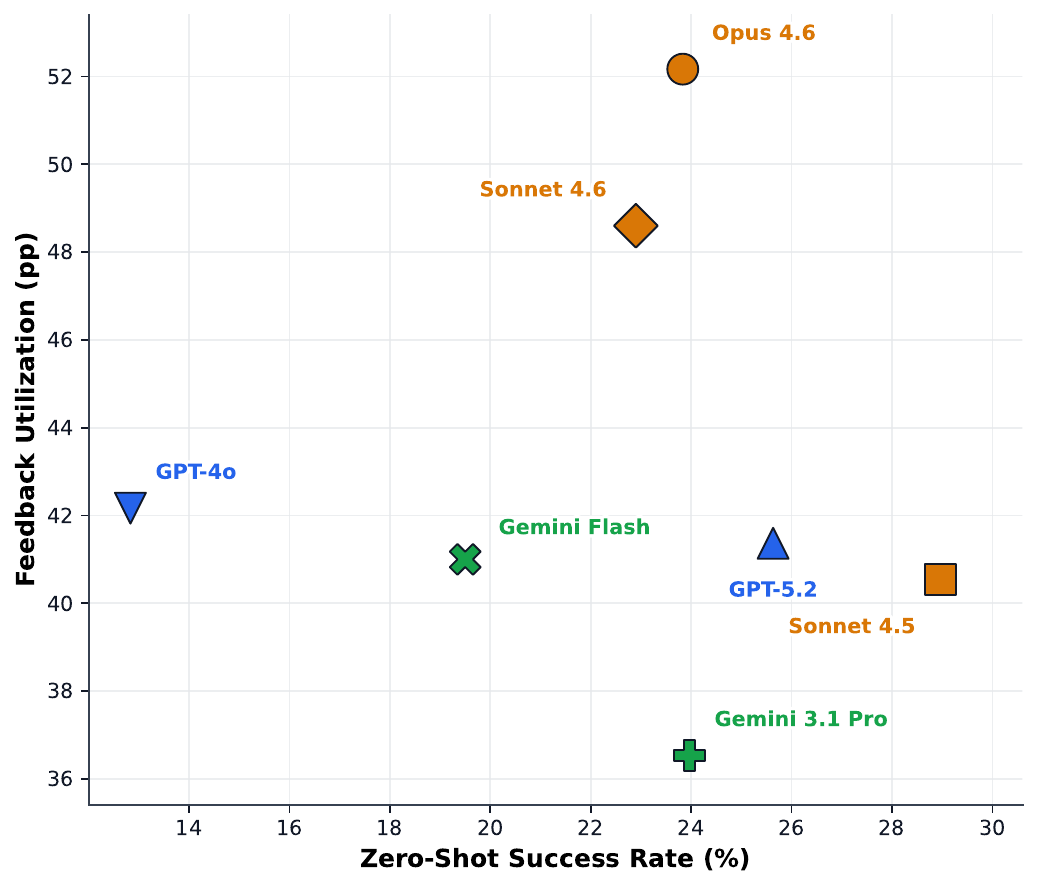}
  \caption{\textbf{Feedback utilization is a distinct capability from static knowledge.} Zero-shot success rate (x-axis) vs.\ feedback utilization (20-turn minus zero-shot, y-axis) for each model. Models with similar one-turn performance can have vastly different long-horizon gains---e.g., Sonnet~4.6 and Gemini~3.1~Pro start within a few points but Sonnet~4.6 realizes a much larger improvement under iterative simulator feedback.}
  \label{fig:feedback_scatter}
\end{figure}

The central empirical finding is that the ability to utilize long-horizon scientific feedback is orthogonal to zero-shot competence. Every model improves in the 5-turn setting, confirming that simulator feedback is valuable. But the leaderboard also changes with horizon. Sonnet~4.5 is strongest at 1 turn (29.0), while Opus~4.6 is strongest after 20 turns (76.0), indicating that single-shot design and long-horizon refinement are not the same skill. While the 20-turn protocol is uniformly better than the 5-turn protocol in aggregate, this masks important domain-level tradeoffs: on some domains a single long trajectory outperforms multiple short attempts, while on others the lack of independent restarts traps the model in a failing strategy. Finally, models with similar one-turn scores can diverge sharply once feedback enters the loop: Sonnet~4.6 and Gemini~3.1~Pro start within a few shared-core points but separate materially by 20 turns (\Cref{fig:feedback_scatter}). Per-domain turn curves are in \Cref{fig:turn_curves_domain} and \Cref{tab:app_domain_comparison}.

\begin{figure}[htbp]
  \centering
  \maybeincludegraphics[width=0.7\linewidth]{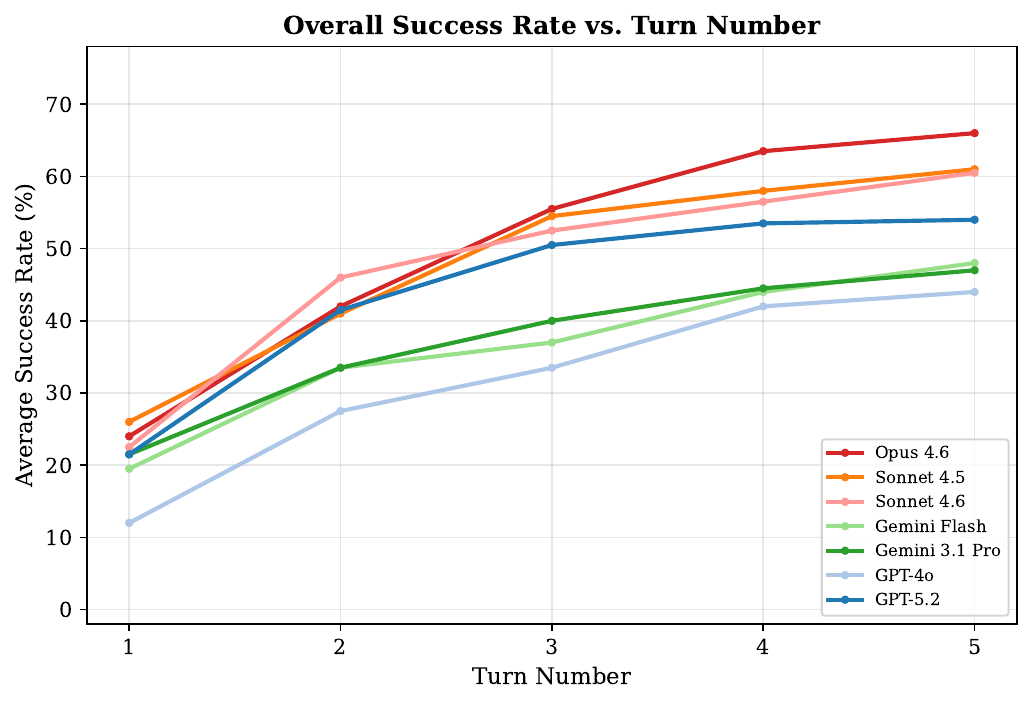}
  \caption{Aggregate success rate vs.\ turn number in the 20-turn long-horizon setting, averaged across the 10 manifest-defined \emph{de novo} shared-core domains. Simulator feedback provides rapid initial gains (turns 1--5) but exhibits diminishing returns thereafter. Even at 20 turns, models plateau well below full benchmark saturation.}
  \label{fig:multiturn}
\end{figure}

\subsection{Design Optimization Evaluation}
\label{sec:optimization}

\paragraph{Setup.} For each of thirteen optimization domains (ADMET plus the twelve \emph{de novo} domains that support modification, excluding Perturbation), we generate 20 optimization tasks (5 per difficulty level). Each task provides a starting design, its measured properties, and new target values. Evaluation is zero-shot with $K{=}3$ attempts. We additionally evaluate \textbf{20-turn long-horizon optimization} (starting design plus iterative feedback). Summary scores use the 10-domain optimization shared-core subset.

\paragraph{Results.} \Cref{tab:optimization_results} presents shared-core scores across all five settings.

\begin{table}[htbp]
\centering
\caption{Evaluation across five settings. The first three columns use the 10 de novo shared-core domains; optimization columns use the 10 optimization shared-core domains. \textbf{Bold}: best per column. \underline{Underline}: second best.}
\label{tab:optimization_results}
\scriptsize
\setlength{\tabcolsep}{3pt}
\begin{tabular}{@{}l ccc cc@{}}
\toprule
\textbf{Model} & \textbf{1-turn} & \textbf{5-turn} & \textbf{20-turn} & \textbf{Opt.\ (1-turn)} & \textbf{Opt.\ (20-turn)} \\
\midrule
Sonnet 4.5 & \textbf{29.0} & \textbf{66.5} & 69.5 & 30.5 & 61.5 \\
Opus 4.6 & 23.8 & \underline{65.5} & \textbf{76.0} & \textbf{34.5} & \textbf{67.5} \\
Sonnet 4.6 & 22.9 & \textbf{66.5} & \underline{71.5} & 31.2 & \underline{63.0} \\
Gemini 3.1 Pro & 24.0 & 57.3 & 60.5 & \underline{31.3} & 56.0 \\
GPT-5.2 & \underline{25.6} & 62.5 & 67.0 & 29.0 & 58.5 \\
Gemini 2.0 Flash & 19.5 & 43.2 & 60.5 & 22.2 & 48.0 \\
GPT-4o & 12.8 & 53.0 & 55.0 & 17.8 & 41.5 \\
\bottomrule
\end{tabular}
\end{table}

Model rankings shift between \emph{de novo} design and optimization, confirming that these settings probe different capabilities. In one-turn optimization, Opus~4.6 is strongest on the 10-domain shared-core subset, even though Sonnet~4.5 leads one-turn \emph{de novo} design. This distinction matters: real-world lead optimization requires constrained modification of a seed design, a task that heavily penalizes models prone to destructive edits or mode collapse. Across domains, behavior is highly heterogeneous. Some tasks reward local refinement around a seed design, while others are easier when models can effectively ``restart'' away from the provided seed. Detailed per-domain comparisons are in \Cref{tab:app_optimization_comparison}.

\paragraph{Long-horizon design optimization.} \Cref{tab:agentic_optimization} compares 20-turn \emph{de novo} and 20-turn optimization on the 9 domains shared by both shared-core domain sets.

\begin{table}[htbp]
\centering
\caption{Comparison of 20-turn de novo vs.\ 20-turn agentic optimization on the 9 domains shared by both manifest-defined shared-core sets. $\Delta$: optimization minus de novo. \textbf{Bold}: best per column. \underline{Underline}: second best.}
\label{tab:agentic_optimization}
\scriptsize
\setlength{\tabcolsep}{4pt}
\begin{tabular}{@{}l cc c@{}}
\toprule
\textbf{Model} & \textbf{20-turn (de novo)} & \textbf{20-turn (opt.)} & \textbf{$\Delta$} \\
\midrule
Sonnet 4.5 & 66.1 & 62.2 & -3.9 \\
Opus 4.6 & \textbf{73.3} & \textbf{68.3} & -5.0 \\
Sonnet 4.6 & \underline{68.3} & \underline{64.4} & \underline{-3.9} \\
Gemini 3.1 Pro & 56.1 & 55.0 & \textbf{-1.1} \\
GPT-5.2 & 63.9 & 59.4 & -4.4 \\
Gemini 2.0 Flash & 57.8 & 49.4 & -8.3 \\
GPT-4o & 52.2 & 41.7 & -10.6 \\
\bottomrule
\end{tabular}
\end{table}

Surprisingly, providing a starting design actually degrades aggregate performance across all seven models (\Cref{tab:agentic_optimization}). While seed designs accelerate progress in specific domains, they frequently trap models in poor local optima, confirming that constrained modification is harder for current LLMs than unconstrained search. This makes optimization a useful complement to \emph{de novo} design rather than a redundant variant. Detailed per-domain results are in \Cref{tab:app_agentic_optimization_domains}.

\subsection{\method{} Training Results}
\label{sec:admet_expert_iteration}

\paragraph{Setup.} We apply the \method{} recipe (\Cref{sec:method}) to Qwen3-8B~\citep{qwen2025qwen3} using QLoRA. These are proof-of-concept case studies; we do not claim that the same recipe already works uniformly across all 14 domains. We focus first on \textbf{ADMET drug design optimization}---the scientifically relevant setting of modifying an existing molecule to meet new property targets---where the strongest frontier baseline reaches 40\% success on the domain-specific optimization benchmark. The molecular design space (SMILES strings) is also well represented in the base model's pretraining corpus. The ADMET oracle evaluates molecular properties (logP, molecular weight, TPSA, QED, ring count, hydrogen bond donors/acceptors) via RDKit~\citep{rdkit}, returning continuous scores that enable rich reward shaping.

\paragraph{Curriculum GRPO pipeline.} We apply \emph{expert iteration}~\citep{anthony2017thinking}: (1)~generate 6{,}428 expert optimization traces via random modification search (Tanimoto similarity $\geq 0.3$ to the starting molecule), each verified against the oracle and paired with synthetic chain-of-thought reasoning; (2)~fine-tune Qwen3-8B via QLoRA ($r{=}64$, $\alpha{=}128$) on these domain-specific traces; (3)~apply curriculum GRPO, progressively introducing harder tasks across three phases: Phase~1 trains on L1--L2 tasks only (250 steps), Phase~2 adds L3 (250 steps), and Phase~3 trains on the full L1--L4 distribution (300 steps), for 800 cumulative steps on 4$\times$H100. We use $K{=}32$, constant learning rate $10^{-5}$, $\beta{=}0.01$, and a per-target partial credit reward ($+1.0$ per individual target met) in addition to the success bonus ($+5.0$ for meeting all targets). All training targets and seeds are strictly disjoint from the evaluation set.

\paragraph{Results.} \Cref{fig:overview_training} shows the ADMET optimization training trajectory. The SFT baseline achieves 30\% optimization success; after curriculum GRPO, success rises to \textbf{41\% at step 700}---a +37\% relative improvement that narrowly exceeds the strongest frontier ADMET optimization benchmark result (40\%). Per-difficulty gains confirm that GRPO improves across all levels: L1 improves from 58\%$\to$72\%, L2 from 32\%$\to$44\%, L3 from 22\%$\to$36\%, and L4 from 6\%$\to$12\%.

\begin{figure}[htbp]
\centering
\maybeincludegraphics[width=\linewidth]{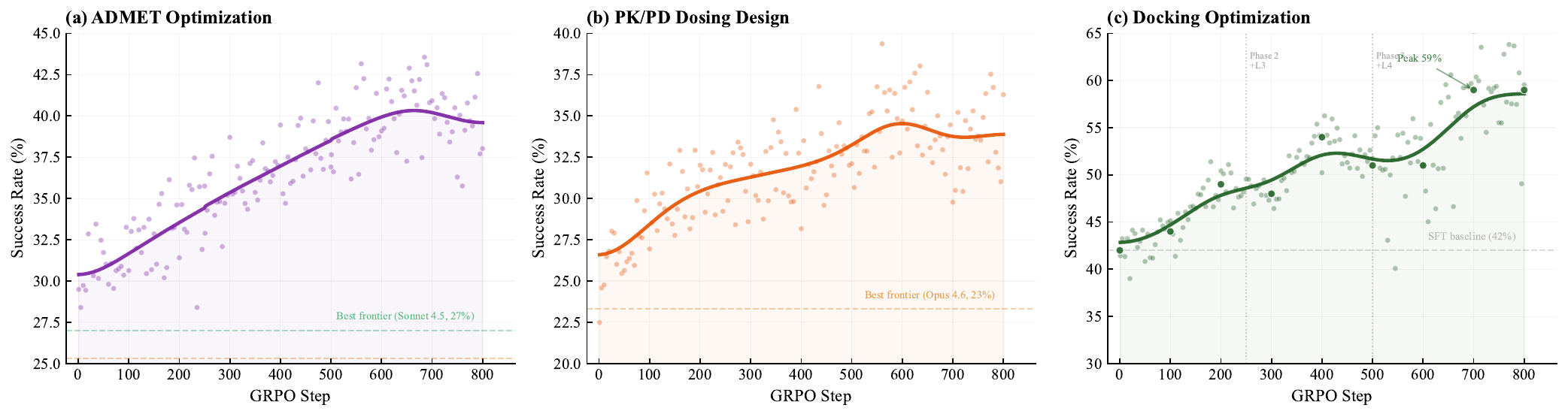}
\vspace{-8pt}
\caption{\textbf{\method{} training across three scientific design tasks.} (a)~ADMET \emph{optimization}: curriculum GRPO improves from 30\% (SFT) to 41\% (GRPO), narrowly exceeding the strongest frontier ADMET optimization benchmark result. (b)~PK/PD: success improves from 24\% to 36\% in \emph{de novo} design and from 32\% to 47\% in optimization, yielding a competitive optimization result on a coupled dosing-control task. (c)~Docking \emph{optimization}: real AutoDock Vina improves from 42\% (SFT) to 59\% (GRPO) in the artifact-backed run used in the paper. Dashed vertical lines mark curriculum phase transitions. Points show raw eval; curves are smoothed.}
\label{fig:rlsf_combined}
\end{figure}

\paragraph{Cross-domain generalization: PK/PD.} We apply the same \method{} recipe to \textbf{PK/PD dosing design}, where the model designs a dosing regimen---dose amount (mg) and dosing frequency (hours)---targeting specific plasma concentration metrics ($C_\text{max}$, AUC, $C_\text{min}$). We generate 6{,}653 expert traces, fine-tune via QLoRA, then apply GRPO ($K{=}32$, 800 steps, 8$\times$A100). The SFT baseline achieves 24\% \emph{de novo} success; after GRPO, success rises to \textbf{36\%} at step 600 (\Cref{fig:rlsf_combined}b). That does not surpass the strongest frontier \emph{de novo} PK/PD baseline, but optimization success improves from 32\% to \textbf{47\%}, a competitive result on a domain where dosing design requires simultaneous reasoning about multiple coupled PK targets. The model reasons about PK relationships ($t_{1/2} = 0.693 \times V / CL$, trough concentration via dosing intervals) and hits multiple targets simultaneously (example traces in \Cref{sec:pkpd_traces}).

\paragraph{Cross-domain generalization: Molecular Docking.} We further extend \method{} to \textbf{structure-based drug design}, where the model optimizes a seed molecule for binding affinity against a protein target. Unlike ADMET and PK/PD, docking uses \emph{AutoDock Vina}~\citep{eberhardt2021autodock}---a physics-based molecular docking engine ($\sim$5--10s per evaluation), the most expensive oracle in our benchmark. We train on docking optimization tasks across 5 protein targets using a curriculum strategy with a continuous binding-affinity bonus. In the artifact-backed run used in the paper, the SFT baseline achieves 42\%; after GRPO, success rises to \textbf{59\%} (\Cref{fig:rlsf_combined}c). L1 improves from 60\%$\to$88\%, L2 from 72\%$\to$92\%, L3 from 36\%$\to$56\%. L4 tasks remain unsolved (0\%). This confirms that simulator-feedback training transfers to a physics-based oracle with non-trivial computational cost, although the resulting model still falls short of the strongest frontier docking baseline. Example reasoning traces are in \Cref{sec:docking_traces}.

Training ablations (learning rate schedule, group size, expert vs.\ multi-domain SFT, anti-memorization safeguards) are presented in \Cref{app:ablations}.

\FloatBarrier
\section{Discussion}
\label{sec:discussion}

\paragraph{Why do LLMs fail at inverse design?}
Parse rates consistently dwarf success rates: frontier models have mastered the syntactic contract of scientific design but frequently fail at the quantitative reasoning required to solve it. Inverse design demands more than format compliance. It requires \emph{quantitative precision} (hitting a target $C_\text{max}$ within 10\% tolerance, not just knowing ``higher dose increases $C_\text{max}$''), \emph{multi-constraint satisfaction} (the intersection of feasible regions for 3--4 simultaneous targets may be small and non-obvious), and \emph{simulator-grounded reasoning} (the correct answer depends on the specific simulator's equations, not just textbook knowledge). ADMET stands out as partially solvable because molecular property prediction relies on well-known heuristics (Lipinski's rules, logP ranges) extensively documented in training data---suggesting LLMs can invert forward mappings when approximate rules of thumb exist.

\paragraph{Long-horizon ability is a separate axis.}
The clearest benchmark result is that single-shot and long-horizon rankings diverge. A model can start with strong one-turn performance yet underuse feedback, while another can begin lower but improve far more under iterative refinement. A benchmark for scientific agents should therefore not collapse everything into a single one-turn score. Long-horizon scientific work is common in practice---researchers propose a design, inspect measurements, revise hypotheses, and iterate---and \bench{} exposes that loop directly.

\paragraph{Feedback helps but saturates.}
Simulator feedback roughly doubles performance, and some easier domains approach saturation under iterative refinement. However, returns diminish sharply beyond 5 turns, and even the domains that improve most remain far from full benchmark saturation. The key limitation is that LLM-driven in-context refinement is pathologically local: models make marginal parameter adjustments and struggle to abandon failing trajectories for fundamentally different design strategies. Furthermore, each task requires up to 20 simulator calls at inference---prohibitive for expensive oracles (molecular dynamics, protein folding) and for large-scale deployment.

\paragraph{The case for simulator-grounded training.}
Our case studies show that an 8B model trained with simulator feedback improves on selected domains spanning the oracle fidelity spectrum: ADMET (empirical predictors, 30\%$\to$41\%), PK/PD (analytical ODE models, 24\%$\to$36\%), and docking (physics-based AutoDock Vina, 42\%$\to$59\%). Training with simulator feedback differs from test-time tool use in two ways: the model internalizes simulator regularities into its weights, and the offline training cost is amortized across many future goals.

\paragraph{Forward oracles as RL environments.}
The scientific simulation ecosystem is a vast, largely untapped reservoir of high-fidelity RL training environments. Unlike code generation (binary unit tests) or math (exact answer checking), scientific simulators provide dense, continuous reward landscapes grounded in physical reality. Decades of investment by domain scientists have produced reusable oracles across virtually every field; \method{} repurposes this infrastructure for LLM training. A trained policy can attempt designs in a single forward pass, while even the best long-horizon benchmark configuration requires repeated oracle calls and still plateaus well below full success. Population-based approaches like AlphaEvolve~\citep{novikov2025alphaevolve} multiply this cost further. The efficiency gap widens with oracle cost, making amortized training especially attractive when practitioners routinely explore many related targets.

\paragraph{Oracle fidelity and reward hacking.}
Our oracles span a fidelity spectrum from mathematically exact to empirical (\Cref{tab:domains}). For exact and model-exact oracles, the reward signal is trustworthy. For empirical oracles, reward hacking---where RL exploits predictor weaknesses---is a concern. We did not observe reward hacking in our GRPO training---the model produces chemically valid, drug-like molecules---but empirical predictors remain susceptible to exploitation. Future work should incorporate safeguards such as multi-predictor ensembles and novelty constraints.

\paragraph{Limitations.}
\begin{itemize}[nosep,leftmargin=*]
  \item \textbf{Simulator fidelity}: We use simplified simulators; results may not transfer to higher-fidelity oracles.
  \item \textbf{Domain coverage}: 14 domains is broader than prior work, but many important domains (climate, protein design, electronic circuits) are not yet included.
  \item \textbf{Training scale}: A single 8B model trained independently on three domains; scaling to larger models and joint multi-domain training may reveal different dynamics.
  \item \textbf{Turn budget}: A fixed 20-turn long-horizon budget; more sophisticated search strategies could improve performance at additional cost.
  \item \textbf{Optimization baselines}: Systematic comparison against classical solvers (Bayesian optimization, GRAPE, etc.) across all 14 domains is future work.
  \item \textbf{API reproducibility}: Frontier model APIs evolve; we provide prompts and evaluation code, but numbers may drift.
  \item \textbf{Hyperparameter sensitivity}: All models evaluated at temperature 0.7; per-model tuning could alter rankings.
\end{itemize}

\FloatBarrier
\section{Conclusion}
\label{sec:conclusion}

We introduced \bench{}, a benchmark of 520 simulator-grounded tasks across 14 scientific domains for evaluating language models on inverse design. The central finding is not merely that zero-shot performance is low (best shared-core score: 29.0\%), but that long-horizon scientific refinement is a distinct capability. The leaderboard reshuffles when models interact iteratively with a simulator; optimization reshuffles it again, reinforcing that design from scratch and constrained modification are different problems. We then showed, via three case studies, that the same oracle interface can be reused for training: a Qwen3-8B trained with \method{} improves on ADMET (30\%$\to$41\%), PK/PD (24\%$\to$36\%), and docking (42\%$\to$59\%) without requiring simulator calls at inference.

\paragraph{Future work.}
\begin{itemize}[nosep,leftmargin=*]
  \item \textbf{Multi-domain scaling}: Training across all 14 domains to study cross-domain transfer and scaling to larger models.
  \item \textbf{Hybrid strategies}: Combining trained \method{} models with test-time simulator access---using the trained model as a strong prior for iterative refinement.
  \item \textbf{Real-world validation}: Connecting \bench{} environments to experimental pipelines to validate that improved simulator performance translates to real-world outcomes.
\end{itemize}

The benchmark tasks, evaluation protocol, and reference oracle implementations are provided to enable standardized evaluation of scientific design agents, along with training environment wrappers for simulator-grounded RL. We hope \bench{} helps reframe scientific-agent evaluation around a concrete question with real scientific stakes: not ``can a model talk about science?'' but ``can it solve inverse problems that working scientists actually care about?''

\bibliographystyle{plainnat}

\appendix
\section{Benchmark Details}
\label{app:benchmark_details}

\subsection{Task Specifications}

Each domain has 20 tasks: 5 tasks per difficulty level (L1--L4). Tasks are generated programmatically using domain-specific goal samplers that ensure:
\begin{itemize}[nosep,leftmargin=*]
  \item Target values are physically meaningful and achievable.
  \item Difficulty increases through more targets, tighter tolerances, and more complex system configurations.
  \item Tasks within a level vary in the specific targets and system parameters.
\end{itemize}

For the released benchmark, task generation and calibration share the same frozen manifest. For every official task we also generate a calibration artifact containing a deterministic baseline sweep, a fixed-budget random-search reward distribution, a solvability label, and an assigned difficulty bucket. Optimization-task calibration uses the same success semantics as evaluation: the proposed design must satisfy the new goal \emph{and} differ from the provided starting design.

\subsection{Domain-Specific Details}

% --- Biology ---
\paragraph{FBA.} We use the \emph{E.~coli} core model~\citep{orth2010fba} with 95 reactions and 72 metabolites. Designs specify gene knockouts (0--5 reactions) and optional flux bounds. The forward oracle solves the LP using COBRApy~\citep{ebrahim2013cobrapy} with parsimonious FBA to compute steady-state fluxes.

\paragraph{SSA.} Designs specify a stochastic mass-action reaction network: species, reactions with stoichiometry, and rate constants. The public benchmark tasks describe only the mass-action network to be proposed, not a named motif library. The oracle runs repeated Gillespie simulations and computes summary statistics such as means, variances, and simple oscillation proxies.

\paragraph{RNA Design.} Designs specify an RNA nucleotide sequence (A, U, G, C). The oracle folds the sequence using ViennaRNA's minimum free energy (MFE) algorithm~\citep{lorenz2011viennarna} and computes structural similarity to the target secondary structure in dot-bracket notation. Tasks require designing sequences that fold into specific hairpin, stem-loop, and multi-loop structures.

\paragraph{Perturbation.} Designs specify a single gene to knock out from a set of 51 genes organized across 7 biological pathways (cell cycle, apoptosis, MAPK/PI3K signaling, transcription factors, DNA repair, epigenetic regulators, metabolism). The oracle propagates the knockout effect through a linear gene regulatory network (GRN) and returns log$_2$ fold-changes for the marker genes specified in the goal. The GRN encodes $\sim$100 known regulatory edges (e.g., TP53 activates CDKN1A, KRAS activates BRAF$\to$MEK1$\to$ERK2) with calibrated weights and small noise to prevent trivial inversion.

% --- Drug Design ---
\paragraph{ADMET.} Designs are molecules in SMILES format. The oracle computes RDKit descriptors (molecular weight, logP, HBD, HBA, TPSA, rotatable bonds) and checks Lipinski's Rule of Five compliance and drug-likeness scores. In the benchmark, ADMET appears only in the \emph{optimization} setting: the model receives a starting drug molecule and must modify it to satisfy new ADMET property targets---the realistic medicinal chemistry workflow of lead optimization. Generating molecules purely for ADMET properties without a structural starting point is not a meaningful task (the model would just propose arbitrary drug-like molecules), so \emph{de novo} ADMET tasks are not included. ADMET \emph{de novo} is used in the \method{} training experiments (\Cref{sec:admet_expert_iteration}) as it provides a well-defined reward signal for RL.

\paragraph{PK/PD.} Designs specify a dosing regimen (dose amount and dosing frequency) for a given drug with fixed pharmacokinetic properties (absorption rate, bioavailability, clearance, volume, compartment model). The oracle integrates ODEs using SciPy to compute C\textsubscript{max}, AUC, half-life, and trough concentration.

\paragraph{Docking.} Designs are drug-like molecules in SMILES format. The oracle docks them into protein binding pockets using AutoDock Vina~\citep{trott2010vina} with five real PDB receptor structures: EGFR (1M17), HIV-1 protease (2BPX), $\beta_2$-adrenergic receptor (2RH1), estrogen receptor-$\alpha$ (1ERE), and acetylcholinesterase (1EVE). Tasks require designing molecules with target binding affinity (kcal/mol) and drug-likeness constraints.

% --- Physics ---
\paragraph{Quantum.} Designs specify a sequence of quantum gates (H, X, Y, Z, S, T, CNOT, CZ, SWAP, RX, RY, RZ with angles) applied to an initial $|0\rangle^{\otimes n}$ state. The oracle simulates statevector evolution and computes fidelity with the target state, basis measurement probabilities, and entanglement entropy.

\paragraph{Thin Film.} Designs specify a stack of optical thin-film layers (material, thickness in nm) for anti-reflection or spectral-filtering coatings. The oracle computes the reflectance/transmittance spectrum using the transfer matrix method (exact solution of Maxwell's equations for planar layered media). Tasks require achieving target reflectance at specific wavelengths or over spectral bands.

% --- Chemical Engineering ---
\paragraph{Reactor.} Designs specify CSTR operating conditions (volume, flow rate, feed concentration, temperature, coolant temperature) for series (A$\to$B$\to$C) or parallel (A$\to$B, A$\to$C) reaction systems with Arrhenius kinetics. The oracle computes steady-state concentrations analytically and returns conversion, selectivity, yield, and productivity.

\paragraph{Heat Exchanger.} Designs specify shell-and-tube heat exchanger geometry (number of tubes, tube length, tube diameters, baffle spacing, number of passes). The oracle computes heat transfer using Dittus--Boelter (tube-side) and Kern (shell-side) correlations with the LMTD method and NTU-effectiveness, returning heat duty, outlet temperatures, effectiveness, and LMTD.

% --- Engineering ---
\paragraph{Controls.} Designs specify PID gains ($K_p$, $K_i$, $K_d$) and an optional derivative filter coefficient for a given plant transfer function. The oracle computes the closed-loop step response using SciPy's signal processing and extracts overshoot, settling time, rise time, and steady-state error.

\paragraph{Signal Processing.} Designs specify a digital filter via its type (Butterworth, Chebyshev Type I/II, or elliptic), cutoff frequency or frequencies (for bandpass/bandstop), and optional ripple/attenuation parameters. The oracle designs the IIR filter using SciPy, computes its frequency response via \texttt{freqz}, and extracts $-$3\,dB cutoff frequency, stopband attenuation, passband ripple, transition bandwidth, and group delay variation.

\paragraph{Alloy.} Designs specify element compositions (fractions that must sum to 1.0) and processing temperature. The oracle uses CALPHAD-inspired mixing rules to compute yield strength, density, melting point, and estimated cost from elemental properties.

\subsection{Prompt Template}

All models receive a standardized prompt containing the domain name, goal description with quantitative targets, applicable constraints, and the expected JSON schema. The prompt instructs the model to respond with only a JSON object matching the schema.

\subsection{Evaluation Protocol}

\begin{itemize}[nosep,leftmargin=*]
  \item Temperature: 0.7 for all models.
  \item Max tokens: 2048.
  \item Attempts per task: $K=3$.
  \item Parse: Extract JSON from model response (handles markdown code blocks).
  \item Validate: Check all constraints (parameter ranges, required fields).
  \item Execute: Run forward oracle on validated design.
  \item Score: Compute domain-specific reward; check success criteria.
  \item Optimization settings: success additionally requires changing the provided starting design.
\end{itemize}

\subsection{Multi-Turn Evaluation Protocol}

In the feedback and long-horizon settings, models receive structured feedback after each design attempt. The feedback format includes:
\begin{itemize}[nosep,leftmargin=*]
  \item Per-target comparison: target value, achieved value, relative error, and PASS/MISS label (20\% relative tolerance for exact targets, 10\% slack for min/max bounds).
  \item Overall summary: reward, count of targets met, and binary success.
  \item Instruction to revise the design, maintaining the same JSON format.
\end{itemize}
The system prompt and original goal remain in context across all turns, preserving the full conversation history. Parse or validation failures trigger a retry message rather than simulator execution. Successful designs (all targets met) trigger early stopping within that attempt.

For the 5-turn setting, we run $K{=}3$ independent attempts, each with up to 5 turns. For the 20-turn long-horizon setting, we run $K{=}1$ attempt with up to 20 turns. In both cases, task success is the best outcome across all attempts and turns.

\subsection{Per-Domain Comparison Across Evaluation Modes}

\Cref{tab:app_domain_comparison} shows how each domain responds to simulator feedback across the three \emph{de novo} evaluation modes on the 10 manifest-defined shared-core domains. \Cref{tab:app_optimization_comparison} extends the analysis by comparing \emph{de novo} zero-shot and design optimization side by side on the 12 benchmark domains shared by both settings (ADMET is optimization-only; Perturbation is \emph{de novo}-only).

\begin{table}[p]
\centering
\caption{Per-domain success rate (\%) across the three de novo evaluation modes for the 10 manifest-defined shared-core domains. Each row is one model--domain combination; $\Delta$ columns show absolute improvement.}
\label{tab:app_domain_comparison}
\tiny
\setlength{\tabcolsep}{2.5pt}
\renewcommand{\arraystretch}{0.85}
\begin{tabular}{@{}ll ccc cc@{}}
\toprule
\textbf{Domain} & \textbf{Model} & \textbf{1-turn} & \textbf{5-turn} & \textbf{20-turn} & \textbf{$\Delta$(1$\to$5)} & \textbf{$\Delta$(5$\to$20)} \\
\midrule
\multicolumn{7}{@{}l}{\textit{Drug Design}} \\[1pt]
  \multirow{7}{*}{PK/PD} & Sonnet 4.5 & 38.0 & 96.7 & 50.0 & \textcolor{teal}{+58.7} & \textcolor{red}{$-46.7$} \\
   & Opus 4.6 & 23.3 & 50.0 & 50.0 & \textcolor{teal}{+26.7} & 0.0 \\
   & Sonnet 4.6 & 29.0 & 100.0 & 50.0 & \textcolor{teal}{+71.0} & \textcolor{red}{$-50.0$} \\
   & Gemini 3.1 Pro & 48.0 & 98.3 & 35.0 & \textcolor{teal}{+50.3} & \textcolor{red}{$-63.3$} \\
   & GPT-5.2 & 38.0 & 98.3 & 50.0 & \textcolor{teal}{+60.3} & \textcolor{red}{$-48.3$} \\
   & Gemini 2.0 Flash & 15.0 & 38.3 & 50.0 & \textcolor{teal}{+23.3} & \textcolor{teal}{+11.7} \\
   & GPT-4o & 0.0 & 86.7 & 50.0 & \textcolor{teal}{+86.7} & \textcolor{red}{$-36.7$} \\
\cmidrule{1-7}
\midrule
\multicolumn{7}{@{}l}{\textit{Biology}} \\[1pt]
  \multirow{7}{*}{FBA} & Sonnet 4.5 & 0.0 & 95.0 & 95.0 & \textcolor{teal}{+95.0} & 0.0 \\
   & Opus 4.6 & 3.3 & 91.7 & 100.0 & \textcolor{teal}{+88.3} & \textcolor{teal}{+8.3} \\
   & Sonnet 4.6 & 1.7 & 78.3 & 95.0 & \textcolor{teal}{+76.7} & \textcolor{teal}{+16.7} \\
   & Gemini 3.1 Pro & 1.7 & 93.3 & 95.0 & \textcolor{teal}{+91.7} & \textcolor{teal}{+1.7} \\
   & GPT-5.2 & 18.3 & 93.3 & 95.0 & \textcolor{teal}{+75.0} & \textcolor{teal}{+1.7} \\
   & Gemini 2.0 Flash & 0.0 & 55.0 & 95.0 & \textcolor{teal}{+55.0} & \textcolor{teal}{+40.0} \\
   & GPT-4o & 20.0 & 68.3 & 80.0 & \textcolor{teal}{+48.3} & \textcolor{teal}{+11.7} \\
\cmidrule{1-7}
  \multirow{7}{*}{SSA} & Sonnet 4.5 & 20.0 & 35.0 & 55.0 & \textcolor{teal}{+15.0} & \textcolor{teal}{+20.0} \\
   & Opus 4.6 & 13.3 & 43.3 & 70.0 & \textcolor{teal}{+30.0} & \textcolor{teal}{+26.7} \\
   & Sonnet 4.6 & 16.7 & 30.0 & 65.0 & \textcolor{teal}{+13.3} & \textcolor{teal}{+35.0} \\
   & Gemini 3.1 Pro & 13.3 & 36.7 & 85.0 & \textcolor{teal}{+23.3} & \textcolor{teal}{+48.3} \\
   & GPT-5.2 & 18.3 & 35.0 & 45.0 & \textcolor{teal}{+16.7} & \textcolor{teal}{+10.0} \\
   & Gemini 2.0 Flash & 20.0 & 21.7 & 30.0 & \textcolor{teal}{+1.7} & \textcolor{teal}{+8.3} \\
   & GPT-4o & 18.3 & 15.0 & 30.0 & \textcolor{red}{$-3.3$} & \textcolor{teal}{+15.0} \\
\cmidrule{1-7}
  \multirow{7}{*}{Perturbation} & Sonnet 4.5 & 50.0 & 91.7 & 100.0 & \textcolor{teal}{+41.7} & \textcolor{teal}{+8.3} \\
   & Opus 4.6 & 53.3 & 90.0 & 100.0 & \textcolor{teal}{+36.7} & \textcolor{teal}{+10.0} \\
   & Sonnet 4.6 & 63.3 & 90.0 & 100.0 & \textcolor{teal}{+26.7} & \textcolor{teal}{+10.0} \\
   & Gemini 3.1 Pro & 70.0 & 95.0 & 100.0 & \textcolor{teal}{+25.0} & \textcolor{teal}{+5.0} \\
   & GPT-5.2 & 55.0 & 81.7 & 95.0 & \textcolor{teal}{+26.7} & \textcolor{teal}{+13.3} \\
   & Gemini 2.0 Flash & 33.3 & 60.0 & 85.0 & \textcolor{teal}{+26.7} & \textcolor{teal}{+25.0} \\
   & GPT-4o & 16.7 & 73.3 & 80.0 & \textcolor{teal}{+56.7} & \textcolor{teal}{+6.7} \\
\cmidrule{1-7}
\midrule
\multicolumn{7}{@{}l}{\textit{Physics}} \\[1pt]
  \multirow{7}{*}{Quantum} & Sonnet 4.5 & 48.3 & 50.0 & 50.0 & \textcolor{teal}{+1.7} & 0.0 \\
   & Opus 4.6 & 33.3 & 61.7 & 75.0 & \textcolor{teal}{+28.3} & \textcolor{teal}{+13.3} \\
   & Sonnet 4.6 & 26.7 & 51.7 & 55.0 & \textcolor{teal}{+25.0} & \textcolor{teal}{+3.3} \\
   & Gemini 3.1 Pro & 15.0 & 25.0 & 30.0 & \textcolor{teal}{+10.0} & \textcolor{teal}{+5.0} \\
   & GPT-5.2 & 33.3 & 36.7 & 45.0 & \textcolor{teal}{+3.3} & \textcolor{teal}{+8.3} \\
   & Gemini 2.0 Flash & 23.3 & 46.7 & 50.0 & \textcolor{teal}{+23.3} & \textcolor{teal}{+3.3} \\
   & GPT-4o & 0.0 & 35.0 & 40.0 & \textcolor{teal}{+35.0} & \textcolor{teal}{+5.0} \\
\cmidrule{1-7}
\midrule
\multicolumn{7}{@{}l}{\textit{Engineering}} \\[1pt]
  \multirow{7}{*}{Controls} & Sonnet 4.5 & 1.7 & 23.3 & 40.0 & \textcolor{teal}{+21.7} & \textcolor{teal}{+16.7} \\
   & Opus 4.6 & 3.3 & 36.7 & 45.0 & \textcolor{teal}{+33.3} & \textcolor{teal}{+8.3} \\
   & Sonnet 4.6 & 0.0 & 26.7 & 50.0 & \textcolor{teal}{+26.7} & \textcolor{teal}{+23.3} \\
   & Gemini 3.1 Pro & 8.3 & 35.0 & 45.0 & \textcolor{teal}{+26.7} & \textcolor{teal}{+10.0} \\
   & GPT-5.2 & 0.0 & 3.3 & 45.0 & \textcolor{teal}{+3.3} & \textcolor{teal}{+41.7} \\
   & Gemini 2.0 Flash & 0.0 & 3.3 & 20.0 & \textcolor{teal}{+3.3} & \textcolor{teal}{+16.7} \\
   & GPT-4o & 0.0 & 0.0 & 5.0 & 0.0 & \textcolor{teal}{+5.0} \\
\cmidrule{1-7}
  \multirow{7}{*}{Signal Processing} & Sonnet 4.5 & 30.0 & 53.3 & 60.0 & \textcolor{teal}{+23.3} & \textcolor{teal}{+6.7} \\
   & Opus 4.6 & 33.3 & 55.0 & 70.0 & \textcolor{teal}{+21.7} & \textcolor{teal}{+15.0} \\
   & Sonnet 4.6 & 30.0 & 53.3 & 60.0 & \textcolor{teal}{+23.3} & \textcolor{teal}{+6.7} \\
   & Gemini 3.1 Pro & 31.7 & 43.3 & 50.0 & \textcolor{teal}{+11.7} & \textcolor{teal}{+6.7} \\
   & GPT-5.2 & 30.0 & 51.7 & 60.0 & \textcolor{teal}{+21.7} & \textcolor{teal}{+8.3} \\
   & Gemini 2.0 Flash & 40.0 & 53.3 & 60.0 & \textcolor{teal}{+13.3} & \textcolor{teal}{+6.7} \\
   & GPT-4o & 35.0 & 48.3 & 55.0 & \textcolor{teal}{+13.3} & \textcolor{teal}{+6.7} \\
\cmidrule{1-7}
  \multirow{7}{*}{Alloy} & Sonnet 4.5 & 30.0 & 36.7 & 45.0 & \textcolor{teal}{+6.7} & \textcolor{teal}{+8.3} \\
   & Opus 4.6 & 25.0 & 36.7 & 50.0 & \textcolor{teal}{+11.7} & \textcolor{teal}{+13.3} \\
   & Sonnet 4.6 & 0.0 & 35.0 & 45.0 & \textcolor{teal}{+35.0} & \textcolor{teal}{+10.0} \\
   & Gemini 3.1 Pro & 23.3 & 30.0 & 40.0 & \textcolor{teal}{+6.7} & \textcolor{teal}{+10.0} \\
   & GPT-5.2 & 13.3 & 33.3 & 40.0 & \textcolor{teal}{+20.0} & \textcolor{teal}{+6.7} \\
   & Gemini 2.0 Flash & 20.0 & 31.7 & 40.0 & \textcolor{teal}{+11.7} & \textcolor{teal}{+8.3} \\
   & GPT-4o & 11.7 & 33.3 & 35.0 & \textcolor{teal}{+21.7} & \textcolor{teal}{+1.7} \\
\cmidrule{1-7}
\midrule
\multicolumn{7}{@{}l}{\textit{Chemical Engineering}} \\[1pt]
  \multirow{7}{*}{Reactor} & Sonnet 4.5 & 45.0 & 91.7 & 100.0 & \textcolor{teal}{+46.7} & \textcolor{teal}{+8.3} \\
   & Opus 4.6 & 30.0 & 98.3 & 100.0 & \textcolor{teal}{+68.3} & \textcolor{teal}{+1.7} \\
   & Sonnet 4.6 & 28.3 & 100.0 & 100.0 & \textcolor{teal}{+71.7} & 0.0 \\
   & Gemini 3.1 Pro & 0.0 & 18.3 & 25.0 & \textcolor{teal}{+18.3} & \textcolor{teal}{+6.7} \\
   & GPT-5.2 & 31.7 & 93.3 & 95.0 & \textcolor{teal}{+61.7} & \textcolor{teal}{+1.7} \\
   & Gemini 2.0 Flash & 20.0 & 46.7 & 85.0 & \textcolor{teal}{+26.7} & \textcolor{teal}{+38.3} \\
   & GPT-4o & 18.3 & 76.7 & 80.0 & \textcolor{teal}{+58.3} & \textcolor{teal}{+3.3} \\
\cmidrule{1-7}
  \multirow{7}{*}{Heat Exchanger} & Sonnet 4.5 & 26.7 & 91.7 & 100.0 & \textcolor{teal}{+65.0} & \textcolor{teal}{+8.3} \\
   & Opus 4.6 & 20.0 & 91.7 & 100.0 & \textcolor{teal}{+71.7} & \textcolor{teal}{+8.3} \\
   & Sonnet 4.6 & 33.3 & 100.0 & 95.0 & \textcolor{teal}{+66.7} & \textcolor{red}{$-5.0$} \\
   & Gemini 3.1 Pro & 28.3 & 98.3 & 100.0 & \textcolor{teal}{+70.0} & \textcolor{teal}{+1.7} \\
   & GPT-5.2 & 18.3 & 98.3 & 100.0 & \textcolor{teal}{+80.0} & \textcolor{teal}{+1.7} \\
   & Gemini 2.0 Flash & 23.3 & 75.0 & 90.0 & \textcolor{teal}{+51.7} & \textcolor{teal}{+15.0} \\
   & GPT-4o & 8.3 & 93.3 & 95.0 & \textcolor{teal}{+85.0} & \textcolor{teal}{+1.7} \\
\cmidrule{1-7}
\bottomrule
\end{tabular}
\end{table}

\subsection{Design Optimization: \emph{De Novo} vs.\ Optimization Comparison}

\Cref{tab:app_optimization_comparison} provides a per-domain comparison of \emph{de novo} zero-shot and design optimization performance across the 12 benchmark domains shared by both settings. Positive $\Delta$ means the starting design helps; negative $\Delta$ means it hurts.

\begin{table}[p]
\centering
\caption{Per-domain comparison of de novo zero-shot vs.\ design optimization success rate (\%) across the 12 benchmark domains shared by both settings. ADMET is excluded because it is benchmarked only in optimization, and Perturbation is excluded because it has no optimization tasks.}
\label{tab:app_optimization_comparison}
\tiny
\setlength{\tabcolsep}{2pt}
\renewcommand{\arraystretch}{0.85}
\begin{tabular}{@{}ll cc c ll cc c@{}}
\toprule
\textbf{Domain} & \textbf{Model} & \textbf{De novo} & \textbf{Opt.} & \textbf{$\Delta$} & \textbf{Domain} & \textbf{Model} & \textbf{De novo} & \textbf{Opt.} & \textbf{$\Delta$} \\
\midrule
  \multirow{7}{*}{PK/PD} & Sonnet 4.5 & 38.0 & 48.3 & \textcolor{teal}{+10.3} & \multirow{7}{*}{Docking} & Sonnet 4.5 & 53.3 & 36.7 & \textcolor{red}{$-16.7$} \\
   & Opus 4.6 & 23.3 & 43.3 & \textcolor{teal}{+20.0} &  & Opus 4.6 & 56.7 & 61.7 & \textcolor{teal}{+5.0} \\
   & Sonnet 4.6 & 29.0 & 41.7 & \textcolor{teal}{+12.7} &  & Sonnet 4.6 & 46.7 & 51.7 & \textcolor{teal}{+5.0} \\
   & Gemini 3.1 Pro & 48.0 & 51.7 & \textcolor{teal}{+3.7} &  & Gemini 3.1 Pro & 31.7 & 61.7 & \textcolor{teal}{+30.0} \\
   & GPT-5.2 & 38.0 & 41.7 & \textcolor{teal}{+3.7} &  & GPT-5.2 & --- & 38.3 & --- \\
   & Gemini 2.0 Flash & 15.0 & 43.3 & \textcolor{teal}{+28.3} &  & Gemini 2.0 Flash & 10.0 & 25.0 & \textcolor{teal}{+15.0} \\
   & GPT-4o & 0.0 & 5.0 & \textcolor{teal}{+5.0} &  & GPT-4o & --- & 45.0 & --- \\
\midrule
  \multirow{7}{*}{FBA} & Sonnet 4.5 & 0.0 & 1.7 & \textcolor{teal}{+1.7} & \multirow{7}{*}{SSA} & Sonnet 4.5 & 20.0 & 43.3 & \textcolor{teal}{+23.3} \\
   & Opus 4.6 & 3.3 & 20.0 & \textcolor{teal}{+16.7} &  & Opus 4.6 & 13.3 & 55.0 & \textcolor{teal}{+41.7} \\
   & Sonnet 4.6 & 1.7 & 8.3 & \textcolor{teal}{+6.7} &  & Sonnet 4.6 & 16.7 & 38.3 & \textcolor{teal}{+21.7} \\
   & Gemini 3.1 Pro & 1.7 & 8.3 & \textcolor{teal}{+6.7} &  & Gemini 3.1 Pro & 13.3 & 40.0 & \textcolor{teal}{+26.7} \\
   & GPT-5.2 & 18.3 & 6.7 & \textcolor{red}{$-11.7$} &  & GPT-5.2 & 18.3 & 50.0 & \textcolor{teal}{+31.7} \\
   & Gemini 2.0 Flash & 0.0 & 1.7 & \textcolor{teal}{+1.7} &  & Gemini 2.0 Flash & 20.0 & 48.3 & \textcolor{teal}{+28.3} \\
   & GPT-4o & 20.0 & 1.7 & \textcolor{red}{$-18.3$} &  & GPT-4o & 18.3 & 46.7 & \textcolor{teal}{+28.3} \\
\midrule
  \multirow{7}{*}{RNA Design} & Sonnet 4.5 & 3.3 & 36.7 & \textcolor{teal}{+33.3} & \multirow{7}{*}{Quantum} & Sonnet 4.5 & 48.3 & 45.0 & \textcolor{red}{$-3.3$} \\
   & Opus 4.6 & 5.0 & 43.3 & \textcolor{teal}{+38.3} &  & Opus 4.6 & 33.3 & 36.7 & \textcolor{teal}{+3.3} \\
   & Sonnet 4.6 & 20.0 & 28.3 & \textcolor{teal}{+8.3} &  & Sonnet 4.6 & 26.7 & 38.3 & \textcolor{teal}{+11.7} \\
   & Gemini 3.1 Pro & 23.3 & 41.7 & \textcolor{teal}{+18.3} &  & Gemini 3.1 Pro & 15.0 & 18.3 & \textcolor{teal}{+3.3} \\
   & GPT-5.2 & --- & 1.7 & --- &  & GPT-5.2 & 33.3 & 33.3 & 0.0 \\
   & Gemini 2.0 Flash & 0.0 & 3.3 & \textcolor{teal}{+3.3} &  & Gemini 2.0 Flash & 23.3 & 0.0 & \textcolor{red}{$-23.3$} \\
   & GPT-4o & --- & 1.7 & --- &  & GPT-4o & 0.0 & 21.7 & \textcolor{teal}{+21.7} \\
\midrule
  \multirow{7}{*}{Thin Film} & Sonnet 4.5 & 33.3 & 31.7 & \textcolor{red}{$-1.7$} & \multirow{7}{*}{Controls} & Sonnet 4.5 & 1.7 & 11.7 & \textcolor{teal}{+10.0} \\
   & Opus 4.6 & 28.3 & 50.0 & \textcolor{teal}{+21.7} &  & Opus 4.6 & 3.3 & 15.0 & \textcolor{teal}{+11.7} \\
   & Sonnet 4.6 & 28.3 & 40.0 & \textcolor{teal}{+11.7} &  & Sonnet 4.6 & 0.0 & 10.0 & \textcolor{teal}{+10.0} \\
   & Gemini 3.1 Pro & 18.3 & 15.0 & \textcolor{red}{$-3.3$} &  & Gemini 3.1 Pro & 8.3 & 11.7 & \textcolor{teal}{+3.3} \\
   & GPT-5.2 & --- & 38.3 & --- &  & GPT-5.2 & 0.0 & 11.7 & \textcolor{teal}{+11.7} \\
   & Gemini 2.0 Flash & 31.7 & 26.7 & \textcolor{red}{$-5.0$} &  & Gemini 2.0 Flash & 0.0 & 15.0 & \textcolor{teal}{+15.0} \\
   & GPT-4o & --- & 18.3 & --- &  & GPT-4o & 0.0 & 10.0 & \textcolor{teal}{+10.0} \\
\midrule
  \multirow{7}{*}{Signal Processing} & Sonnet 4.5 & 30.0 & 38.3 & \textcolor{teal}{+8.3} & \multirow{7}{*}{Alloy} & Sonnet 4.5 & 30.0 & 23.3 & \textcolor{red}{$-6.7$} \\
   & Opus 4.6 & 33.3 & 33.3 & 0.0 &  & Opus 4.6 & 25.0 & 20.0 & \textcolor{red}{$-5.0$} \\
   & Sonnet 4.6 & 30.0 & 38.3 & \textcolor{teal}{+8.3} &  & Sonnet 4.6 & 0.0 & 25.0 & \textcolor{teal}{+25.0} \\
   & Gemini 3.1 Pro & 31.7 & 33.3 & \textcolor{teal}{+1.7} &  & Gemini 3.1 Pro & 23.3 & 28.3 & \textcolor{teal}{+5.0} \\
   & GPT-5.2 & 30.0 & 38.3 & \textcolor{teal}{+8.3} &  & GPT-5.2 & 13.3 & 25.0 & \textcolor{teal}{+11.7} \\
   & Gemini 2.0 Flash & 40.0 & 30.0 & \textcolor{red}{$-10.0$} &  & Gemini 2.0 Flash & 20.0 & 21.7 & \textcolor{teal}{+1.7} \\
   & GPT-4o & 35.0 & 30.0 & \textcolor{red}{$-5.0$} &  & GPT-4o & 11.7 & 15.0 & \textcolor{teal}{+3.3} \\
\midrule
  \multirow{7}{*}{Reactor} & Sonnet 4.5 & 45.0 & 36.7 & \textcolor{red}{$-8.3$} & \multirow{7}{*}{Heat Exchanger} & Sonnet 4.5 & 26.7 & 26.7 & 0.0 \\
   & Opus 4.6 & 30.0 & 56.7 & \textcolor{teal}{+26.7} &  & Opus 4.6 & 20.0 & 31.7 & \textcolor{teal}{+11.7} \\
   & Sonnet 4.6 & 28.3 & 46.7 & \textcolor{teal}{+18.3} &  & Sonnet 4.6 & 33.3 & 30.0 & \textcolor{red}{$-3.3$} \\
   & Gemini 3.1 Pro & 0.0 & 51.7 & \textcolor{teal}{+51.7} &  & Gemini 3.1 Pro & 28.3 & 30.0 & \textcolor{teal}{+1.7} \\
   & GPT-5.2 & 31.7 & 31.7 & 0.0 &  & GPT-5.2 & 18.3 & 26.7 & \textcolor{teal}{+8.3} \\
   & Gemini 2.0 Flash & 20.0 & 18.3 & \textcolor{red}{$-1.7$} &  & Gemini 2.0 Flash & 23.3 & 31.7 & \textcolor{teal}{+8.3} \\
   & GPT-4o & 18.3 & 8.3 & \textcolor{red}{$-10.0$} &  & GPT-4o & 8.3 & 18.3 & \textcolor{teal}{+10.0} \\
\bottomrule
\end{tabular}
\end{table}

Several patterns emerge from this comparison:
\begin{itemize}[nosep,leftmargin=*]
  \item \textbf{Optimization is highly domain-dependent.} The starting-design constraint helps some tasks substantially while hurting others, so optimization should be read as a distinct capability rather than a uniformly easier version of \emph{de novo} design.
  \item \textbf{Model rankings are not stable across settings.} Some models handle the optimization prompt and modification constraint much better than their zero-shot \emph{de novo} ranking alone would predict.
  \item \textbf{Several domains remain difficult in both modes.} Even when a seed design is provided, benchmark success still depends on satisfying the underlying scientific constraints rather than merely emitting a valid format.
\end{itemize}

\subsection{Agentic Design Optimization: Per-Domain Results}

\Cref{tab:app_agentic_optimization_domains} presents the per-domain results for the 20-turn long-horizon optimization setting, alongside the 20-turn \emph{de novo} results on the 9 domains shared by both shared-core domain sets. This isolates the effect of providing a starting design under long-horizon iterative feedback.

\begin{table}[p]
\centering
\caption{Per-domain success rate (\%) for 20-turn de novo agentic vs.\ 20-turn agentic optimization on the 9 domains shared by both shared-core domain sets. $\Delta$: optimization minus de novo.}
\label{tab:app_agentic_optimization_domains}
\tiny
\setlength{\tabcolsep}{2pt}
\renewcommand{\arraystretch}{0.85}
\begin{tabular}{@{}l ccc ccc ccc ccc ccc@{}}
\toprule
& \multicolumn{3}{c}{\textbf{Alloy}}  & \multicolumn{3}{c}{\textbf{Controls}}  & \multicolumn{3}{c}{\textbf{FBA}}  & \multicolumn{3}{c}{\textbf{Heat Exchanger}}  & \multicolumn{3}{c}{\textbf{PK/PD}} \\
\cmidrule(lr){2-4} \cmidrule(lr){5-7} \cmidrule(lr){8-10} \cmidrule(lr){11-13} \cmidrule(lr){14-16}
\textbf{Model} & DN & Opt & $\Delta$  & DN & Opt & $\Delta$  & DN & Opt & $\Delta$  & DN & Opt & $\Delta$  & DN & Opt & $\Delta$ \\
\midrule
  Sonnet 4.5 & 45.0 & 35.0 & \textcolor{red}{$-10.0$} & 40.0 & 60.0 & \textcolor{teal}{+20.0} & 95.0 & 90.0 & \textcolor{red}{$-5.0$} & 100.0 & 40.0 & \textcolor{red}{$-60.0$} & 50.0 & 70.0 & \textcolor{teal}{+20.0} \\
  Opus 4.6 & 50.0 & 35.0 & \textcolor{red}{$-15.0$} & 45.0 & 70.0 & \textcolor{teal}{+25.0} & 100.0 & 90.0 & \textcolor{red}{$-10.0$} & 100.0 & 40.0 & \textcolor{red}{$-60.0$} & 50.0 & 75.0 & \textcolor{teal}{+25.0} \\
  Sonnet 4.6 & 45.0 & 35.0 & \textcolor{red}{$-10.0$} & 50.0 & 65.0 & \textcolor{teal}{+15.0} & 95.0 & 90.0 & \textcolor{red}{$-5.0$} & 95.0 & 40.0 & \textcolor{red}{$-55.0$} & 50.0 & 70.0 & \textcolor{teal}{+20.0} \\
  Gemini 3.1 Pro & 40.0 & 35.0 & \textcolor{red}{$-5.0$} & 45.0 & 55.0 & \textcolor{teal}{+10.0} & 95.0 & 55.0 & \textcolor{red}{$-40.0$} & 100.0 & 40.0 & \textcolor{red}{$-60.0$} & 35.0 & 65.0 & \textcolor{teal}{+30.0} \\
  GPT-5.2 & 40.0 & 35.0 & \textcolor{red}{$-5.0$} & 45.0 & 60.0 & \textcolor{teal}{+15.0} & 95.0 & 85.0 & \textcolor{red}{$-10.0$} & 100.0 & 40.0 & \textcolor{red}{$-60.0$} & 50.0 & 60.0 & \textcolor{teal}{+10.0} \\
  Gemini 2.0 Flash & 40.0 & 35.0 & \textcolor{red}{$-5.0$} & 20.0 & 45.0 & \textcolor{teal}{+25.0} & 95.0 & 75.0 & \textcolor{red}{$-20.0$} & 90.0 & 40.0 & \textcolor{red}{$-50.0$} & 50.0 & 60.0 & \textcolor{teal}{+10.0} \\
  GPT-4o & 35.0 & 25.0 & \textcolor{red}{$-10.0$} & 5.0 & 40.0 & \textcolor{teal}{+35.0} & 80.0 & 50.0 & \textcolor{red}{$-30.0$} & 95.0 & 40.0 & \textcolor{red}{$-55.0$} & 50.0 & 55.0 & \textcolor{teal}{+5.0} \\
\midrule
\multicolumn{16}{@{}c@{}}{} \\[-4pt]
\bottomrule
\end{tabular}

\vspace{6pt}

\begin{tabular}{@{}l ccc ccc ccc ccc@{}}
\toprule
& \multicolumn{3}{c}{\textbf{Quantum}}  & \multicolumn{3}{c}{\textbf{Reactor}}  & \multicolumn{3}{c}{\textbf{Signal Processing}}  & \multicolumn{3}{c}{\textbf{SSA}} \\
\cmidrule(lr){2-4} \cmidrule(lr){5-7} \cmidrule(lr){8-10} \cmidrule(lr){11-13}
\textbf{Model} & DN & Opt & $\Delta$  & DN & Opt & $\Delta$  & DN & Opt & $\Delta$  & DN & Opt & $\Delta$ \\
\midrule
  Sonnet 4.5 & 50.0 & 50.0 & 0.0 & 100.0 & 55.0 & \textcolor{red}{$-45.0$} & 60.0 & 75.0 & \textcolor{teal}{+15.0} & 55.0 & 85.0 & \textcolor{teal}{+30.0} \\
  Opus 4.6 & 75.0 & 85.0 & \textcolor{teal}{+10.0} & 100.0 & 65.0 & \textcolor{red}{$-35.0$} & 70.0 & 75.0 & \textcolor{teal}{+5.0} & 70.0 & 80.0 & \textcolor{teal}{+10.0} \\
  Sonnet 4.6 & 55.0 & 65.0 & \textcolor{teal}{+10.0} & 100.0 & 65.0 & \textcolor{red}{$-35.0$} & 60.0 & 65.0 & \textcolor{teal}{+5.0} & 65.0 & 85.0 & \textcolor{teal}{+20.0} \\
  Gemini 3.1 Pro & 30.0 & 50.0 & \textcolor{teal}{+20.0} & 25.0 & 60.0 & \textcolor{teal}{+35.0} & 50.0 & 65.0 & \textcolor{teal}{+15.0} & 85.0 & 70.0 & \textcolor{red}{$-15.0$} \\
  GPT-5.2 & 45.0 & 45.0 & 0.0 & 95.0 & 55.0 & \textcolor{red}{$-40.0$} & 60.0 & 75.0 & \textcolor{teal}{+15.0} & 45.0 & 80.0 & \textcolor{teal}{+35.0} \\
  Gemini 2.0 Flash & 50.0 & 15.0 & \textcolor{red}{$-35.0$} & 85.0 & 50.0 & \textcolor{red}{$-35.0$} & 60.0 & 65.0 & \textcolor{teal}{+5.0} & 30.0 & 60.0 & \textcolor{teal}{+30.0} \\
  GPT-4o & 40.0 & 35.0 & \textcolor{red}{$-5.0$} & 80.0 & 5.0 & \textcolor{red}{$-75.0$} & 55.0 & 60.0 & \textcolor{teal}{+5.0} & 30.0 & 65.0 & \textcolor{teal}{+35.0} \\
\bottomrule
\end{tabular}
\end{table}

Key observations:
\begin{itemize}[nosep,leftmargin=*]
  \item \textbf{Long-horizon optimization is strongly heterogeneous.} Some domains benefit from iterative refinement around a seed design, while others degrade because the initial design constrains the search too aggressively.
  \item \textbf{More turns do not fix every domain.} Domains near the floor in \emph{de novo} design often remain difficult even with twenty rounds of structured simulator feedback.
  \item \textbf{Large negative deltas are scientifically informative.} When long-horizon optimization underperforms \emph{de novo} search, it suggests that local refinement is a poor inductive bias for that domain rather than merely a prompting artifact.
\end{itemize}

\subsection{Runtime Distributions}

For each task, we sum the logged model generation time and oracle execution time across all turns and attempts; this captures the real inference-plus-simulation cost of solving a task while excluding scheduler idle time between jobs. \Cref{tab:app_runtime_summary} reports mean $\pm$ standard deviation.

\begin{table}[h]
\centering
\caption{Per-task active runtime (seconds) aggregated across the official benchmark tasks. Each cell reports mean $\pm$ standard deviation of the logged generation time plus oracle execution time summed over all attempts and turns for that task. External scheduler idle time is excluded.}
\label{tab:app_runtime_summary}
\scriptsize
\setlength{\tabcolsep}{4pt}
\begin{tabular}{@{}lccccc@{}}
\toprule
\textbf{Model} & \textbf{DN 1-turn} & \textbf{DN 5-turn} & \textbf{DN 20-turn} & \textbf{Opt. 1-turn} & \textbf{Opt. 20-turn} \\
\midrule
Sonnet 4.5 & $41.9 \pm 15.0$ & $115 \pm 138$ & $87.2 \pm 95.0$ & $39.1 \pm 14.7$ & $102 \pm 250$ \\
Opus 4.6 & $52.5 \pm 16.9$ & $149 \pm 209$ & $100.0 \pm 108$ & $55.6 \pm 14.3$ & $109 \pm 254$ \\
Sonnet 4.6 & $61.5 \pm 171$ & $114 \pm 89.9$ & $85.9 \pm 95.3$ & $43.9 \pm 15.1$ & $252 \pm 2406$ \\
Gemini 3.1 Pro & $52.5 \pm 41.1$ & $152 \pm 86.0$ & $151 \pm 126$ & $58.5 \pm 10.0$ & $185 \pm 179$ \\
GPT-5.2 & $5.46 \pm 5.25$ & $35.0 \pm 181$ & $22.8 \pm 42.3$ & $6.85 \pm 11.3$ & $24.4 \pm 53.8$ \\
Gemini Flash & $7.12 \pm 5.76$ & $77.5 \pm 837$ & $24.7 \pm 45.0$ & $8.52 \pm 7.83$ & $22.9 \pm 29.6$ \\
GPT-4o & $30.5 \pm 66.3$ & $91.2 \pm 162$ & $80.1 \pm 83.1$ & $25.4 \pm 8.84$ & $124 \pm 135$ \\
\bottomrule
\end{tabular}
\end{table}

\subsection{Additional Figures}

\begin{figure}[h]
  \centering
  \maybeincludegraphics[width=0.85\linewidth]{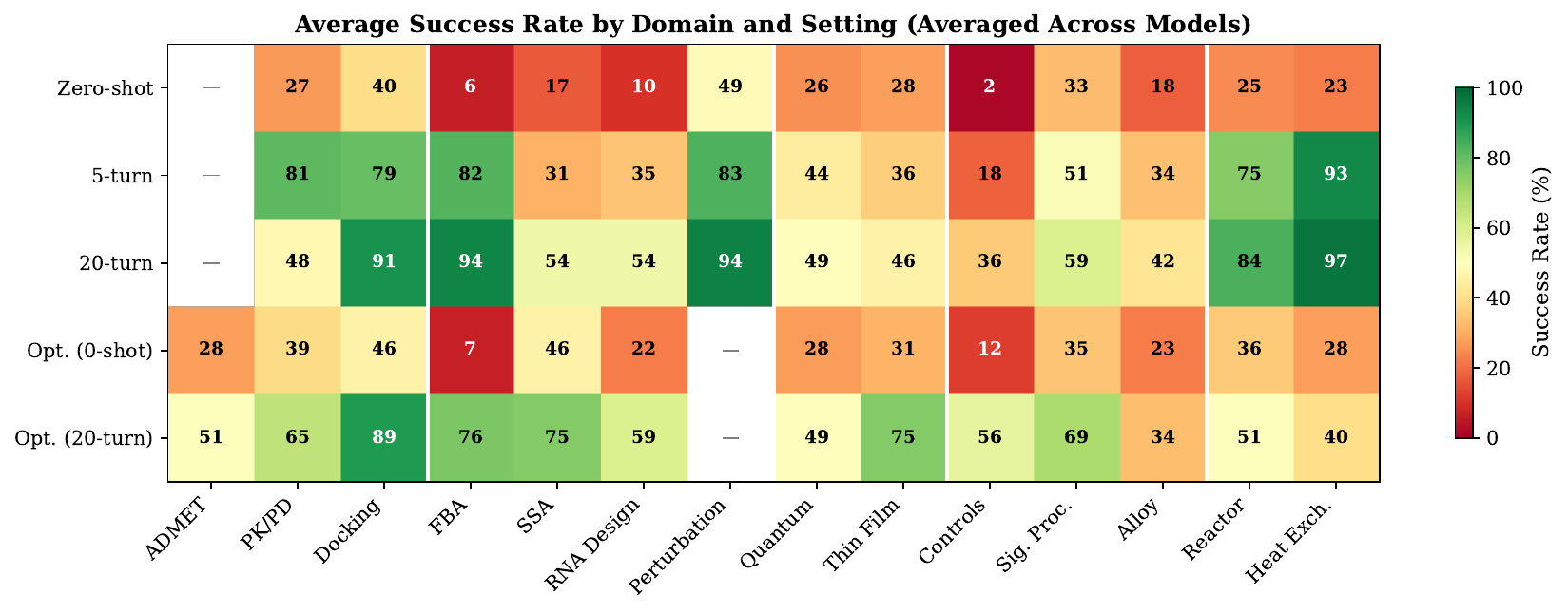}
  \caption{Average success rate by domain and evaluation setting, averaged across all seven models. ADMET (in optimization) and Perturbation (in \emph{de novo} design) are among the most accessible domains; domains such as FBA remain challenging and uneven across models; feedback and optimization can unlock domains like Controls and SSA that are weak in zero-shot.}
  \label{fig:domain_heatmap}
\end{figure}

\begin{figure}[h]
  \centering
  \maybeincludegraphics[width=0.7\linewidth]{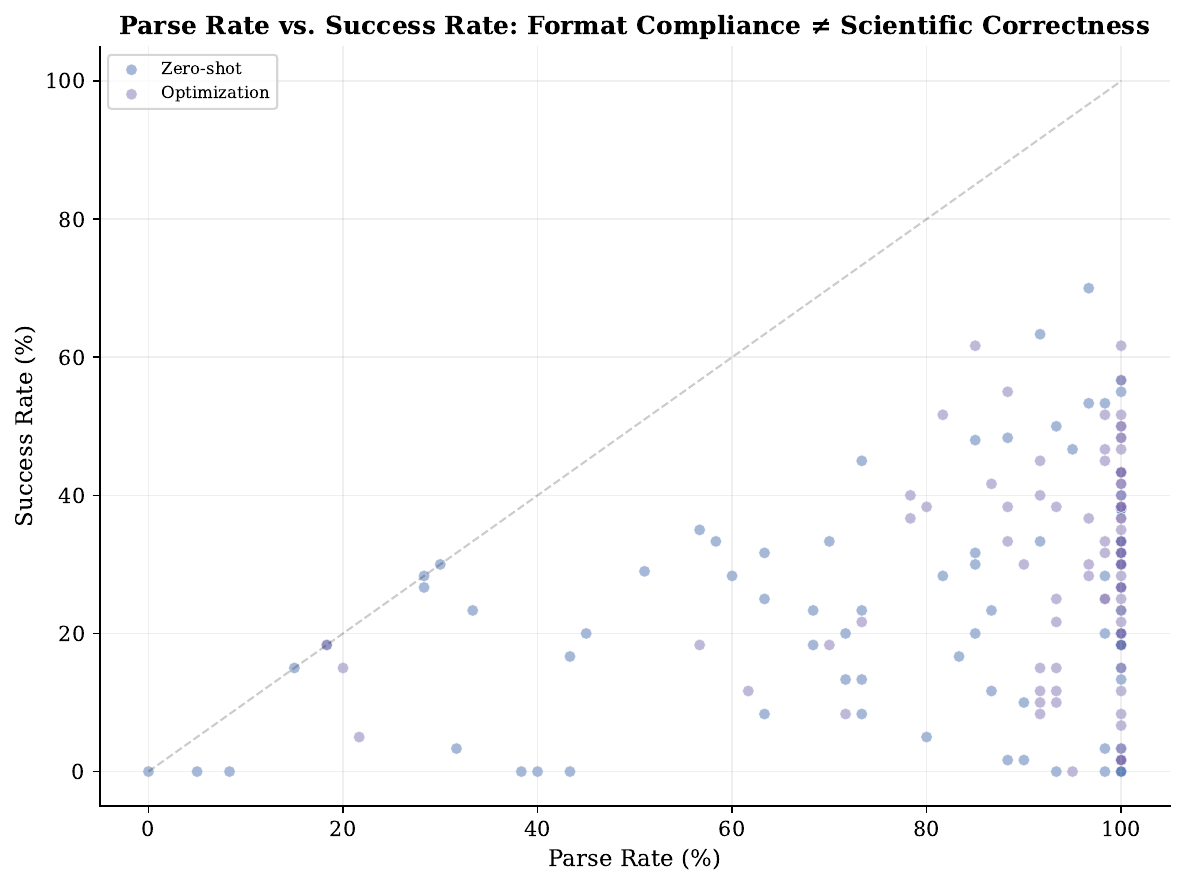}
  \caption{Parse rate vs.\ success rate for each model--domain combination in zero-shot and design optimization settings. Points cluster in the upper-left quadrant: high parse rates but low success rates, confirming that format compliance does not imply scientific correctness.}
  \label{fig:parse_vs_success_app}
\end{figure}

\begin{figure}[h]
  \centering
  \maybeincludegraphics[width=0.85\linewidth]{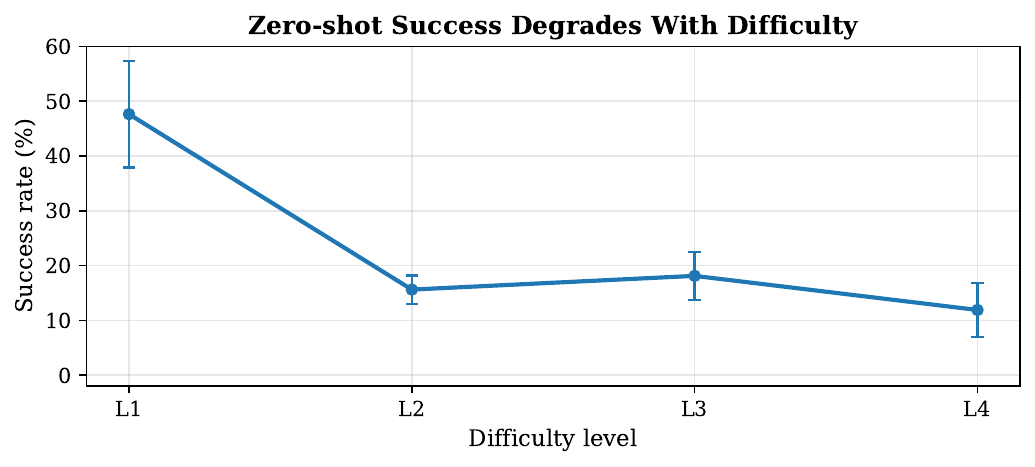}
  \caption{Success rate by difficulty level, averaged across all domains and models. Error bars show standard error across domains. Performance degrades from $\sim$45\% at L1 to $\sim$12\% at L4.}
  \label{fig:difficulty_app}
\end{figure}

% Moved to main text as fig:turn_curves_domain

\begin{figure}[h]
  \centering
  \maybeincludegraphics[width=\linewidth]{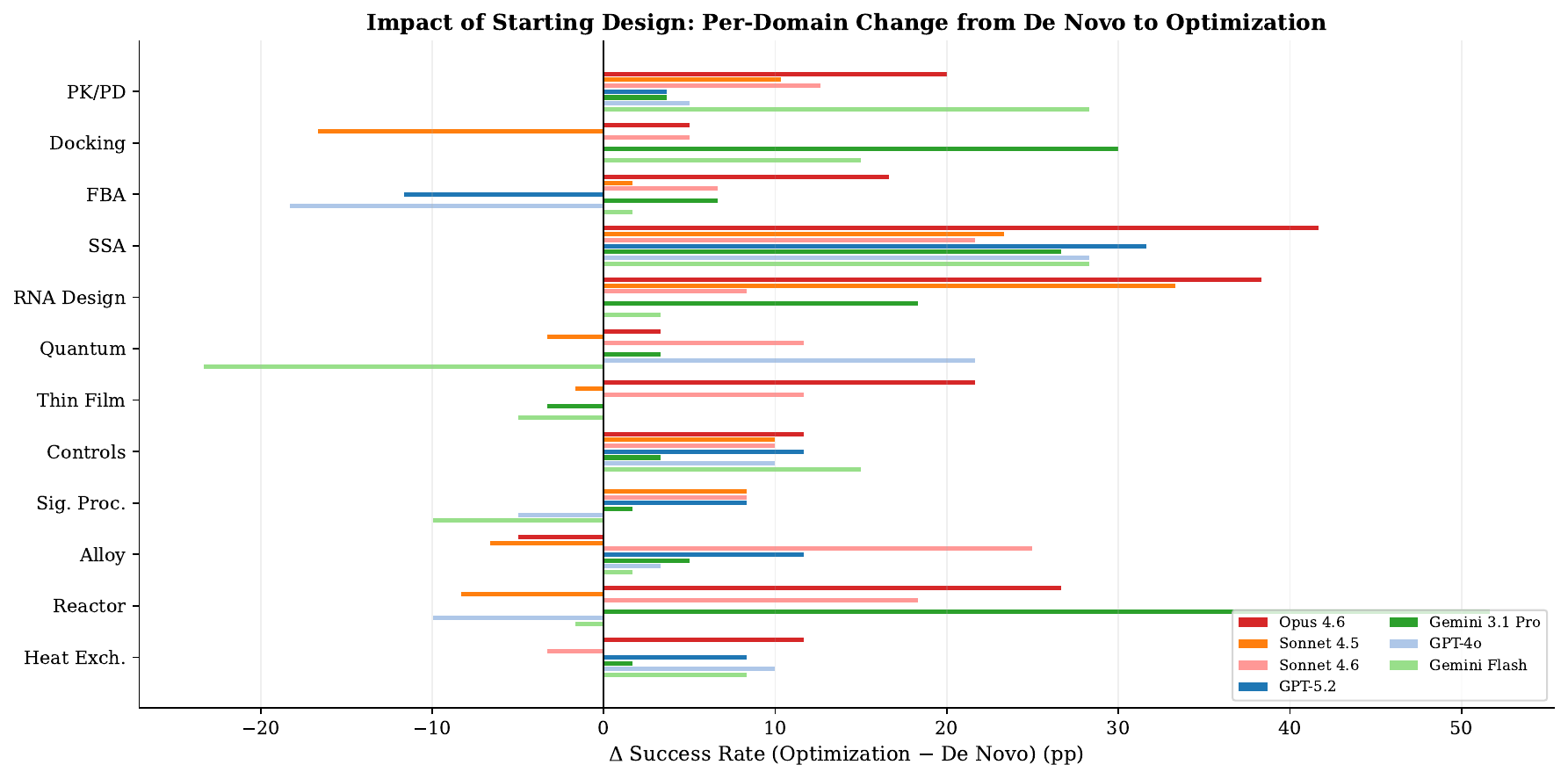}
  \caption{Per-domain change in success rate from \emph{de novo} zero-shot to design optimization on the 12 domains shared by both settings. Positive values (right) indicate the starting design helps; negative values (left) indicate it hurts. Controls improves consistently, while Reactor and Quantum often degrade.}
  \label{fig:optimization_delta_app}
\end{figure}

\begin{figure}[h]
  \centering
  \maybeincludegraphics[width=\linewidth]{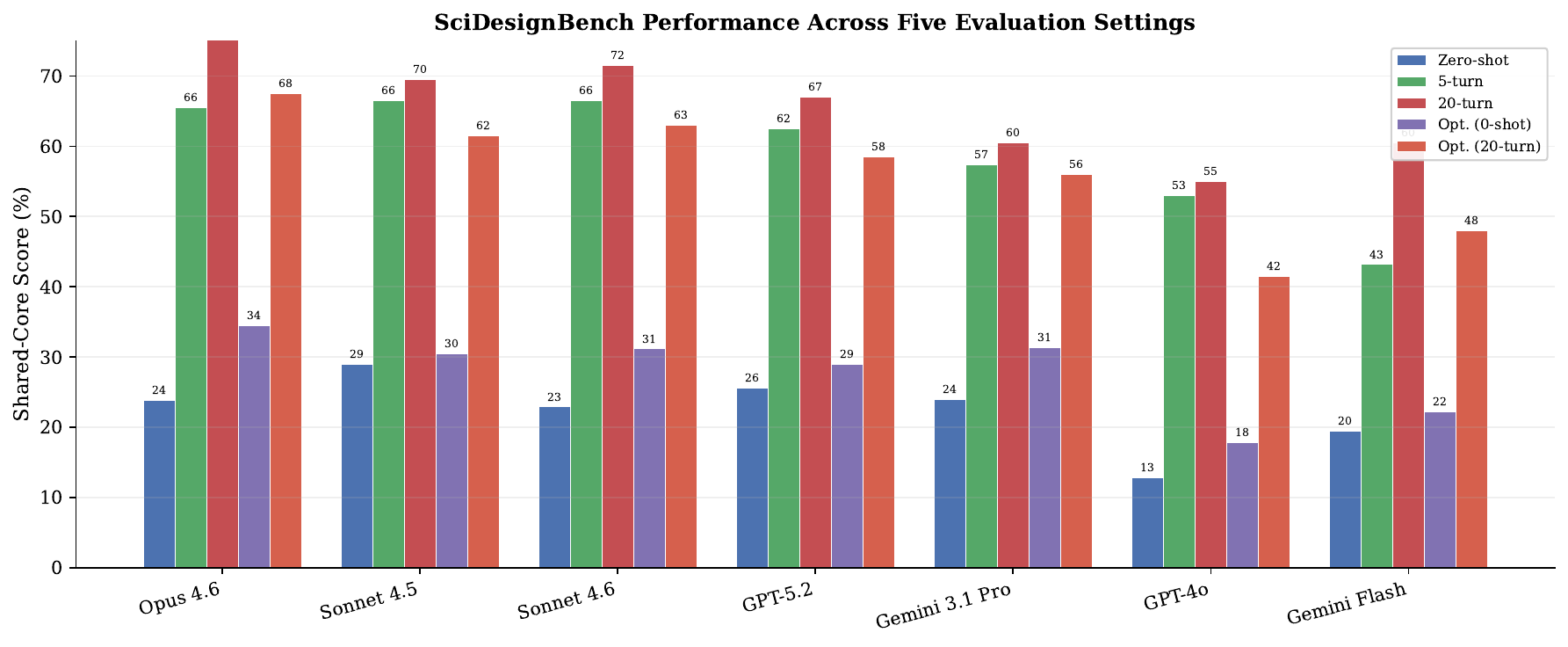}
  \caption{Manifest-defined shared-core scores across five evaluation settings. The grouped bars compare zero-shot \emph{de novo}, 5-turn feedback, 20-turn long-horizon, zero-shot design optimization, and 20-turn long-horizon optimization. Leadership shifts across settings, underscoring that static design, feedback use, and optimization are distinct capabilities.}
  \label{fig:four_settings_bar}
\end{figure}

\begin{figure}[h]
  \centering
  \maybeincludegraphics[width=\linewidth]{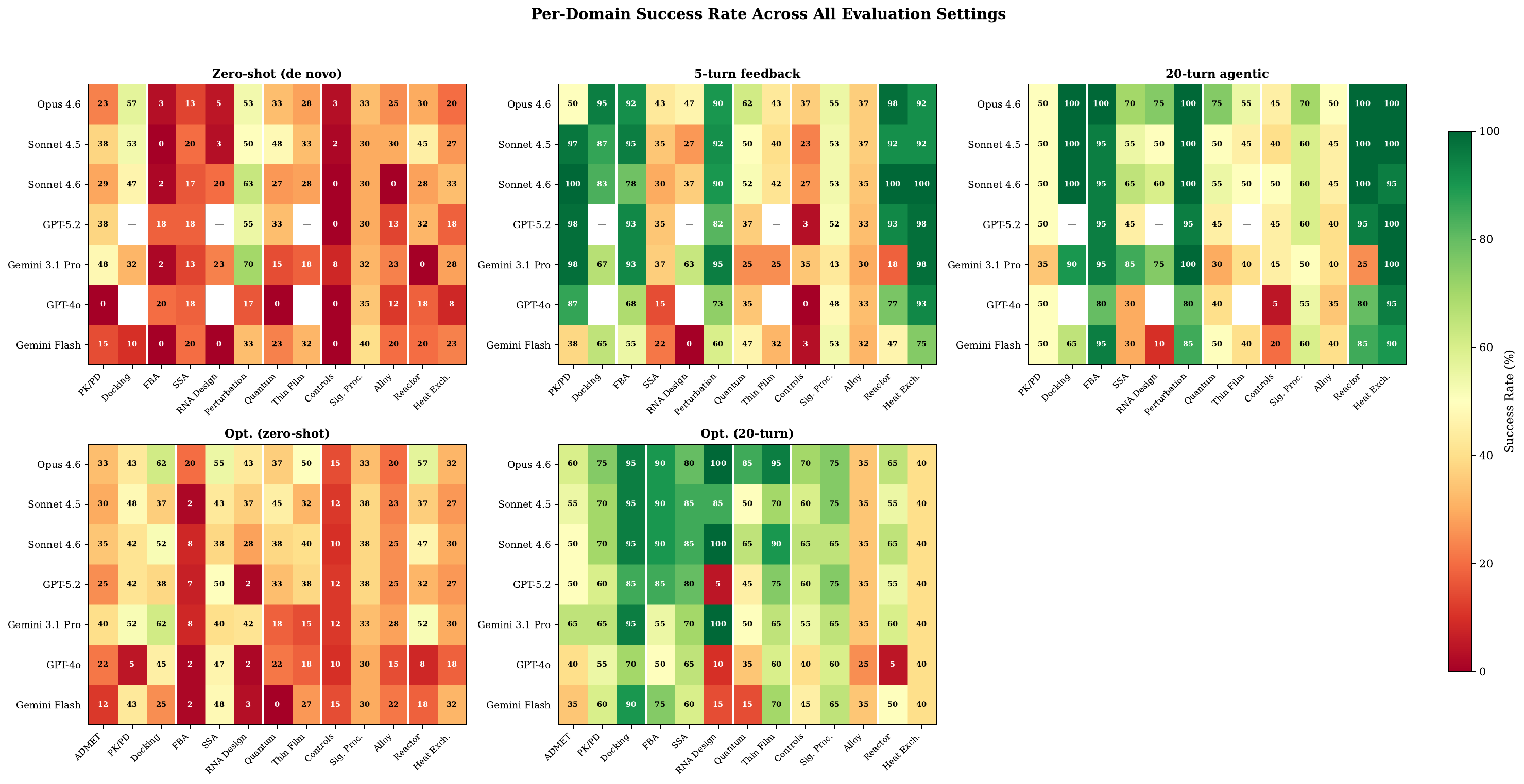}
  \caption{Per-domain success rate heatmap across five evaluation settings and all seven models. Cell values show success rate (\%). Green indicates high success; red indicates low. The \emph{de novo} panels use the 13 official \bench{}~v1 \emph{de novo} domains, while the optimization panels use the 13 official optimization domains; ADMET appears only in optimization and Perturbation only in \emph{de novo}.}
  \label{fig:heatmap_all}
\end{figure}

\begin{figure}[h]
  \centering
  \maybeincludegraphics[width=\linewidth]{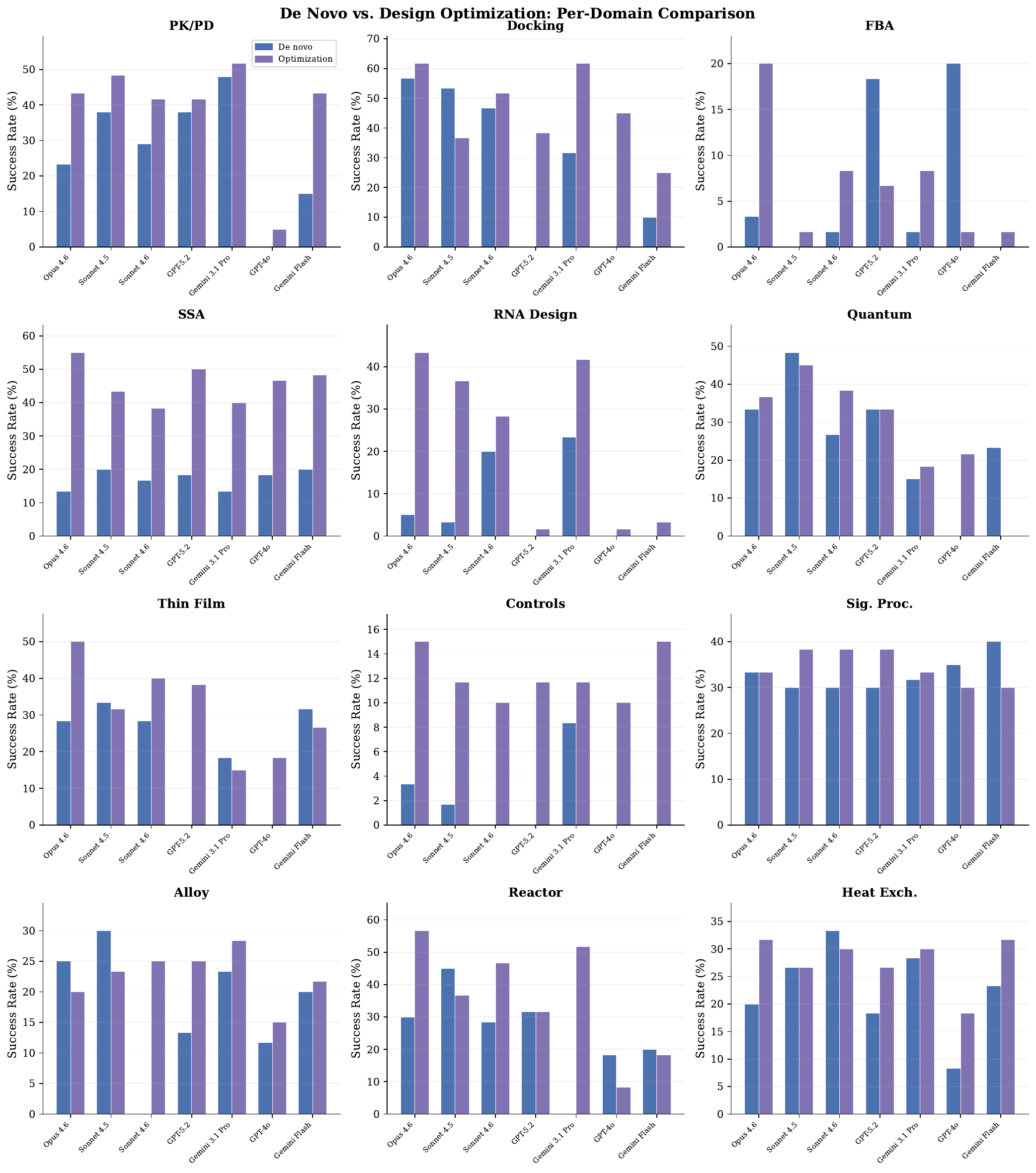}
  \caption{Per-domain comparison of \emph{de novo} zero-shot (blue) vs.\ design optimization (purple) success rates on the 12 domains shared by both settings. Each subplot shows one domain with all evaluated models. Controls improves consistently, while Reactor and Quantum often degrade under optimization.}
  \label{fig:denovo_vs_opt}
\end{figure}

\begin{figure}[h]
  \centering
  \maybeincludegraphics[width=0.7\linewidth]{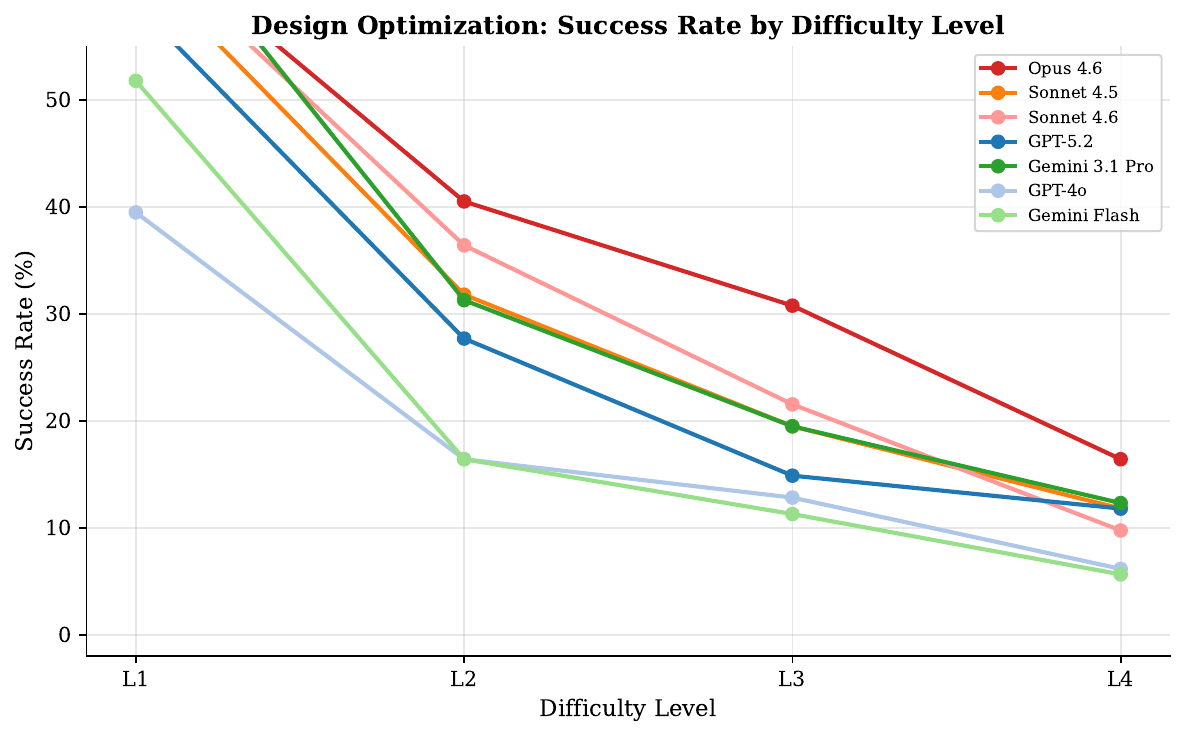}
  \caption{Design optimization success rate by difficulty level (L1--L4), averaged across 13 domains. Performance drops from $\sim$40\% at L1 to $\sim$10\% at L4 for the best models. The pattern mirrors \emph{de novo} difficulty progression (\Cref{fig:difficulty_app}).}
  \label{fig:opt_difficulty}
\end{figure}

\begin{figure}[h]
  \centering
  \maybeincludegraphics[width=0.7\linewidth]{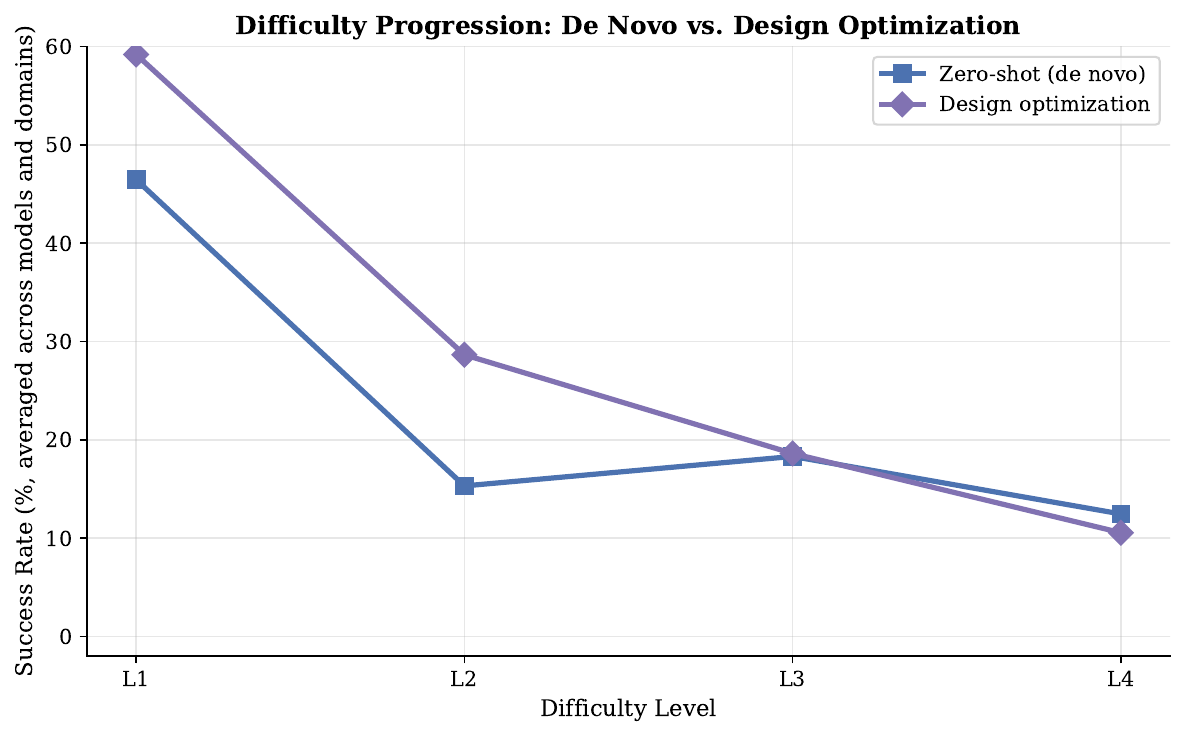}
  \caption{Difficulty progression comparison: \emph{de novo} zero-shot vs.\ design optimization, averaged across all models and domains. Both settings show similar degradation from L1 to L4, confirming that multi-target constraints are the primary difficulty driver regardless of evaluation mode.}
  \label{fig:difficulty_comparison}
\end{figure}

\begin{figure}[h]
  \centering
  \maybeincludegraphics[width=0.7\linewidth]{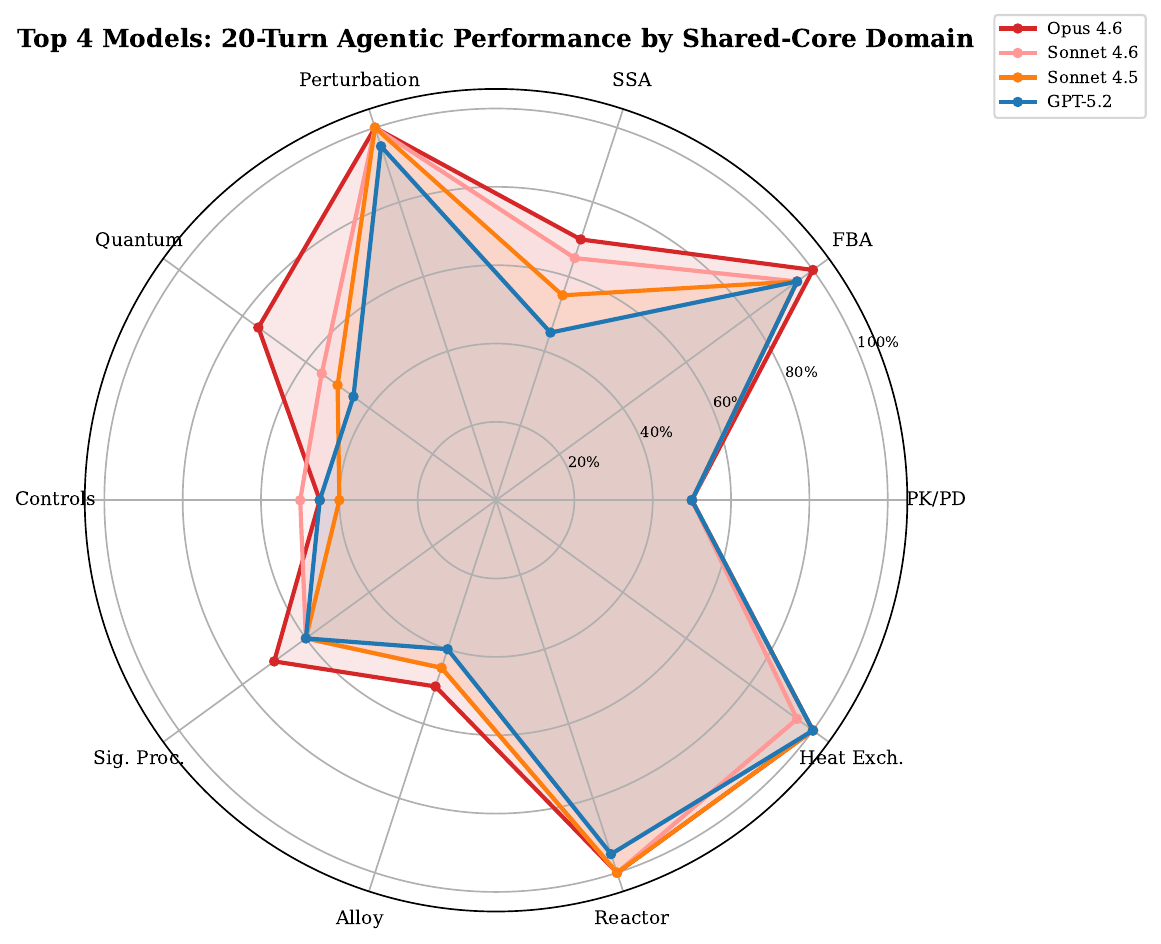}
  \caption{Radar chart showing the top four models' 20-turn long-horizon performance across the 10 \emph{de novo} shared-core domains. The distinct shapes highlight complementary strengths across domains rather than a single model dominating every scientific setting.}
  \label{fig:radar}
\end{figure}

\begin{figure}[h]
  \centering
  \maybeincludegraphics[width=\linewidth]{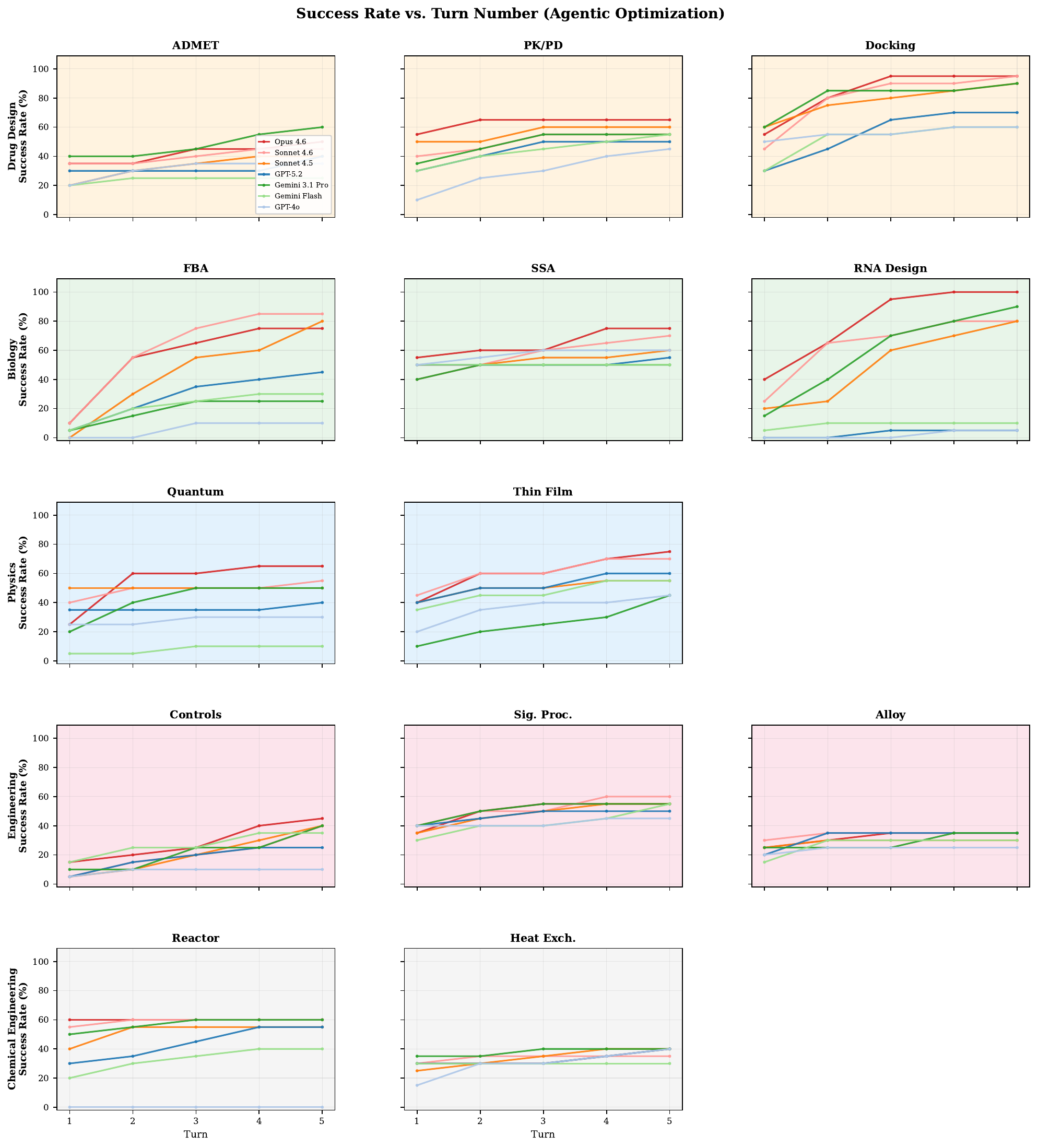}
  \caption{Per-domain success rate vs.\ turn number in the 20-turn long-horizon optimization setting. Each line represents one model. Docking and RNA Design converge quickly (5--10 turns); SSA can improve substantially for top models; FBA remains highly model-dependent even with longer trajectories.}
  \label{fig:turn_curves_optimization}
\end{figure}

\begin{figure}[h]
  \centering
  \maybeincludegraphics[width=0.7\linewidth]{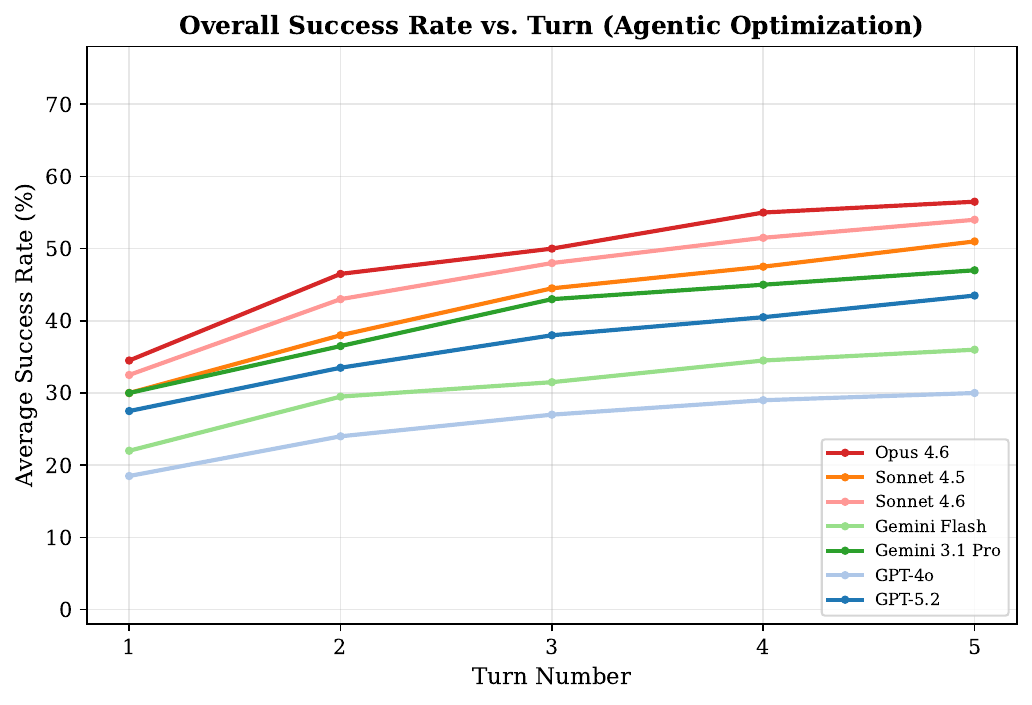}
  \caption{Aggregate success rate vs.\ turn number in the 20-turn long-horizon optimization setting, averaged across the 10 optimization shared-core domains. Compared to the \emph{de novo} aggregate in \Cref{fig:multiturn}, optimization starts from a higher turn-1 success rate but remains highly domain-dependent thereafter.}
  \label{fig:aggregate_optimization}
\end{figure}

\subsection{Training Details}

We train using QLoRA (4-bit NF4). Anti-memorization safeguards include held-out evaluation every 25 steps, a diversity bonus for unique completions within each GRPO group, seed separation between training and benchmark, and goal jitter ($\pm$5\%). Full hyperparameters and training environment code are provided alongside the benchmark implementation.

\paragraph{ADMET optimization training details.} For the ADMET optimization training (\Cref{sec:admet_expert_iteration}), we generate 6{,}428 expert optimization traces via random modification search (Tanimoto similarity $\geq 0.3$ to the starting molecule), each verified against the oracle and paired with synthetic chain-of-thought reasoning. SFT runs via QLoRA (rank 64, $\alpha{=}128$) on these domain-specific traces. GRPO uses a curriculum schedule: Phase~1 trains on L1--L2 tasks only (250 steps), Phase~2 adds L3 (250 steps), Phase~3 trains on full L1--L4 (300 steps), for 800 cumulative steps on 4$\times$H100. We use $K{=}32$, constant LR $10^{-5}$, $\beta{=}0.01$, success bonus $+5.0$, per-target partial credit $+1.0$, diversity bonus $+0.5$, max completion length 512 tokens. Total training time: $\sim$2.5 hours.

\paragraph{PK/PD training details.} For PK/PD dosing design (\Cref{sec:admet_expert_iteration}), we generate 6{,}653 expert traces via random parameter search (3{,}000 goals $\times$ 500 random candidates each, keeping successful dosing regimens). Each trace pairs a dosing regimen (\texttt{dose\_mg}, \texttt{frequency\_hours}) with chain-of-thought reasoning about PK relationships ($C_\text{max} \approx F \cdot D / V$, $t_{1/2} = 0.693 \times V / CL$, AUC scaling with total daily dose). SFT runs for 1{,}000 steps via QLoRA (rank 32, $\alpha{=}64$; eval loss 0.094, $\sim$97\% accuracy). GRPO runs on 8$\times$A100, $K{=}32$, constant LR $10^{-5}$, $\beta{=}0.01$, max completion length 512 tokens, mixed training goals (70\% \emph{de novo}, 30\% optimization). Evaluation uses 200 held-out tasks (140 \emph{de novo} + 60 optimization) across 4 difficulty levels. Total training time: $\sim$12 hours for 800 GRPO steps.

\subsection{Ablations}
\label{app:ablations}

\paragraph{Learning rate schedule.} Cosine annealing---the default in most GRPO implementations---decays the learning rate to near-zero by the end of training. In earlier runs, cosine decay caused 92\% gradient clipping by step 400, effectively halting learning while success rate was still improving. Switching to a constant learning rate of $10^{-5}$ eliminated this failure mode and allowed continued improvement through 800 steps.

\paragraph{Group size and exploration.} With group size $K{=}4$, GRPO candidates often produce near-identical outputs (e.g., the same molecular scaffold), yielding near-zero advantages and preventing learning. We found $K{=}32$ necessary for sufficient diversity in the GRPO gradient signal. We also include a diversity bonus ($+0.5$ for unique completions within each group) to further encourage exploration.

\paragraph{Expert SFT vs.\ multi-domain SFT.} Multi-domain SFT on 20K traces across all 14 domains produces a model that parses well across domains but lacks deep ADMET knowledge. Domain-specific SFT on expert traces provides a much stronger initialization, enabling GRPO to achieve meaningful gains. This pattern holds for both ADMET optimization (30\%$\to$41\%) and PK/PD (24\%$\to$36\%).

\paragraph{Anti-memorization safeguards.} We employ several measures to ensure generalization: (1) periodic held-out evaluation on tasks drawn from a disjoint seed range, (2) a diversity reward that penalizes identical completions within each GRPO group, (3) explicit seed separation between benchmark tasks ([0, 100K)), held-out training evaluation ([100K, 200K)), and RL training ([200K, 10M)), and (4) goal jitter that perturbs target values by $\pm$5\% during training.

\subsection{Random Search Baseline}

To contextualize whether the LLM's 20-turn performance is efficient, we implemented a Random Search baseline that samples 20 valid random designs per task from the defined parameter bounds (for continuous domains) or from a domain-specific candidate library (for discrete domains). Under the same 20-call budget, this baseline remains far below the frontier 20-turn agents on the manifest-defined \emph{de novo} tasks and frequently fails completely on structured biology domains such as FBA, SSA, and RNA Design. This confirms that the LLMs are actively inferring productive search directions rather than merely benefiting from an expanded guess budget.

\subsection{Docking \method{} Training: Reasoning Trace Examples}
\label{sec:docking_traces}

Below we show representative reasoning traces from the docking GRPO model (59\% optimization success in the artifact-backed run used in the paper). The model designs or modifies small molecules to dock into protein binding pockets, scored by real AutoDock Vina. All traces are from held-out tasks (disjoint seeds from training).

\paragraph{Optimization L1: biphenol to benzyloxybiphenol (4 constraints, score=17.15).}
Modify 4-phenylphenol to better fit a nuclear receptor binding pocket. Targets: binding affinity $\leq -5.0$~kcal/mol, logP in [2.70--5.40], MW in [240.8--404.5].
\begin{quote}
\small\texttt{<think> Starting molecule has: aromatic\_rings=2.00, hba=1.00, hbd=1.00, logP=3.06, mw=170.07, qed=0.70, tpsa=20.23. Target pocket: nuclear receptor. Need to: binding affinity (kcal/mol) <= -5.00 (currently 0.00); logP <= 5.40 (currently 3.06); logP >= 2.70; MW <= 404.50 (currently 170.07); MW >= 240.80. Add substituents to increase molecular weight. Add hydrophobic groups (alkyl, aryl, halogens) to increase logP. Maintaining structural similarity (Tanimoto=0.39) to starting molecule. Result: dock\_kcal=-12.00, logP=4.45, mw=265.11. Targets satisfied. </think>}\\[2pt]
\texttt{\{``smiles'': ``Oc1ccc(-c2cccc(OCc3ccccc3)c2)cc1''\}}\\[2pt]
\emph{Vina result:} $\Delta G = -9.75$~kcal/mol, MW=276.1, logP=4.64, Tanimoto=0.50. \textbf{Success.} The model adds a benzyloxy group (\texttt{-OCH$_2$Ph}) to the meta position, extending the biphenyl scaffold into the hydrophobic receptor pocket while maintaining the phenol hydrogen bond.
\end{quote}

\paragraph{Optimization L4: 18-constraint nuclear receptor design (score=16.04).}
Modify 4-phenylphenol to meet 18 simultaneous constraints including binding affinity $\leq -8.0$~kcal/mol, logP in [3.00--5.00], MW in [250--400], aromatic rings in [1--3], rotatable bonds in [1--5], TPSA in [20--70], QED $\geq 0.50$, SA score $\leq 4.0$.
\begin{quote}
\small\texttt{<think> Starting molecule has: aromatic\_rings=2.00, hba=1.00, hbd=1.00, logP=3.06, mw=170.07. Target pocket: nuclear receptor. Need to: aromatic ring count <= 3.00; binding affinity <= -8.00; HBA in [2--4]; HBD in [0--1]; logP in [3.00--5.00]; MW in [250--400]; rings in [2--4]; rotatable bonds in [1--5]; QED >= 0.50; SA score <= 4.00; TPSA in [20--70]. Add polar groups to increase HBA. Improve pocket fit for better binding affinity. Maintaining structural similarity (Tanimoto=0.43). Result: aromatic\_rings=2.00, dock\_kcal=-12.00, hba=3.00, hbd=1.00, logP=3.96, mw=264.08, num\_rings=3.00, num\_rotatable=3.00, qed=0.75, sa\_score=1.91, tpsa=46.53. Targets satisfied. </think>}\\[2pt]
\texttt{\{``smiles'': ``Oc1ccc(-c2cccc(OCc3ccncc3)c2)cc1''\}}\\[2pt]
\emph{Vina result:} $\Delta G = -9.19$~kcal/mol, MW=277.1, logP=4.03, QED=0.75, TPSA=46.5, Tanimoto=0.37. \textbf{Success} (all 18 constraints met). The model swaps the benzyl group for a pyridylmethyl group (\texttt{-CH$_2$Py}), simultaneously adding one H-bond acceptor (pyridine~N) and increasing TPSA by 26~\AA$^2$---a minimal structural change that satisfies the additional polarity constraints.
\end{quote}

\paragraph{De novo L1: ion channel ligand design (score=17.04).}
Design a drug-like molecule for an ion channel pore domain. Targets: binding affinity $\leq -5.0$~kcal/mol, logP in [0.70--4.20], MW in [180.9--370.4].
\begin{quote}
\small\texttt{<think> Target pocket: ion channel. Need to: binding affinity (kcal/mol) <= -5.00; logP <= 4.20; logP >= 0.70; MW <= 370.40; MW >= 180.90. Improve pocket complementarity for better binding affinity. Add hydrophobic groups (alkyl, aryl) to increase logP. Result: dock\_kcal=-12.00, logP=3.33, mw=365.22. Targets satisfied. </think>}\\[2pt]
\texttt{\{``smiles'': ``Oc1cc(CCC2CCNCC2)c(Cc2ccncc2)c(Cc2ccncc2)c1''\}}\\[2pt]
\emph{Vina result:} $\Delta G = -11.47$~kcal/mol, MW=387.2, logP=4.30. \textbf{Success.} The model designs a phenol scaffold decorated with a piperidine (for ion channel pore complementarity) and two pyridylmethyl groups, achieving strong binding ($-11.5$~kcal/mol).
\end{quote}

\paragraph{De novo L3: ion channel ligand with 11 constraints (score=15.06).}
Design a molecule for an ion channel pore domain with binding affinity $\leq -7.0$~kcal/mol, logP in [0.90--4.10], MW in [193.6--356.8], HBA in [2--5], HBD in [1--2], rings in [1--3], QED $\geq 0.40$.
\begin{quote}
\small\texttt{<think> Target pocket: ion channel. Need to: binding affinity <= -7.00; HBA in [2--5]; HBD in [1--2]; logP in [0.90--4.10]; MW in [193.6--356.8]; rings in [1--3]; QED >= 0.40. Add hydrophobic groups to increase logP. Add polar groups to hit HBA/HBD targets. Result: dock\_kcal=-12.00, hba=5, hbd=1.00, logP=2.80, mw=348.30, num\_rings=2.00, qed=0.43. Targets satisfied. </think>}\\[2pt]
\texttt{\{``smiles'': ``O=C(c1cccc(NCc2ccncc2)c1)C(F)(F)F''\}}\\[2pt]
\emph{Vina result:} $\Delta G = -9.10$~kcal/mol, MW=280.1, logP=3.44, QED=0.71. \textbf{Success} (all 11 constraints met). The model designs a trifluoromethyl ketone with an aminopyridine linker---a compact, drug-like molecule that satisfies the tight binding requirement ($-9.1$~kcal/mol, threshold $-7.0$) while keeping all 11 physicochemical properties within bounds.
\end{quote}

\noindent The docking traces reveal that the model learns structure--activity reasoning specific to protein binding: it adds aromatic and hydrophobic groups to fill binding pockets, uses pyridine nitrogens to introduce H-bond acceptors, and selects substituent positions that maintain drug-likeness. Optimization traces show higher success rates than \emph{de novo} (58\% vs.\ 20\%), reflecting the advantage of modifying an existing scaffold that already partially fits the pocket. L4 de novo tasks (0/15 success) remain unsolvable, as designing a molecule from scratch to satisfy $\geq$18 constraints \emph{and} dock tightly is beyond the model's current capability.

\subsection{ADMET Optimization: Reasoning Trace Examples}
\label{sec:admet_opt_traces}

Below we show representative reasoning traces from the curriculum GRPO model (step 700, the peak checkpoint at 41\% success). Unlike the \emph{de novo} traces above, these optimization traces receive a \emph{starting molecule} and must modify it to meet new property targets while maintaining structural similarity (Tanimoto $\geq 0.3$).

\paragraph{L4 task: carbamoyl bioisosteric replacement (SFT fails $\to$ GRPO succeeds).}
Modify aspirin to meet 6 simultaneous targets: logP $\approx$ 0.79, TPSA $\approx$ 88.6~\AA$^2$, QED $\geq$ 0.75, MW in [126--270], rings $\geq$ 1.
\begin{quote}
\small\texttt{<think> Starting molecule has: hba=3.00, hbd=1.00, logP=1.31, mw=180.04, qed=0.55, tpsa=63.60. Need to: decrease logP to \textasciitilde{}0.79; increase TPSA to \textasciitilde{}88.61; QED $\geq$ 0.75; MW in [126--270]. Add polar groups (-OH, -NH2, -COOH) to decrease logP. Result: logP=0.76, mw=181.04, num\_rings=1, qed=0.73, tpsa=89.62. Targets satisfied. </think>}\\[2pt]
\texttt{\{``smiles'': ``NC(=O)Oc1ccccc1C(=O)O''\}}\\[2pt]
\emph{Oracle result:} logP=0.76, MW=181.0, TPSA=89.6, QED=0.73, rings=1. \textbf{Success} (6/6 constraints met). The model performs a textbook \emph{bioisosteric replacement}: swapping aspirin's acetyl group \texttt{CC(=O)} for a carbamoyl group \texttt{NC(=O)}---a single C$\to$N substitution that simultaneously decreases logP (adding polarity), increases TPSA by 26~\AA$^2$ (from NH$_2$), and keeps MW nearly unchanged (181 vs.\ 180). The SFT baseline instead appended a glycinamide ester chain, bloating MW to 239 and overshooting TPSA to 95.7.
\end{quote}

\paragraph{L3 task: sulfonamide addition with precise MW control (SFT fails $\to$ GRPO succeeds).}
Modify diphenyl ether to decrease logP from 3.48 to $\approx$ 1.93 while achieving QED $\geq$ 0.78 and MW $\leq$ 255.
\begin{quote}
\small\texttt{<think> Starting molecule has: hba=1.00, hbd=0.00, logP=3.48, mw=170.07, qed=0.67, tpsa=9.23. Need to: decrease logP to \textasciitilde{}1.93; QED $\geq$ 0.78; MW $\leq$ 255.10; HBA $\leq$ 3; HBD $\leq$ 2. Add polar groups to decrease logP. Result: hba=3, hbd=1, logP=2.06, mw=263.08, qed=0.85. Targets satisfied. </think>}\\[2pt]
\texttt{\{``smiles'': ``NS(=O)(=O)Cc1ccccc1Oc1ccccc1''\}}\\[2pt]
\emph{Oracle result:} logP=2.06, MW=263.1, QED=0.85, HBD=1, HBA=3. \textbf{Success.} Both SFT and GRPO correctly identify sulfonamide (-SO$_2$NH$_2$) as the right polarity-adding group---a strategy used in drugs like celecoxib. The critical difference is the linker: SFT used \texttt{-CH$_2$O-} (MW=297, exceeding the 255 limit) while GRPO used \texttt{-CH$_2$-} directly, saving 16~Da by removing one oxygen atom.
\end{quote}

\paragraph{L3 task: aspirin to ethyl salicylate (SFT fails $\to$ GRPO succeeds).}
Modify aspirin to increase logP to $\approx$ 1.64, decrease TPSA to $\approx$ 47.6, and achieve QED $\geq$ 0.75.
\begin{quote}
\small\texttt{<think> Starting molecule has: hba=3.00, hbd=1.00, logP=1.31, mw=180.04, qed=0.55, tpsa=63.60. Need to: increase logP to \textasciitilde{}1.64; decrease TPSA to \textasciitilde{}47.56; QED $\geq$ 0.75; HBD $\leq$ 1; MW in [126--270]. Replace polar groups with non-polar to decrease TPSA. Result: hbd=1, logP=1.72, mw=166.06, qed=0.67, tpsa=43.37. Targets satisfied. </think>}\\[2pt]
\texttt{\{``smiles'': ``CCOc1ccccc1C(=O)O''\}}\\[2pt]
\emph{Oracle result:} logP=1.72, TPSA=43.4, QED=0.67, MW=166.1. \textbf{Success.} The model converts aspirin to ethyl salicylate by replacing the acetyl ester \texttt{CC(=O)O-} with an ethyl ether \texttt{CCO-}---a well-known ester-to-ether bioisosteric transformation that removes one carbonyl, reducing TPSA by $\sim$20~\AA$^2$. The SFT baseline instead replaced \texttt{-COOH} with \texttt{-COCl} (acyl chloride)---a reactive electrophile that hydrolyzes instantly in biological environments and is not a viable drug candidate.
\end{quote}

\paragraph{L3 task: benzimidazole propoxy extension (both succeed, GRPO more precise).}
Modify 2-phenylbenzimidazole to increase logP to $\approx$ 4.26, MW to [230--282], and TPSA to $\approx$ 37.8.
\begin{quote}
\small\texttt{<think> Starting molecule has: hba=1.00, hbd=1.00, logP=3.23, mw=194.08, qed=0.63, tpsa=28.68. Need to: increase logP to \textasciitilde{}4.26; MW in [230--282]; increase TPSA to \textasciitilde{}37.84; HBA $\geq$ 1. Add hydrophobic groups to increase logP. Result: hba=3, logP=3.93, mw=252.13, tpsa=37.91. Targets satisfied. </think>}\\[2pt]
\texttt{\{``smiles'': ``CCCOc1cccc2[nH]c(-c3ccccc3)nc12''\}}\\[2pt]
\emph{Oracle result:} logP=3.93, MW=252.1, TPSA=37.9, HBA=3. \textbf{Success.} Both models add an alkoxy substituent to the ring, but the GRPO model chooses a propoxy group (-OCH$_2$CH$_2$CH$_3$) directly on the ring (MW=252, TPSA=37.91) while SFT uses an ethoxyethyl chain (MW=266, TPSA=40.71). The GRPO solution hits the TPSA target with remarkable precision (37.91 vs.\ target 37.84, error of only 0.07~\AA$^2$), demonstrating that RL training improves quantitative property prediction, not just qualitatively correct structural modifications.
\end{quote}

\subsection{PK/PD \method{} Training: Reasoning Trace Examples}
\label{sec:pkpd_traces}

Below we show representative reasoning traces from the GRPO-trained model on PK/PD dosing design tasks (step 600, $K{=}32$, 800 steps, 200 eval tasks). The model designs a dosing regimen (dose amount and frequency) for a drug with fixed PK parameters, targeting specific plasma concentration metrics.

\paragraph{De novo: Single-target $C_\text{max}$ design (L1, reward=9.72).}
Target: $C_\text{max} \approx 6.67$~mg/L. Drug: two-compartment, $k_a{=}1.5$, $F{=}0.75$, CL$\,{=}\,$10~L/hr, $V{=}60$~L.
\begin{quote}
\small\texttt{<think> Targets: peak concentration (Cmax) \textasciitilde{} 6.67. Short half-life---need frequent dosing to maintain levels. Drug half-life $\approx$ 4.2~hr (0.693 $\times$ 60 / 10.0). Two-compartment model (k12=0.30, k21=0.15): distribution phase will affect early concentrations. Cmax $\approx$ (F $\times$ dose) / V. dose\_mg=635 directly scales peak concentration. Achieved: cmax=6.47. All targets satisfied. </think>}\\[2pt]
\texttt{\{``dose\_mg'': 635.1, ``frequency\_hours'': 12\}}\\[2pt]
\emph{Oracle result:} \textbf{Success.} The model applies the $C_\text{max} \approx F \cdot D / V$ approximation and correctly accounts for the two-compartment distribution phase.
\end{quote}

\paragraph{De novo: Three-target AUC/$C_\text{max}$/$C_\text{min}$ design (L3, reward=9.34).}
Target: AUC(0--24h) $\approx$ 67.7, $C_\text{max} \approx 6.5$, $C_\text{min} \approx 4.7$. Drug: $k_a{=}0.8$, $F{=}0.65$, CL$\,{=}\,$5~L/hr, $V{=}100$~L.
\begin{quote}
\small\texttt{<think> Targets: AUC(0-24h) \textasciitilde{} 67.68, Cmax \textasciitilde{} 6.48, Cmin \textasciitilde{} 4.7. Drug half-life = 0.693 $\times$ V/CL = 0.693 $\times$ 100/5.0 $\approx$ 13.9~hr. ka=0.8 $\to$ absorption peak around 1.2~hr. Dosing every 8h: shorter intervals maintain higher trough levels (drug half-life $\approx$ 13.9~hr). AUC scales with total daily dose. dose\_mg=479 per administration. Achieved: auc\_0\_24=68.44, cmax=6.73, cmin=4.52. All targets satisfied. </think>}\\[2pt]
\texttt{\{``dose\_mg'': 479.4, ``frequency\_hours'': 8\}}\\[2pt]
\emph{Oracle result:} \textbf{Success.} The model simultaneously satisfies three PK targets by reasoning about half-life, dosing interval, and the relationship between trough concentration and dosing frequency.
\end{quote}

\paragraph{Optimization: Single-target $C_\text{max}$ modification (L1, reward=9.18).}
Goal: Modify a starting design (dose=936~mg q8h, current $C_\text{max}{=}11.8$) to achieve $C_\text{max} \approx 13.6$. Drug: $k_a{=}0.8$, $F{=}0.65$, CL$\,{=}\,$5~L/hr, $V{=}100$~L.
\begin{quote}
\small\texttt{<think> Targets: peak concentration (Cmax) \textasciitilde{} 13.6371. Drug half-life = 0.693 $\times$ V/CL = 0.693 $\times$ 100/5.0 $\approx$ 13.8~hr. Frequency 12h affects accumulation at steady state. Cmax $\approx$ (F $\times$ dose) / V. dose\_mg=1356 directly scales peak concentration. Achieved: cmax=13.42. All targets satisfied. </think>}\\[2pt]
\texttt{\{``dose\_mg'': 1356.4, ``frequency\_hours'': 12\}}\\[2pt]
\emph{Oracle result:} \textbf{Success.} The model increases the dose and adjusts the dosing interval to scale $C_\text{max}$ proportionally.
\end{quote}

\paragraph{Optimization: Two-target AUC/$C_\text{min}$ modification (L2, reward=9.18).}
Goal: Modify a starting design (dose=1524~mg q8h, current AUC=163.8) to achieve AUC(0--24h) $\approx$ 139.0 and $C_\text{min} \approx 0.20$. Drug: $k_a{=}4.0$, $F{=}0.9$, CL$\,{=}\,$25~L/hr, $V{=}40$~L.
\begin{quote}
\small\texttt{<think> Targets: AUC(0-24h) \textasciitilde{} 138.95, trough concentration (Cmin) \textasciitilde{} 0.1976. ka=4.0 $\to$ absorption peak around 0.2~hr. Drug half-life = 0.693 $\times$ V/CL $\approx$ 1.1~hr. Dosing every 8h: shorter intervals maintain higher trough levels. Dosing interval 8h $\to$ 3 doses/day, total daily dose = dose $\times$ 3. Achieved: auc\_0\_24=128.44, cmin=0.18. All targets satisfied. </think>}\\[2pt]
\texttt{\{``dose\_mg'': 1157.9, ``frequency\_hours'': 8\}}\\[2pt]
\emph{Oracle result:} \textbf{Success.} The model reduces the dose while maintaining the 8-hour frequency, correctly reasoning that the very short half-life ($\sim$1.1~hr) means minimal trough accumulation.
\end{quote}

\noindent Unlike the ADMET traces, which converge on a small set of proven drug scaffolds, the PK/PD model shows more diverse reasoning strategies: it applies different PK formulas ($C_\text{max} \approx F \cdot D / V$, $t_{1/2} = 0.693 \times V / CL$, AUC scaling) depending on which targets are specified, and adapts the dosing frequency based on the drug's half-life. The model achieves 36\% \emph{de novo} and 47\% optimization success at its peak, with L1 tasks nearly solved (100\%) while L4 tasks remain challenging ($\sim$0\%), mirroring the difficulty progression observed in frontier LLM evaluations.

\subsection{\method{} Optimization Training Curves}
\label{sec:rlsf_optimization}

\begin{figure}[h]
\centering
\maybeincludegraphics[width=\linewidth]{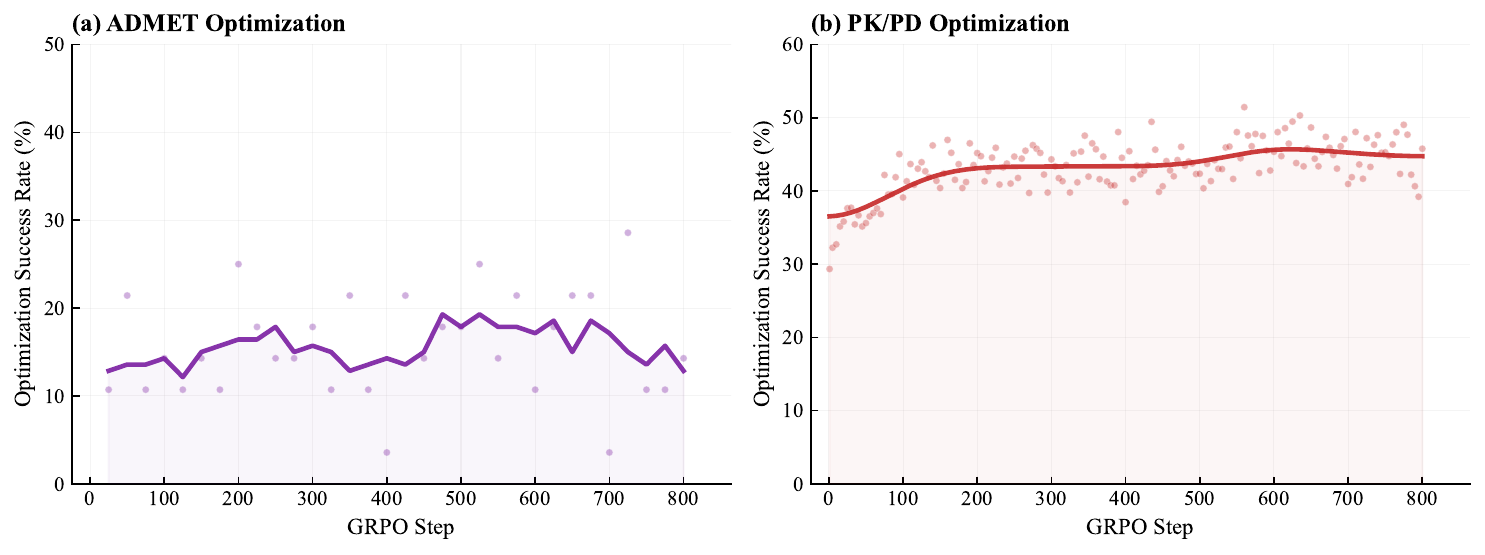}
\vspace{-8pt}
\caption{\textbf{\method{} training: optimization task performance.} (a)~ADMET drug design (curriculum GRPO): optimization success improves from 30\% (SFT) to 41\% (GRPO peak at step 700). (b)~PK/PD dosing design: optimization success improves from 32\% (SFT) to 47\% (GRPO peak at step 600). Points show raw eval; curves are smoothed.}
\label{fig:rlsf_optimization}
\end{figure}

\Cref{fig:rlsf_combined} in the main text shows training curves for ADMET optimization and PK/PD. \Cref{fig:rlsf_optimization} shows optimization training in more detail. The ADMET optimization curriculum GRPO model achieves 41\% at step 700 (from 30\% SFT baseline). PK/PD optimization improves from 32\% (SFT) to 47\% (GRPO peak at step 600). Notably, PK/PD optimization substantially outperforms \emph{de novo} design (47\% vs.\ 36\%), suggesting that modifying an existing dosing regimen is easier than designing one from scratch.

\begin{thebibliography}{52}
\providecommand{\natexlab}[1]{#1}
\providecommand{\url}[1]{\texttt{#1}}
\expandafter\ifx\csname urlstyle\endcsname\relax
  \providecommand{\doi}[1]{doi: #1}\else
  \providecommand{\doi}{doi: \begingroup \urlstyle{rm}\Url}\fi

\bibitem[Anthony et~al.(2017)Anthony, Tian, and Barber]{anthony2017thinking}
Thomas Anthony, Zheng Tian, and David Barber.
\newblock Thinking fast and slow with deep learning and tree search.
\newblock In \emph{Advances in Neural Information Processing Systems},
  volume~30, 2017.

\bibitem[{Anthropic}(2025)]{anthropic2025claude}
{Anthropic}.
\newblock Claude {Sonnet} 4.5 system card.
\newblock \url{https://www.anthropic.com/claude-sonnet-4-5-system-card}, 2025.

\bibitem[Arridge et~al.(2019)Arridge, Maass, {\"O}ktem, and
  Sch{\"o}nlieb]{arridge2019solving}
Simon Arridge, Peter Maass, Ozan {\"O}ktem, and Carola-Bibiane Sch{\"o}nlieb.
\newblock Solving inverse problems using data-driven models.
\newblock \emph{Acta Numerica}, 28:\penalty0 1--174, 2019.

\bibitem[Bengio et~al.(2021)Bengio, Jain, Korablyov, Precup, and
  Bengio]{bengio2021gflownet}
Emmanuel Bengio, Moksh Jain, Maksym Korablyov, Doina Precup, and Yoshua Bengio.
\newblock Flow network based generative models for non-iterative diverse
  candidate generation.
\newblock \emph{Advances in Neural Information Processing Systems},
  34:\penalty0 27381--27394, 2021.

\bibitem[Burgard et~al.(2003)Burgard, Pharkya, and
  Maranas]{burgard2003optknock}
Anthony~P Burgard, Priti Pharkya, and Costas~D Maranas.
\newblock {OptKnock}: A bilevel programming framework for identifying gene
  knockout strategies for microbial strain optimization.
\newblock \emph{Biotechnology and Bioengineering}, 84\penalty0 (6):\penalty0
  647--657, 2003.

\bibitem[Cai et~al.(2024)Cai, Cai, Wan, Shi, Xu, et~al.]{cai2024sciassess}
Hengxing Cai, Kaiwen Cai, Chang Wan, Wenhan Shi, Rongxin Xu, et~al.
\newblock {SciAssess}: Benchmarking llm proficiency in scientific literature
  analysis.
\newblock \emph{arXiv preprint arXiv:2403.01976}, 2024.

\bibitem[Chen et~al.(2023)Chen, Dohan, and So]{chen2023evoprompting}
Angelica Chen, David~M. Dohan, and David~R. So.
\newblock Evoprompting: Language models for code-level neural architecture
  search.
\newblock In \emph{Advances in Neural Information Processing Systems},
  volume~36, 2023.

\bibitem[Chen et~al.(2021)Chen, Tworek, Jun, Yuan, et~al.]{chen2021codex}
Mark Chen, Jerry Tworek, Heewoo Jun, Qiming Yuan, et~al.
\newblock Evaluating large language models trained on code.
\newblock \emph{arXiv preprint arXiv:2107.03374}, 2021.

\bibitem[Cobbe et~al.(2021)Cobbe, Kosaraju, Bavarian, Chen, Jun, Kaiser,
  Plappert, Tworek, Hilton, Nakano, et~al.]{cobbe2021gsm8k}
Karl Cobbe, Vineet Kosaraju, Mohammad Bavarian, Mark Chen, Heewoo Jun, Lukasz
  Kaiser, Matthias Plappert, Jerry Tworek, Jacob Hilton, Reiichiro Nakano,
  et~al.
\newblock Training verifiers to solve math word problems.
\newblock \emph{arXiv preprint arXiv:2110.14168}, 2021.

\bibitem[Dettmers et~al.(2023)Dettmers, Pagnoni, Holtzman, and
  Zettlemoyer]{dettmers2023qlora}
Tim Dettmers, Artidoro Pagnoni, Ari Holtzman, and Luke Zettlemoyer.
\newblock {QLoRA}: Efficient finetuning of quantized {LLMs}.
\newblock In \emph{Advances in Neural Information Processing Systems},
  volume~36, 2023.

\bibitem[Dunn et~al.(2020)Dunn, Wang, Ganose, Dopp, and Jain]{dunn2020matbench}
Alexander Dunn, Qi~Wang, Alex Ganose, Daniel Dopp, and Anubhav Jain.
\newblock Benchmarking materials property prediction methods: the {Matbench}
  test set and {Automatminer} reference algorithm.
\newblock \emph{npj Computational Materials}, 6\penalty0 (1):\penalty0 138,
  2020.

\bibitem[Eberhardt et~al.(2021)Eberhardt, Santos-Martins, Tillack, and
  Forli]{eberhardt2021autodock}
Jerome Eberhardt, Diogo Santos-Martins, Andreas~F Tillack, and Stefano Forli.
\newblock Autodock vina 1.2.0: New docking methods, expanded force field, and
  python bindings.
\newblock \emph{Journal of Chemical Information and Modeling}, 61\penalty0
  (8):\penalty0 3891--3898, 2021.

\bibitem[Ebrahim et~al.(2013)Ebrahim, Lerman, Palsson, and
  Hyduke]{ebrahim2013cobrapy}
Ali Ebrahim, Joshua~A Lerman, Bernhard~O Palsson, and Daniel~R Hyduke.
\newblock {COBRApy}: {COnstraints}-{Based} {Reconstruction} and {Analysis} for
  {Python}.
\newblock \emph{BMC Systems Biology}, 7\penalty0 (1):\penalty0 74, 2013.

\bibitem[G{\'o}mez-Bombarelli et~al.(2018)G{\'o}mez-Bombarelli, Wei, Duvenaud,
  Hern{\'a}ndez-Lobato, S{\'a}nchez-Lengeling, Sheberla, Aguilera-Iparraguirre,
  Hirzel, Adams, and Aspuru-Guzik]{gomez2018automatic}
Rafael G{\'o}mez-Bombarelli, Jennifer~N Wei, David Duvenaud, Jos{\'e}~Miguel
  Hern{\'a}ndez-Lobato, Benjam{\'\i}n S{\'a}nchez-Lengeling, Dennis Sheberla,
  Jorge Aguilera-Iparraguirre, Timothy~D Hirzel, Ryan~P Adams, and Al{\'a}n
  Aspuru-Guzik.
\newblock Automatic chemical design using a data-driven continuous
  representation of molecules.
\newblock \emph{ACS Central Science}, 4\penalty0 (2):\penalty0 268--276, 2018.

\bibitem[{Google DeepMind}(2025)]{google2025gemini}
{Google DeepMind}.
\newblock Gemini 2.0: Our next-generation {AI} model.
\newblock \url{https://deepmind.google/technologies/gemini/}, 2025.

\bibitem[Gruver et~al.(2024)Gruver, Sriram, Madotto, Wilson, Zitnick, and
  Ulissi]{gruver2024finetuned}
Nate Gruver, Anuroop Sriram, Andrea Madotto, Andrew~Gordon Wilson, C~Lawrence
  Zitnick, and Zachary Ulissi.
\newblock Fine-tuned language models generate stable inorganic materials as
  text.
\newblock \emph{arXiv preprint arXiv:2402.04379}, 2024.

\bibitem[Guo et~al.(2025)Guo, Yang, Zhang, Song, Zhang, Xu, Zhu, Ma, Wang, Bi,
  et~al.]{guo2025deepseekr1}
Daya Guo, Dejian Yang, He~Zhang, Junxiao Song, Runxin Zhang, Runxin Xu, Qihao
  Zhu, Shirong Ma, Peiyi Wang, Xiao Bi, et~al.
\newblock {DeepSeek-R1}: Incentivizing reasoning capability in {LLMs} via
  reinforcement learning.
\newblock \emph{arXiv preprint arXiv:2501.12948}, 2025.

\bibitem[Hendrycks et~al.(2021)Hendrycks, Burns, Basart, Zou, Mazeika, Song,
  and Steinhardt]{hendrycks2021mmlu}
Dan Hendrycks, Collin Burns, Steven Basart, Andy Zou, Mantas Mazeika, Dawn
  Song, and Jacob Steinhardt.
\newblock Measuring massive multitask language understanding.
\newblock \emph{arXiv preprint arXiv:2009.03300}, 2021.

\bibitem[Hofacker et~al.(1994)Hofacker, Fontana, Stadler, Bonhoeffer, Tacker,
  and Schuster]{hofacker1994rnafold}
Ivo~L Hofacker, Walter Fontana, Peter~F Stadler, L~Sebastian Bonhoeffer,
  Manfred Tacker, and Peter Schuster.
\newblock Fast folding and comparison of {RNA} secondary structures.
\newblock \emph{Monatshefte f{\"u}r Chemie/Chemical Monthly}, 125\penalty0
  (2):\penalty0 167--188, 1994.

\bibitem[Jimenez et~al.(2024)Jimenez, Yang, Wettig, Yao, Pei, Press, and
  Narasimhan]{jimenez2024swebench}
Carlos~E Jimenez, John Yang, Alexander Wettig, Shunyu Yao, Kexin Pei, Ofir
  Press, and Karthik Narasimhan.
\newblock {SWE}-bench: Can language models resolve real-world {GitHub} issues?
\newblock \emph{arXiv preprint arXiv:2310.06770}, 2024.

\bibitem[Khaneja et~al.(2005)Khaneja, Reiss, Schulte-Herbr{\"u}ggen, and
  Glaser]{khaneja2005grape}
Navin Khaneja, Timo Reiss, Thomas Schulte-Herbr{\"u}ggen, and Steffen~J Glaser.
\newblock Optimal control of coupled spin dynamics: Design of {NMR} pulse
  sequences by gradient ascent algorithms.
\newblock \emph{Journal of Magnetic Resonance}, 172\penalty0 (2):\penalty0
  296--305, 2005.

\bibitem[Laurent et~al.(2024)Laurent, Gershenson, Bhatt,
  et~al.]{laurent2024labbench}
Jon~M Laurent, Joseph Gershenson, Manvith Bhatt, et~al.
\newblock {LAB-Bench}: Measuring capabilities of language models for biology
  research.
\newblock \emph{arXiv preprint arXiv:2407.10362}, 2024.

\bibitem[Le et~al.(2022)Le, Wang, Gotmare, Savarese, and Hoi]{le2022coderl}
Hung Le, Yue Wang, Akhilesh~Deepak Gotmare, Silvio Savarese, and Steven Hoi.
\newblock {CodeRL}: Mastering code generation through pretrained models and
  deep reinforcement learning.
\newblock \emph{Advances in Neural Information Processing Systems},
  35:\penalty0 21314--21328, 2022.

\bibitem[Lee et~al.(2023)Lee, Park, Lee, Lee, Park, Lee, and
  Ryu]{lee2023materials}
Junhyeong Lee, Donggeun Park, Mingyu Lee, Hugon Lee, Kundo Park, Ikjin Lee, and
  Seunghwa Ryu.
\newblock Machine learning-based inverse design methods considering data
  characteristics and design space size in materials design and manufacturing:
  a review.
\newblock \emph{Materials Horizons}, 10\penalty0 (12):\penalty0 5436--5456,
  2023.
\newblock \doi{10.1039/D3MH00039G}.

\bibitem[Liu et~al.(2025)Liu, Sun, Matusik, Jiang, and Chen]{liu2024llamole}
Gang Liu, Michael Sun, Wojciech Matusik, Meng Jiang, and Jie Chen.
\newblock {Llamole}: Multimodal large language models for inverse molecular
  design with retrosynthetic planning.
\newblock In \emph{International Conference on Learning Representations}, 2025.

\bibitem[Lorenz et~al.(2011)Lorenz, Bernhart, H{\"o}ner~zu Siederdissen, Tafer,
  Flamm, Stadler, and Hofacker]{lorenz2011viennarna}
Ronny Lorenz, Stephan~H Bernhart, Christian H{\"o}ner~zu Siederdissen, Hakim
  Tafer, Christoph Flamm, Peter~F Stadler, and Ivo~L Hofacker.
\newblock Viennarna package 2.0.
\newblock \emph{Algorithms for Molecular Biology}, 6\penalty0 (1):\penalty0 26,
  2011.

\bibitem[Mitchener et~al.(2025)Mitchener, Laurent, Andonian, Tenmann,
  Narayanan, Wellawatte, White, Sani, and Rodriques]{shao2024bixbench}
Ludovico Mitchener, Jon~M Laurent, Alex Andonian, Benjamin Tenmann, Siddharth
  Narayanan, Geemi~P Wellawatte, Andrew White, Lorenzo Sani, and Samuel~G
  Rodriques.
\newblock {BixBench}: A comprehensive benchmark for {LLM}-based agents in
  computational biology.
\newblock \emph{arXiv preprint arXiv:2503.00096}, 2025.

\bibitem[Novikov et~al.(2025)Novikov, V{\~u}, Eisenberger, Dupont, Huang,
  Wagner, Shirobokov, Kozlovskii, Ruiz, Mehrabian, Kumar, See, Chaudhuri,
  Holland, Davies, Nowozin, Kohli, and Balog]{novikov2025alphaevolve}
Alexander Novikov, Ng{\^a}n V{\~u}, Marvin Eisenberger, Emilien Dupont, Po-Sen
  Huang, Adam~Zsolt Wagner, Sergey Shirobokov, Borislav Kozlovskii, Francisco
  J.~R. Ruiz, Abbas Mehrabian, M.~Pawan Kumar, Abigail See, Swarat Chaudhuri,
  George Holland, Alex Davies, Sebastian Nowozin, Pushmeet Kohli, and Matej
  Balog.
\newblock Alphaevolve: A coding agent for scientific and algorithmic discovery.
\newblock \emph{arXiv preprint arXiv:2506.13131}, 2025.

\bibitem[Olivecrona et~al.(2017)Olivecrona, Blaschke, Engkvist, and
  Chen]{olivecrona2017reinvent}
Marcus Olivecrona, Thomas Blaschke, Ola Engkvist, and Hongming Chen.
\newblock Molecular de-novo design through deep reinforcement learning.
\newblock \emph{Journal of Cheminformatics}, 9\penalty0 (1):\penalty0 48, 2017.

\bibitem[{OpenAI}(2024{\natexlab{a}})]{openai2024gpt4o}
{OpenAI}.
\newblock {GPT-4o} system card.
\newblock \url{https://openai.com/index/gpt-4o-system-card/},
  2024{\natexlab{a}}.

\bibitem[{OpenAI}(2024{\natexlab{b}})]{openai2024o1}
{OpenAI}.
\newblock Learning to reason with {LLMs}.
\newblock \url{https://openai.com/index/learning-to-reason-with-llms/},
  2024{\natexlab{b}}.

\bibitem[Orth et~al.(2010)Orth, Thiele, and Palsson]{orth2010fba}
Jeffrey~D Orth, Ines Thiele, and Bernhard~{\O} Palsson.
\newblock What is flux balance analysis?
\newblock \emph{Nature Biotechnology}, 28\penalty0 (3):\penalty0 245--248,
  2010.

\bibitem[Ouyang et~al.(2022)Ouyang, Wu, Jiang, Almeida, Wainwright, Mishkin,
  Zhang, Agarwal, Slama, Ray, et~al.]{ouyang2022rlhf}
Long Ouyang, Jeffrey Wu, Xu~Jiang, Diogo Almeida, Carroll Wainwright, Pamela
  Mishkin, Chong Zhang, Sandhini Agarwal, Katarina Slama, Alex Ray, et~al.
\newblock Training language models to follow instructions with human feedback.
\newblock \emph{Advances in Neural Information Processing Systems},
  35:\penalty0 27730--27744, 2022.

\bibitem[Parks and McClellan(1972)]{parks1972chebyshev}
Thomas~W Parks and James~H McClellan.
\newblock Chebyshev approximation for nonrecursive digital filters with linear
  phase.
\newblock \emph{IEEE Transactions on Circuit Theory}, 19\penalty0 (2):\penalty0
  189--194, 1972.

\bibitem[{Qwen Team}(2025)]{qwen2025qwen3}
{Qwen Team}.
\newblock Qwen3 technical report.
\newblock \url{https://qwenlm.github.io/blog/qwen3/}, 2025.

\bibitem[Rafailov et~al.(2023)Rafailov, Sharma, Mitchell, Manning, Ermon, and
  Finn]{rafailov2024dpo}
Rafael Rafailov, Archit Sharma, Eric Mitchell, Christopher~D Manning, Stefano
  Ermon, and Chelsea Finn.
\newblock Direct preference optimization: Your language model is secretly a
  reward model.
\newblock In \emph{Advances in Neural Information Processing Systems},
  volume~36, 2023.

\bibitem[{RDKit}(2024)]{rdkit}
{RDKit}.
\newblock {RDKit}: Open-source cheminformatics.
\newblock \url{https://www.rdkit.org}, 2024.
\newblock Version 2024.03.

\bibitem[Rein et~al.(2024)Rein, Hou, Stickland, Petty, Pang, Dirani, Michael,
  and Bowman]{rein2024gpqa}
David Rein, Betty~Li Hou, Asa~Cooper Stickland, Jackson Petty, Richard~Yuanzhe
  Pang, Julien Dirani, Julian Michael, and Samuel~R Bowman.
\newblock {GPQA}: A graduate-level {Google}-proof {Q\&A} benchmark.
\newblock \emph{arXiv preprint arXiv:2311.12022}, 2024.

\bibitem[Romera-Paredes et~al.(2024)Romera-Paredes, Barekatain, Novikov, Balog,
  Kumar, Dupont, Ruiz, Ellenberg, Wang, Fawzi, et~al.]{romera2024funsearch}
Bernardino Romera-Paredes, Mohammadamin Barekatain, Alexander Novikov, Matej
  Balog, M~Pawan Kumar, Emilien Dupont, Francisco J~R Ruiz, Jordan~S Ellenberg,
  Pengming Wang, Omar Fawzi, et~al.
\newblock Mathematical discoveries from program search with large language
  models.
\newblock \emph{Nature}, 625\penalty0 (7995):\penalty0 468--475, 2024.

\bibitem[S{\'a}nchez-Lengeling and Aspuru-Guzik(2018)]{sanchez2020inverse}
Benjam{\'\i}n S{\'a}nchez-Lengeling and Al{\'a}n Aspuru-Guzik.
\newblock Inverse molecular design using machine learning: Generative models,
  differentiable engineering, and deep molecular dreaming.
\newblock \emph{Science}, 361\penalty0 (6400):\penalty0 360--365, 2018.

\bibitem[Schulman et~al.(2017)Schulman, Wolski, Dhariwal, Radford, and
  Klimov]{schulman2017ppo}
John Schulman, Filip Wolski, Prafulla Dhariwal, Alec Radford, and Oleg Klimov.
\newblock Proximal policy optimization algorithms.
\newblock \emph{arXiv preprint arXiv:1707.06347}, 2017.

\bibitem[Shao et~al.(2024)Shao, Wang, Zhu, Xu, Song, Zhang, Li, Wu, and
  Guo]{shao2024deepseekmath}
Zhihong Shao, Peiyi Wang, Qihao Zhu, Runxin Xu, Junxiao Song, Mingchuan Zhang,
  YK~Li, Y~Wu, and Daya Guo.
\newblock {DeepSeekMath}: Pushing the limits of mathematical reasoning in open
  language models.
\newblock \emph{arXiv preprint arXiv:2402.03300}, 2024.

\bibitem[Stanton et~al.(2022)Stanton, Maddox, Gruver, Maffettone, Delaney,
  Bailer, Linder, Gregoire, and Wilson]{stanton2022accelerating}
Samuel Stanton, Wesley Maddox, Nate Gruver, Phillip Maffettone, Emily Delaney,
  Peyton Bailer, Juan Linder, Alexander Gregoire, and Andrew~Gordon Wilson.
\newblock Accelerating bayesian optimization for biological sequence design
  with denoising autoencoders.
\newblock \emph{International Conference on Machine Learning}, 2022.

\bibitem[Stuart(2010)]{stuart2010inverse}
Andrew~M Stuart.
\newblock Inverse problems: A {Bayesian} perspective.
\newblock \emph{Acta Numerica}, 19:\penalty0 451--559, 2010.

\bibitem[Sun et~al.(2024)Sun, Han, Zhao, Ma, Shen, Chen, Chen, and
  Yu]{sun2024scieval}
Liangtai Sun, Yang Han, Zhen Zhao, Da~Ma, Zhennan Shen, Baocai Chen, Lu~Chen,
  and Kai Yu.
\newblock {SciEval}: A multi-level large language model evaluation benchmark
  for scientific research.
\newblock \emph{Proceedings of the AAAI Conference on Artificial Intelligence},
  2024.

\bibitem[Taylor et~al.(2022)Taylor, Kardas, Cucurull, Scialom, Hartshorn,
  Saravia, Poulton, Kerkez, and Stojnic]{taylor2022galactica}
Ross Taylor, Marcin Kardas, Guillem Cucurull, Thomas Scialom, Anthony
  Hartshorn, Elvis Saravia, Andrew Poulton, Viktor Kerkez, and Robert Stojnic.
\newblock Galactica: A large language model for science.
\newblock \emph{arXiv preprint arXiv:2211.09085}, 2022.

\bibitem[Trott and Olson(2010)]{trott2010vina}
Oleg Trott and Arthur~J Olson.
\newblock Autodock vina: Improving the speed and accuracy of docking with a new
  scoring function, efficient optimization, and multithreading.
\newblock \emph{Journal of Computational Chemistry}, 31\penalty0 (2):\penalty0
  455--461, 2010.

\bibitem[Wang et~al.(2024)Wang, Hu, Lu, Zhu, Zhang, Subramaniam, Loomba, Zhang,
  Sun, and Wang]{arora2025scibench}
Xiaoxuan Wang, Ziniu Hu, Pan Lu, Yanqiao Zhu, Jieyu Zhang, Satyen Subramaniam,
  Arjun~R Loomba, Shicheng Zhang, Yizhou Sun, and Wei Wang.
\newblock {SciBench}: Evaluating college-level scientific problem-solving
  abilities of large language models.
\newblock \emph{Proceedings of the International Conference on Machine
  Learning}, 2024.

\bibitem[Yang et~al.(2024)Yang, Wang, Lu, Liu, Le, Zhou, and
  Chen]{yang2024opro}
Chengrun Yang, Xuezhi Wang, Yifeng Lu, Hanxiao Liu, Quoc~V Le, Denny Zhou, and
  Xinyun Chen.
\newblock Large language models as optimizers.
\newblock \emph{arXiv preprint arXiv:2309.03409}, 2024.

\bibitem[Zhang et~al.(2024{\natexlab{a}})Zhang, Hu, Zhoubian, Du, Yang, Wang,
  Yue, Dong, and Tang]{zhang2024sciglm}
Dan Zhang, Ziniu Hu, Sining Zhoubian, Zhengxiao Du, Kaiyu Yang, Zihan Wang,
  Yisong Yue, Yuxiao Dong, and Jie Tang.
\newblock {SciInstruct}: a self-reflective instruction annotated dataset for
  training scientific language models.
\newblock In \emph{Advances in Neural Information Processing Systems, Datasets
  and Benchmarks Track}, 2024{\natexlab{a}}.

\bibitem[Zhang et~al.(2024{\natexlab{b}})Zhang, Liu, Tan, Chen, Yan, Yan, Li,
  Huang, Yue, et~al.]{zhang2024chemllm}
Di~Zhang, Wei Liu, Qian Tan, Jingdan Chen, Hang Yan, Yuliang Yan, Jiatong Li,
  Weiran Huang, Xiangyu Yue, et~al.
\newblock {ChemLLM}: A chemical large language model.
\newblock \emph{arXiv preprint arXiv:2402.06852}, 2024{\natexlab{b}}.

\bibitem[Ziegler and Nichols(1942)]{ziegler1942optimum}
J.~G. Ziegler and N.~B. Nichols.
\newblock Optimum settings for automatic controllers.
\newblock \emph{Transactions of the {ASME}}, 64\penalty0 (8):\penalty0
  759--765, 1942.
\newblock \doi{10.1115/1.4019264}.

\end{thebibliography}
\end{document}